\documentclass[10pt,twocolumn,letterpaper]{article}

\usepackage{cvpr}              

\usepackage{graphicx}
\usepackage{amsmath}
\usepackage{amssymb}
\usepackage{booktabs}
\usepackage{mathtools}
\usepackage{tabularx}

\usepackage{graphicx}
\usepackage{comment}
\usepackage{multirow}
\usepackage{nicefrac}
\usepackage[accsupp]{axessibility}  
\graphicspath{ {./images/} }

\newcommand\blfootnote[1]{%
  \begingroup
  \renewcommand\thefootnote{}\footnote{#1}%
  \addtocounter{footnote}{-1}%
  \endgroup
}
\usepackage[breaklinks=true,
            bookmarks=false,
            colorlinks,
            linkcolor=red,       
            anchorcolor=blue,  
            citecolor=green, 
            ]{hyperref}

\usepackage[capitalize]{cleveref}
\crefname{section}{Sec.}{Secs.}
\Crefname{section}{Section}{Sections}
\Crefname{table}{Table}{Tables}
\crefname{table}{Tab.}{Tabs.}
\newtheorem{corollary}{Corollary}

\def\cvprPaperID{7361} 
\def\confName{CVPR}
\def\confYear{2022}

\usepackage{overpic}
\usepackage{enumitem} 
\usepackage{overpic} 
\usepackage{color}

\definecolor{turquoise}{cmyk}{0.65,0,0.1,0.3}
\definecolor{purple}{rgb}{0.65,0,0.65}
\definecolor{dark_green}{rgb}{0, 0.5, 0}
\definecolor{orange}{rgb}{0.8, 0.6, 0.2}
\definecolor{red}{rgb}{0.8, 0.2, 0.2}
\definecolor{darkred}{rgb}{0.6, 0.1, 0.05}
\definecolor{blueish}{rgb}{0.0, 0.3, .6}
\definecolor{light_gray}{rgb}{0.7, 0.7, .7}
\definecolor{pink}{rgb}{1, 0, 1}
\definecolor{greyblue}{rgb}{0.25, 0.25, 1}

\usepackage{blindtext}

\renewcommand{\paragraph}[1]{\noindent\textbf{#1}.}

\usepackage{amssymb}
\usepackage{pifont}

\begin{document}
\title{3DeformRS: Certifying Spatial Deformations on Point Clouds}
\author{
    Gabriel Pérez~S.$^{*,1,2}$
    \and
    Juan~C. Pérez$^{*,2}$
    \and
    Motasem Alfarra$^{*,2}$
    \and
    Silvio Giancola$^{2}$
    \and
    Bernard Ghanem$^{2}$\\
    \normalsize $^{1}$Universidad Nacional de Colombia, \normalsize $^{2}$King Abdullah University of Science and Technology (KAUST)\\
    \tt\small $^{1}$gaperezsa\texttt{@unal.edu.co}\\
    \tt\small $^{2}$\{\texttt{juan.perezsantamaria,motasem.alfarra,silvio.giancola,bernard.ghanem}\}\texttt{@kaust.edu.sa}\\
}
\maketitle
\begin{abstract}
    3D computer vision models are commonly used in security-critical applications such as autonomous driving and surgical robotics.
    Emerging concerns over the robustness of these models against real-world deformations must be addressed practically and reliably.
    In this work, we propose 3DeformRS, a method to certify the robustness of point cloud Deep Neural Networks~(DNNs) against real-world deformations.
    We developed 3DeformRS by building upon recent work that generalized Randomized Smoothing~(RS) from pixel-intensity perturbations to vector-field deformations.
    In particular, we specialized RS to certify DNNs against parameterized deformations (\eg rotation, twisting), while enjoying practical computational costs.
    We leverage the virtues of 3DeformRS to conduct a comprehensive empirical study on the certified robustness of four representative point cloud DNNs on two datasets and against seven different deformations. 
    Compared to previous approaches for certifying point cloud DNNs, 3DeformRS is fast, scales well with point cloud size, and provides comparable-to-better certificates.
    For instance, when certifying a plain PointNet against a 3$^\circ$ $z-$rotation on 1024-point clouds, 3DeformRS grants a certificate $3\times$ larger and $20\times$ faster than previous work\blfootnote{* Denotes Equal Contribution. The order is random.}\blfootnote{Work done during Gabriel's internship at KAUST.}\footnote{Code: \url{https://github.com/gaperezsa/3DeformRS}}.
\end{abstract}
\section{Introduction}\label{sec:intro}
\begin{figure}[t]
    \centering
    \includegraphics[width=\columnwidth]{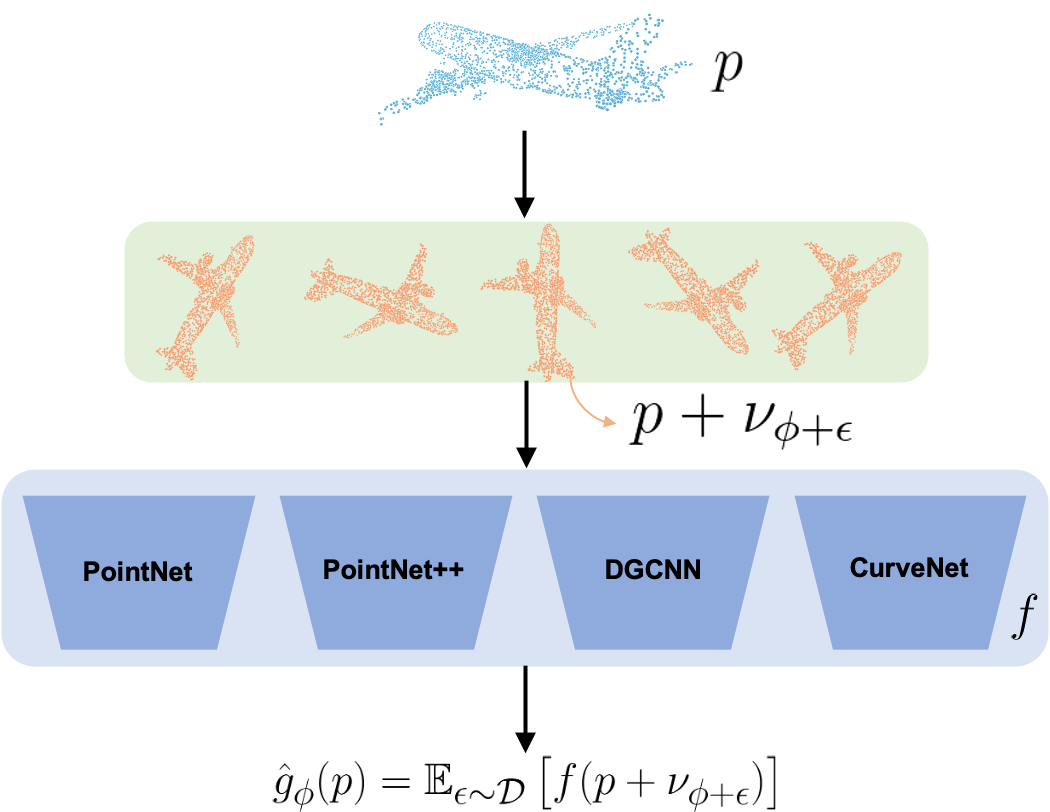} 
    \caption{\label{fig:PullingFigure}
        \textbf{Randomized Smoothing for Point Cloud DNNs.}
        We certify the prediction of a DNN on the input point cloud via randomized smoothing by constructing a \emph{smooth} DNN $\hat{g}_\phi$ around the original DNN $f$.
        For an input $p$, a parametric transformation $\nu_{\phi+\epsilon}$, and a parameter $\sigma$, the Smooth DNN predicts the expected value of the predictions from transformed versions of $p$.
    }
\end{figure}
\begin{figure*}[t]
    \centering
    \frame{\includegraphics[trim=14cm 7cm 14cm 7cm,clip,width=0.196\textwidth,height=0.14\textheight]{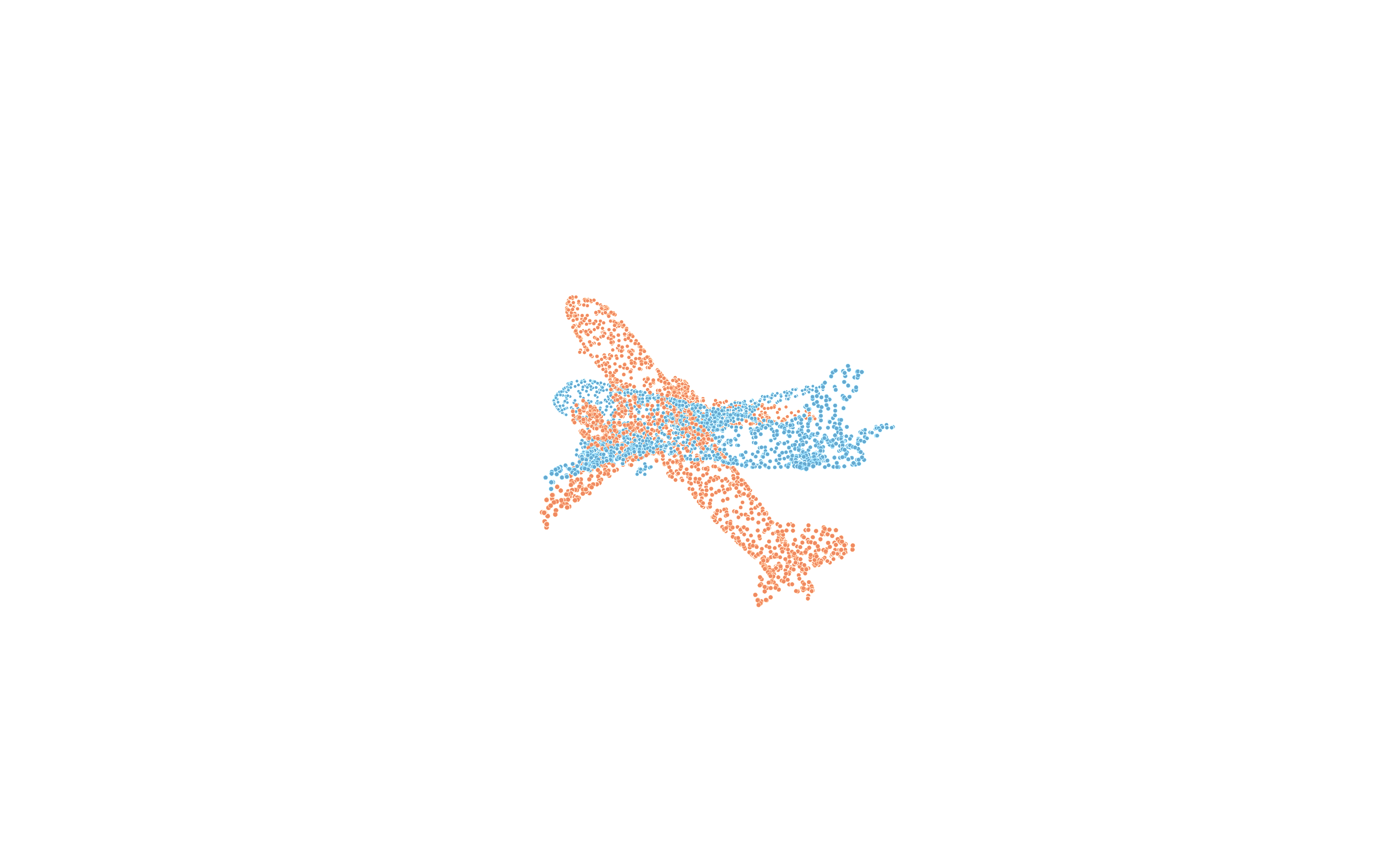}}
    \frame{\includegraphics[trim=12cm 7cm 8cm 5cm,clip,width=0.196\textwidth,height=0.14\textheight]{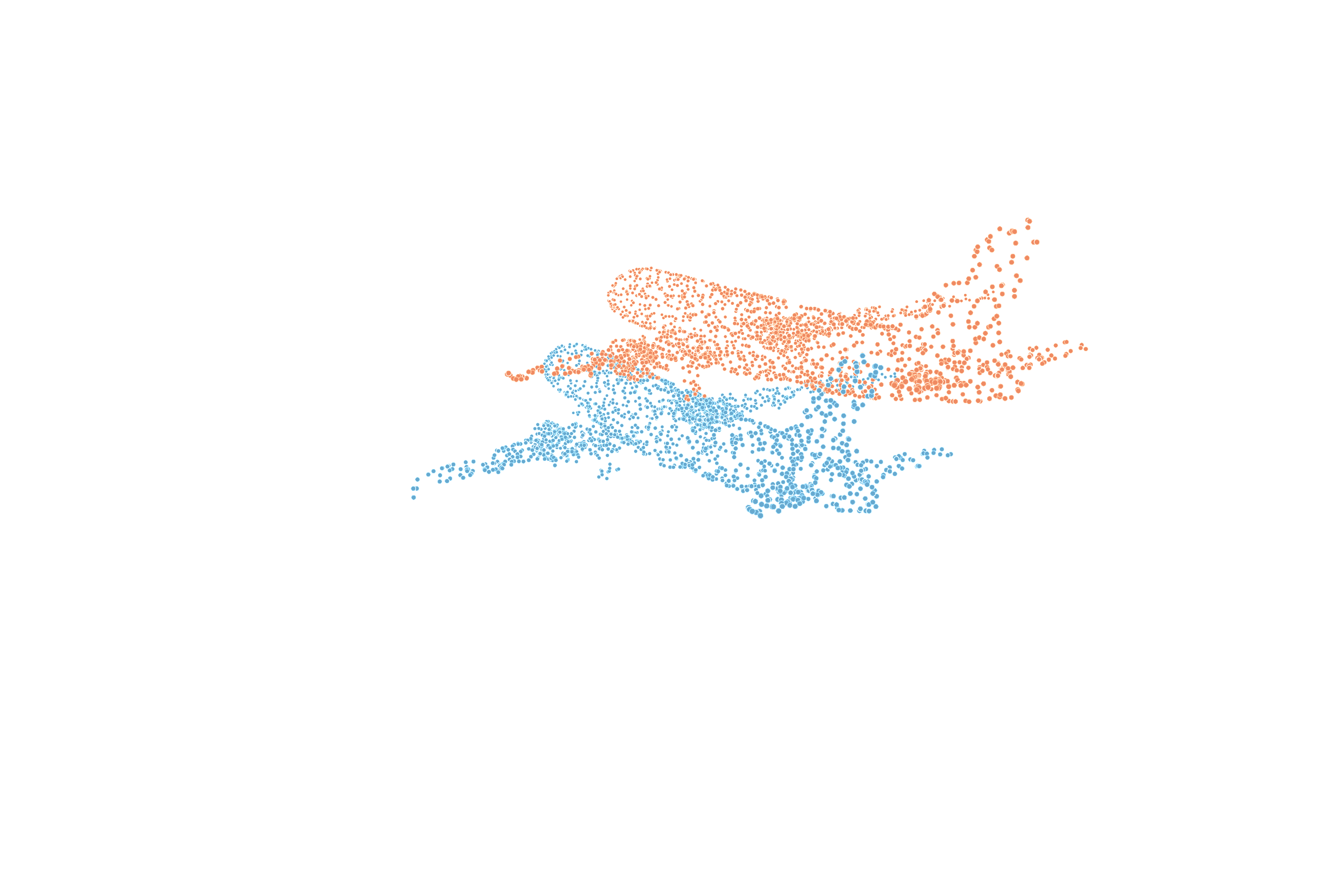}}
    \frame{\includegraphics[trim=10cm 5cm 10cm 5cm,clip,width=0.196\textwidth,height=0.14\textheight]{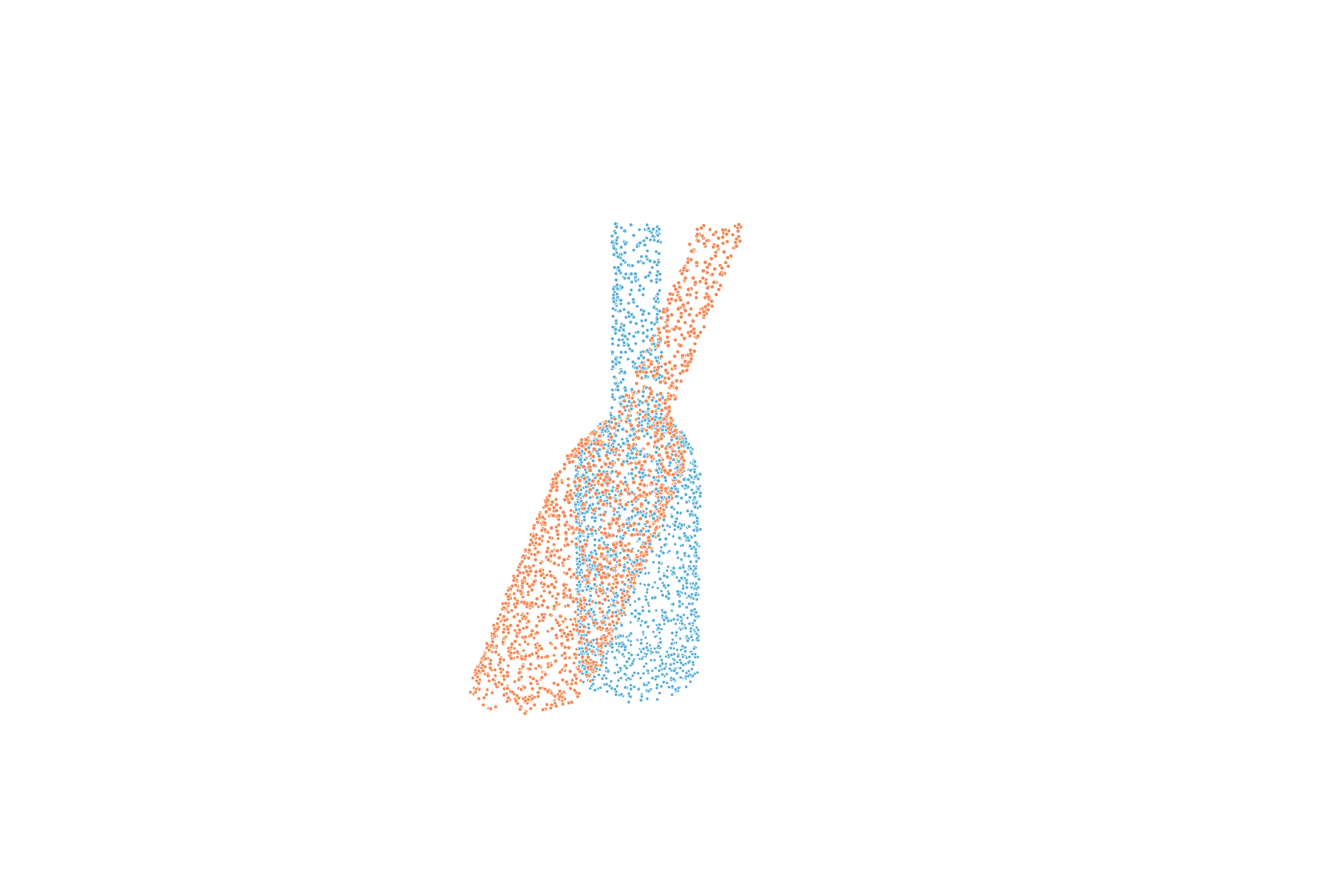}}
    \frame{\includegraphics[trim=10cm 5cm 10cm 6cm,clip,width=0.196\textwidth,height=0.14\textheight]{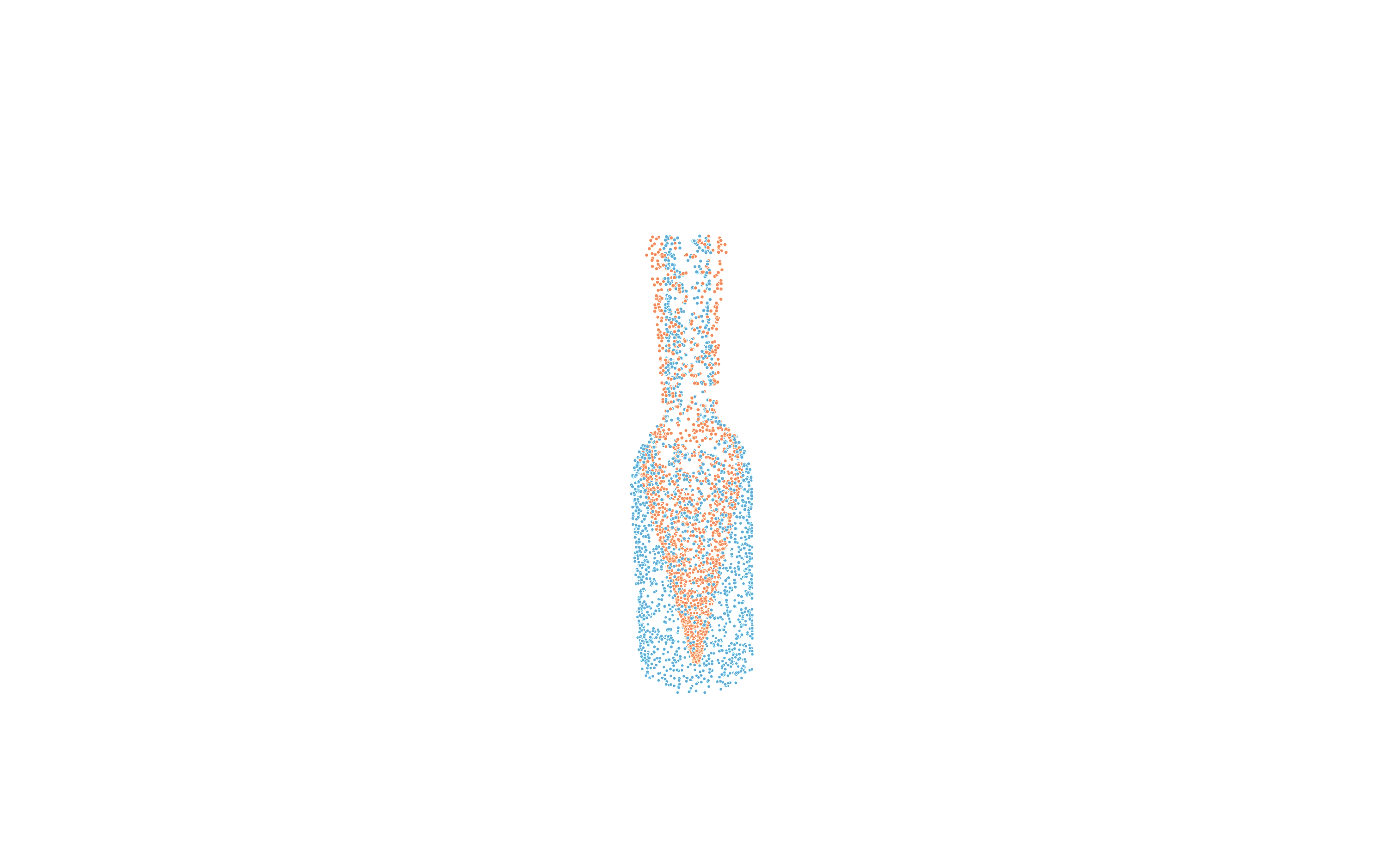}}
    \frame{\includegraphics[trim=11cm 6cm 11cm 6cm,clip,width=0.196\textwidth,height=0.14\textheight]{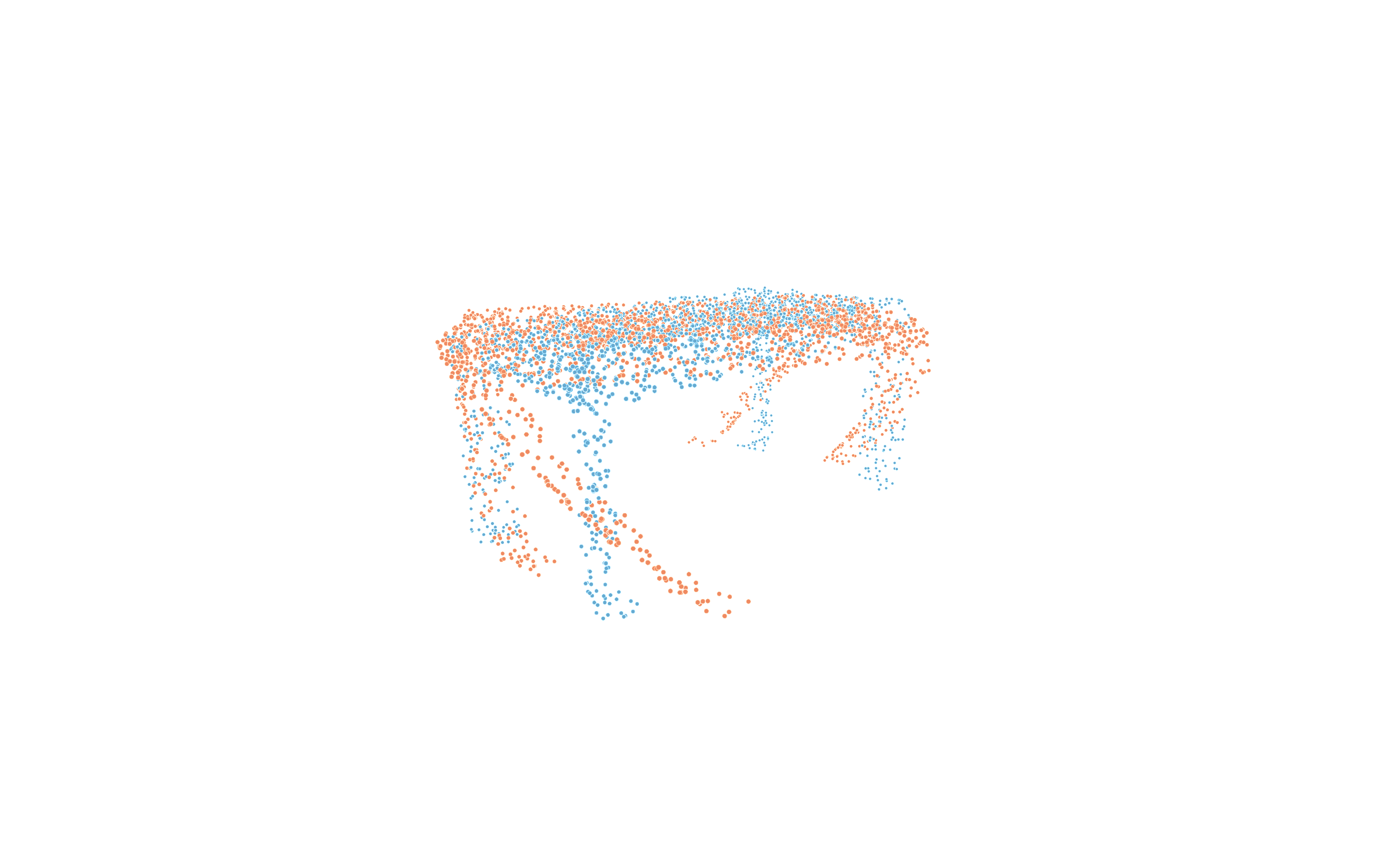}}\\
    \caption{\label{fig:deformationsExamples}
    \textbf{Visualization of Spatially-Deformed Point Clouds.} 
    In blue the original point cloud and in red its deformed version.
    \emph{From left to right:} Rotation (plane), Translation (plane), Shearing (bottle), Tapering (bottle) and Twisting (table).
    }
\end{figure*}

Perception of 3D scenes plays a critical role in applications such as autonomous navigation~\cite{favaro2017examining} and robotics~\cite{marescaux2001transatlantic}.
The success of autonomous driving cars and surgical robots largely depends on their ability to understand the surrounding scenes.
Recent works~\cite{PointNet,PointNet++,DGCNN,CurveNet} demonstrated the capability of Deep Neural Networks (DNNs) to process 3D point clouds.
These advances allowed DNNs to achieve exceptional performance in challenging 3D tasks, such as shape classification~\cite{ModelNet,ScanObjectNN}, object detection~\cite{KITTI} and semantic segmentation~\cite{S3DIS,ScanNet}.
Despite the superior performance of DNNs over traditional approaches, explaining the logic behind their decisions and understanding their limitations is extremely challenging.
For instance, earlier works~\cite{xiang2019generating,hamdi2020advpc} elucidated the limitations of point cloud DNNs to withstand tiny and meaningless perturbations in their input, known as adversarial attacks.
The vulnerability of point cloud DNNs against adversarial attacks underscores the importance of considering their robustness for security-critical applications.
 
The study of adversarial robustness in the image domain demonstrated that defenses~\cite{guo2018countering} are often later broken by more sophisticated attacks~\cite{carlini2017towards,athalye2018obfuscated,croce2020reliable}.
This phenomenon poses a difficulty for evaluating robustness~\cite{carlini2019evaluating}.
Such difficulty found response in the domain of certified robustness, wherein a defense's robustness is theoretically guaranteed and thus independent of the attack's sophistication.
Recent methods for certifying DNNs on images have spurred interest in extending such methods to point cloud DNNs.
In particular, the seminal work of Liu~\etal\cite{liu2021pointguard} certified point cloud DNNs against modification, addition, and deletion of points.
Furthermore, Lorenz~\etal~\cite{lorenz2021robustness} recently proposed a verification approach to certify against 3D transformations.

In this work, we propose 3DeformRS, a certification approach for point cloud DNNs against spatial and deformable transformations.
Figure~\ref{fig:PullingFigure} provides an overview of 3DeformRS's inner workings.
Our approach builds upon the theoretical background of Randomized Smoothing~(RS)~\cite{cohen2019certified}.
Specifically, we build 3DeformRS by leveraging DeformRS~\cite{deformrs}, an RS reformulation that generalized from pixel-intensity perturbations to vector field deformations, and specializing it to point cloud data.
In contrast to previous approaches, our work considers spatial deformations on any point cloud DNN, providing efficiency and practicality.
Compared to previous work, we find that 3DeformRS produces comparable-to-better certificates, while requiring significantly less computation time. 
We thus build on 3DeformRS' virtues and provide a comprehensive empirical study on the certified robustness of four representative point cloud DNNs (PointNet~\cite{PointNet}, PointNet++~\cite{PointNet++}, DGCNN~\cite{DGCNN}, and CurveNet~\cite{CurveNet}) on two datasets (ModelNet40~\cite{ModelNet} and ScanObjectNN~\cite{ScanObjectNN}), and against seven different types of spatial deformations (Rotation, Translation, Affine, Twisting, Tapering, Shearing, and Gaussian Noise).

\paragraph{Contributions}
In summary, our contributions are three-fold:
\textbf{(i)} We propose 3DeformRS by extending Randomized Smoothing (RS) from certifying image deformations to point cloud transformations. 
We further show that RS' classical formulation for certifying input perturbations can be seen as a special case of 3DeformRS. 
\textbf{(ii)} We conduct a comprehensive empirical study with 3DeformRS, where we assess the certified robustness of four point cloud DNNs, on two classification datasets and against seven spatial deformations. 
\textbf{(iii)} We compare 3DeformRS with an earlier point cloud certification approach in certification and runtime.
We show that 3DeformRS delivers consistent improvements while inheriting RS' scalability and efficiency.

\section{Related Work}\label{sec:related}
\paragraph{3D Computer Vision}
Images enjoy a canonical representation as fixed-sized matrices; in contrast, several representations exist for 3D data.
Such representations include point clouds~\cite{PointNet,PointNet++,CurveNet}, meshes~\cite{MeshNet,DualConvMeshNet,PicassoNet}, voxels~\cite{VoxelNet,OctNet,MinkowskiNet}, multi-view~\cite{MVCNN,ViewGCN,MVTN} and implicit representations~\cite{OccupancyNet,DeepSDF,NeRF}.
Given the prevalence and practicality of point clouds, we focus our attention on this representation and the associated DNNs that process it.
PointNet~\cite{PointNet} was the first successful attempt at using DNNs on point clouds.
This architecture introduced a point-wise MLP with a global set pooling, and achieved remarkable performance in shape classification and semantic segmentation.
PointNet++~\cite{PointNet++} then introduced intermediate pooling operations in point clouds for local neighborhood aggregation.
Afterwards, DGCNN~\cite{DGCNN} modeled convolutional operations in point clouds based on dynamically-generated graphs between closest point features.
Recently, CurveNet~\cite{CurveNet} learned point sequences for local aggregation, achieving state-of-the-art performance on 3D computer vision tasks.
In this work, we conduct a comprehensive empirical study on certified robustness by analyzing the robustness of these four point cloud DNNs.

\paragraph{Robustness}
Szegedy~\etal's~\cite{szegedy2014intriguing} seminal work exposed the vulnerability of DNNs against small input modifications, now known as adversarial examples.
Later works observed the pervasiveness of this phenomenon~\cite{goodfellow2015explaining,carlini2017towards}, spurring an arms race between defenses that enhanced DNNs' adversarial robustness~\cite{madry2018towards,Xie_2019_CVPR,guo2018countering} and attacks that could break such defenses~\cite{athalye2018obfuscated,carlini2017towards}.
The conflict between ever-more complex defenses and attacks also incited interest towards ``certified robustness''~\cite{wong2018provable}, wherein defenses enjoy theoretical guarantees about the inexistence of adversarial examples that could fool them.
A set of works focused on exact verification~\cite{gowal2018effectiveness,zhang2020towards}, while others considered probabilistic certification~\cite{lecuyer2019certified,li2018second}.
Randomized Smoothing~(RS)~\cite{cohen2019certified} has emerged as a certification approach from the probabilistic paradigm that scales well with models and datasets.
Notably, RS has been successfully combined with adversarial training~\cite{salman2019provably}, regularization~\cite{zhai2019macer}, and smoothing-parameters' optimization~\cite{DBLP:journals/corr/abs-2107-04570,alfarra2020data}.
Recently, DeformRS~\cite{deformrs} reformulated RS to consider general parameterized vector-field deformations.
In this work, we develop 3DeformRS by specializing and extending DeformRS to certify point cloud DNNs against spatial deformations.

\paragraph{Certification on 3D Point Clouds}
Since the seminal work of Xiang~\etal~\cite{xiang2019generating} attacked point cloud DNNs, several works studied the robustness of such DNNs~\cite{xiang2019generating,tsai2020robust,zhou2019dup,hamdi2020advpc,ma2020efficient,kim2021minimal,ma2021towards,sun2021adversarially}.
Despite growing interest in the robustness of point cloud DNNs, only two works have addressed their certification.
PointGuard~\cite{liu2021pointguard} provided tight robustness guarantees against modification, addition, and deletion of points.
3DCertify~\cite{lorenz2021robustness} generalized DeepG~\cite{balunovic2019certifying} to 3D point clouds and proposed 3DCertify, a verification approach to certify robustness against common 3D transformations.
PointGuard has the benefit of low computational cost~(compared to that of exact verification), but does not allow for spatial deformations. 
Conversely, 3DCertify considers such transformations, but suffers from impractical computational costs.
In contrast, our 3DeformRS approach combines the best of both worlds, thus allowing for spatial deformations while enjoying low computational cost.

\section{Our approach: 3DeformRS}
\begin{table*}[t]
    \centering
    \resizebox{\linewidth}{!}{
    \begin{tabular}{c|l|c||c|l|c||c|l|c}
    \toprule
        Name & Flow ($\tilde p$) & $\phi$ &
        Name &  Flow ($\tilde p$) & $\phi$ & Name & Flow ($\tilde p$) & $\phi$  \\
        \midrule
        Translation
        & $\begin{array}{lcl}
           \Tilde{x} = t_x  \\
           \Tilde{y} = t_y \\
           \Tilde{z} = t_z
        \end{array}$
        & $\begin{bmatrix} t_x \\ t_y \\ t_z \end{bmatrix}$
        &
        \begin{tabular}{@{}c@{}}$z-$ \\ Rotation \end{tabular}
        & $\begin{array}{lcl}
          \Tilde{x} = (c_\gamma-1) x - s_\gamma y \\
          \Tilde{y} = s_\gamma x + (c_\gamma-1) y \\
          \Tilde{z} = 0  
        \end{array}$
        & $\begin{bmatrix} \gamma \end{bmatrix}$
        &
        Affine
        & $\begin{array}{lcl}
           \Tilde{x} = ax + by + cz + d \\
           \Tilde{y} = ex + fy + gz + h \\
           \Tilde{z} = ix + jy + kz + l
        \end{array}$
        &  $\begin{bmatrix} a \\ \vdots \\ l \end{bmatrix}$ \\
        
        \midrule
        
        \begin{tabular}{@{}c@{}}$z-$ \\ Shearing \end{tabular}
        & $\begin{array}{lcl}
           \Tilde{x} = az  \\
           \Tilde{y} = bz \\
           \Tilde{z} = 0
        \end{array}$
        & $\begin{bmatrix} a \\ b \end{bmatrix}$
        &
        \begin{tabular}{@{}c@{}}$z-$ \\ Twisting \end{tabular}
        & $\begin{array}{lcl}
           \Tilde{x} = (c_{\gamma z}-1) x - s_{\gamma z} y \\
           \Tilde{y} = s_{\gamma z} x + (c_{\gamma z}-1) y \\
           \Tilde{z} = 0  
        \end{array}$
        & $\begin{bmatrix} \gamma \end{bmatrix}$ 
        &
        \begin{tabular}{@{}c@{}}$z-$ \\ Tapering \end{tabular}
        & $\begin{array}{lcl}
          \Tilde{x} = (\frac{1}{2}a^2+b)zx \\
          \Tilde{y} = (\frac{1}{2}a^2+b)zy \\
          \Tilde{z} = 0
        \end{array}$
        & $\begin{bmatrix} a \\ b \end{bmatrix}$ 
        \\
        \bottomrule
    \end{tabular}
    }
    \caption{\label{tab:deformations}
    \textbf{Per-point Deformation Flows} for semantically-viable spatial deformations.
    Convention: $c_\alpha = \cos(\alpha)$ and $s_\alpha = \sin(\alpha)$.
    Without loss of generality, we only show the rotation around the $z-$axis, and leave the formulation of other rotations to the \textbf{Appendix}.
    }
\end{table*}
We present 3DeformRS, a probabilistic certification for point cloud DNNs against spatial deformations.

\paragraph{Preliminaries}
Our methodology builds on Randomized Smoothing~(RS) \cite{cohen2019certified}, arguably the most scalable DNN-certification method. 
Given a classifier $f: \mathbb R^d \rightarrow \mathcal{P(Y)}$ that maps an input $x\in\mathbb R^d$ to the probability simplex $\mathcal{P(Y)}$, RS constructs a smooth classifier $g$ that outputs the most probable class when $f$'s input is subjected to additive perturbations sampled from a distribution $\mathcal{D}$. 
While RS certified against additive pixel perturbations in images, DeformRS~\cite{deformrs} extended this formulation to certify against image \emph{deformations} by proposing a \emph{parametric-domain smooth classifier}.
Specifically, given image $x$ with coordinates $p$, a parametric deformation function $\nu_\phi$ with parameters $\phi$~(\eg if $\nu$ is a rotation, then $\phi$ is the rotation angle), and an interpolation function $I_T$, DeformRS defined a parametric-domain smooth classifier
\begin{equation}\label{eq:parametric-smooth-classifier}
g_\phi(x, p) = \mathbb E_{\epsilon\sim \mathcal D}\left[ f(I_T(x, p+\nu_{\phi+\epsilon})) \right].   
\end{equation}
In a nutshell, $g$ outputs $f$'s average prediction over transformed versions of $x$. 
Note that, in contrast to RS, this formulation deforms pixel \emph{locations} instead of their intensity.
DeformRS showed that parametric-domain smooth classifiers are certifiably robust against perturbations to the deformation function's parameters via the following corollary.

\begin{corollary}\label{cor:parametric-certification}~(restated from~\cite{deformrs}). 
Suppose $g$ assigns class $c_A$ to an input $x$, \ie $c_A = \arg\max_i g^i_\phi(x, p)$ with
\[
p_A = g_\phi^{c_A}(x, p) \,\, \text{and} \,\, p_B = \max_{c\neq c_A}g_\phi^c(x, p)
\]
Then $\arg\max_c g^c_{\phi+\delta}(x, p) = c_A$ for all parametric perturbations $\delta$ satisfying:
\begin{align*}
    &\|\delta\|_1 \leq \lambda \left(p_A - p_B\right) &\text{for } \mathcal D = \mathcal{U}[-\lambda, \lambda], \\
    &\|\delta\|_2 \leq \frac{\sigma}{2}\left(\Phi^{-1}(p_A) - \Phi^{-1}(p_B)\right) &\text{for } \mathcal D = \mathcal N(0, \sigma^2I).
\end{align*}
\end{corollary}
In short, Corollary~\eqref{cor:parametric-certification} states that, as long as the parameters of the deformation function (\eg rotation angle) are perturbed by a quantity upper bounded by the certified radius, $g_\phi$'s prediction will remain constant.
This result allowed certification against various image deformations. 
In this work, we build upon DeformRS and specialize it to certify against spatial deformations in point clouds.

\subsection{Parametric Certification for 3D DNNs}
We specialize the result in Corollary~\eqref{cor:parametric-certification} to point clouds. 
In this setup, $p \in \mathbb R^{N \times 3}$ is a point cloud consisting of $N$ 3-dimensional points. 
We highlight two key differences between certifying images and point clouds. 
\textbf{(i)}~The interpolation function $I_T$, while essential in images for the pixels' discrete locations, is irrelevant for 3D coordinates and so can be omitted. 
\textbf{(ii)}~Most recent DNNs neglect the points' color information and \emph{exclusively} rely on the points' location.
We combine these observations and modify the parametric domain smooth classifier from Eq.~\eqref{eq:parametric-smooth-classifier} to propose
\begin{equation}\label{eq:3d-domain-smooth-classifier}
    \hat g_\phi(p) = \mathbb E_{\epsilon\sim \mathcal D}\left[ f(p+\nu_{\phi+\epsilon}) \right].
\end{equation}
Our $\hat g_\phi$, inheriting $g_\phi$'s structure, is certifiable against parametric perturbations via Corollary~\eqref{cor:parametric-certification}. 
Note that, since point cloud DNNs exclusively rely on location, our \emph{parametric} certification is thus equivalent to \emph{input} certification. 

To consider input deformations, let us study the general case where $\nu_\phi = \phi \in \mathbb R^{N\times 3}$. 
Under this setup, and by setting $p' = p + \phi$, our smooth classifier becomes:
\begin{equation}\label{eq:3d-vf-smooth-classifier}
    \Tilde g_\phi(p) = \mathbb E_{\epsilon\sim \mathcal D}\left[ f(p'+\epsilon) \right].
\end{equation}
The classifier in Eq.~\eqref{eq:3d-vf-smooth-classifier} is a general case of both the domain smooth classifier~\cite{deformrs} and the input smooth classifier proposed earlier in~\cite{cohen2019certified}. 
Note that this expression elucidates how parametric certification of point cloud DNNs is equivalent to input certification. 
At last, we note that the smooth classifier in Eq.~\eqref{eq:3d-vf-smooth-classifier} is also certifiable via Corollary~\ref{cor:parametric-certification}.
We highlight here that directly certifying $\tilde g$ against parametric transformations (\eg rotation) will perform poorly, as empirically observed in~\cite{liu2021pointguard}. 
However, such deficient performance is not an inherent weakness of RS for certifying point cloud transformations, but rather due to a sub-optimal formulation of RS for spatial deformations. 
We argue that our formulation, \ie the parametric domain smooth classifier in Eq.~\eqref{eq:3d-domain-smooth-classifier}, is more suitable for modeling spatial deformations than the one presented in~\cite{liu2021pointguard}.

Next, we outline various spatial deformations we consider for assessing the robustness point cloud DNNs.

\subsection{Modeling Spatial Deformations}
We now detail the spatial deformations we consider to assess the robustness of point cloud DNNs.
Our formulation from Eq.~\eqref{eq:3d-vf-smooth-classifier} requires modeling spatial deformations as additive perturbations on the points' coordinates.
Thus, given a point cloud $p$ that is transformed to yield $p'$, we define the flow that additively perturbs $p$ as $\tilde p = p' - p$, where $\tilde p \in \mathbb R^{N \times 3}$.
Modeling transformations is then equivalent to modeling the induced per-point flow.
Thus, we are required to model each deformation via a \emph{parametric flow}, whose parameters are those of the corresponding transformation.

We consider six parameterizable flows, corresponding to four linear and two nonlinear transformations.
In particular, we consider four linear, rigid and deformable transformations: (1)~rotation, (2)~translation, (3)~shearing, and (4)~the general affine transformation.
Furthermore, we follow prior work~\cite{lorenz2021robustness} and consider two nonlinear transformations: (5) tapering and (6)~twisting.
As a summary, we report all the deformation flows and their corresponding parameters in Table~\ref{tab:deformations}, and visualize some of their effects in Figure~\ref{fig:deformationsExamples}.

Note that the affine transformation is the most general transformation we consider, capable of modeling any linear transformation.
Presumably, a DNN enjoying certified robustness against affine transformations would also enjoy robustness against combinations of the other spatial deformations.
We leave formulations of the aforementioned transformations in homogeneous coordinates to the \textbf{Appendix}.

\paragraph{Gaussian Noise}
In addition to the above transformations, we also consider Gaussian noise perturbations.
Notably, this noise deforms the underlying vector field and, thus, is the most general perturbation. 
To certify against Gaussian noise, we construct the smooth classifier in Eq.~\eqref{eq:3d-vf-smooth-classifier} with $\epsilon\in\mathbb{R}^{N\times 3}$ and $\mathcal D = \mathcal N(0, \sigma^2 I)$. 
While this deformation is very general, the high dimensionality of the certified parameters limits its applicability to imperceptible perturbations. 
Nevertheless, as adversaries may take such form~\cite{xiang2019generating}, we consider this perturbation in our experiments.

\paragraph{Design Choices}
For each spatial deformation and point cloud DNN, we assess certified robustness by constructing the smooth classifier from~Eq.~\eqref{eq:3d-domain-smooth-classifier}.
In deformations whose parameter space is bounded (\eg~rotations, where angles beyond~$\pm \pi$ radians are redundant), we smooth with a uniform distribution and thus obtain an $\ell_1$ certificate.
For the remaining deformations, we employ Gaussian smoothing and thus obtain an $\ell_2$ certificate. 

\begin{figure*}
    \centering
    \includegraphics[trim={1cm 1cm 1cm 0.5cm},width=.5\linewidth,height=0.3cm,clip]{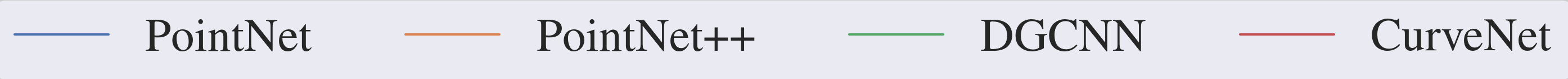}\\
    \includegraphics[trim={0cm 0cm 0cm 0cm},width=.22\linewidth,height=3.07cm,clip]{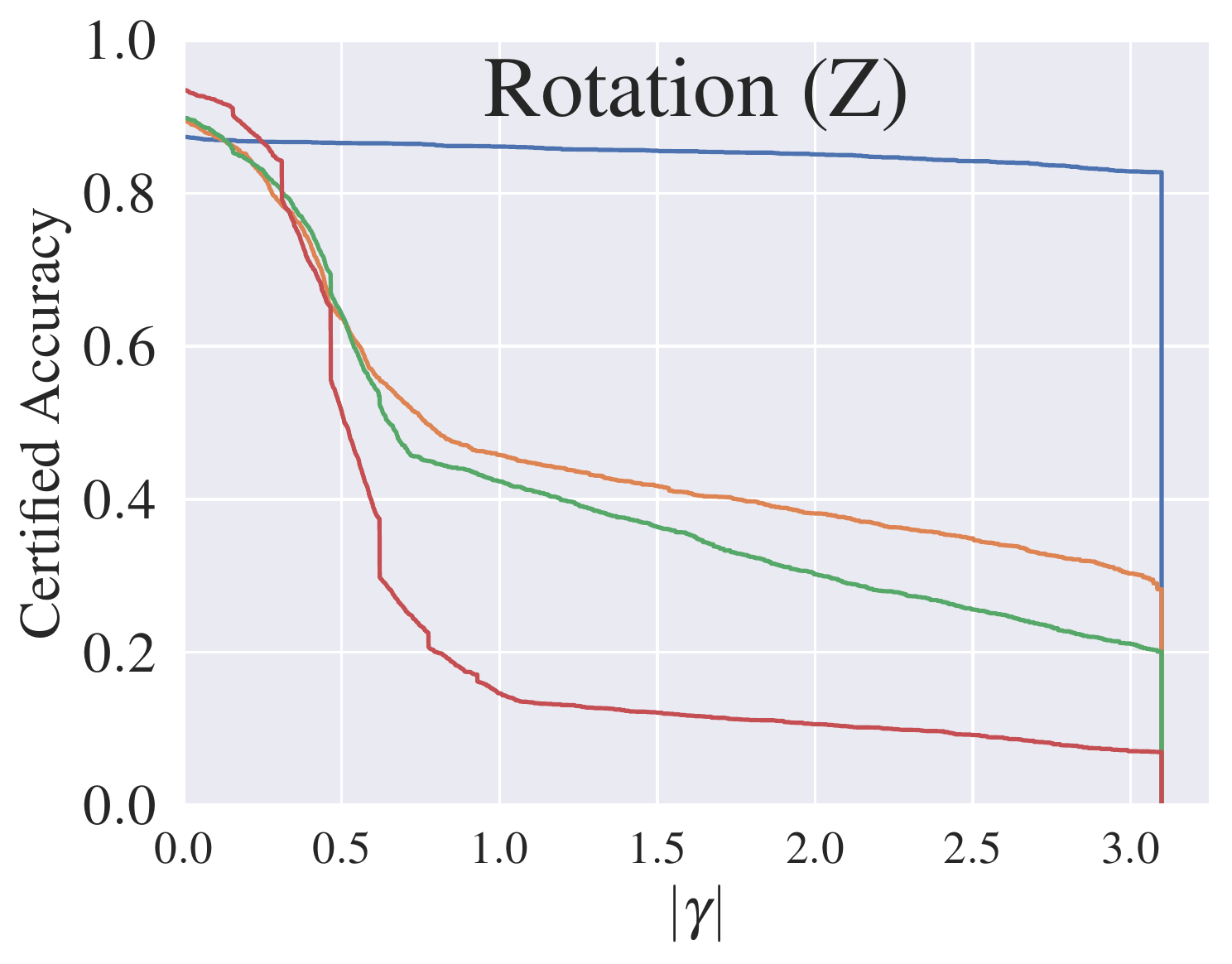}
    \includegraphics[trim={0cm 0cm 0cm 0cm},width=.19\linewidth,height=3cm,clip]{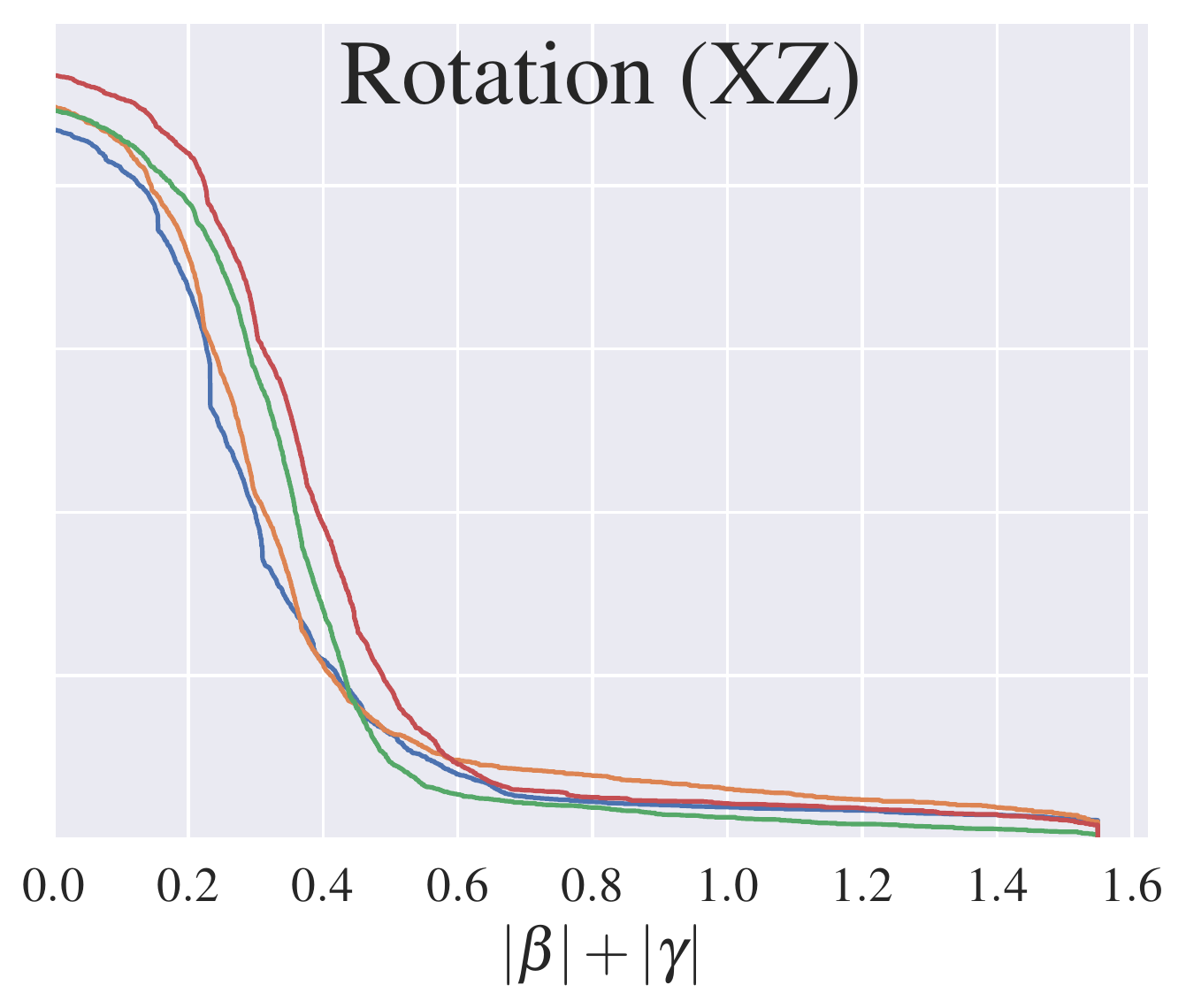} 
    \includegraphics[trim={0cm 0cm 0cm 0cm},width=.19\linewidth,height=3cm,clip]{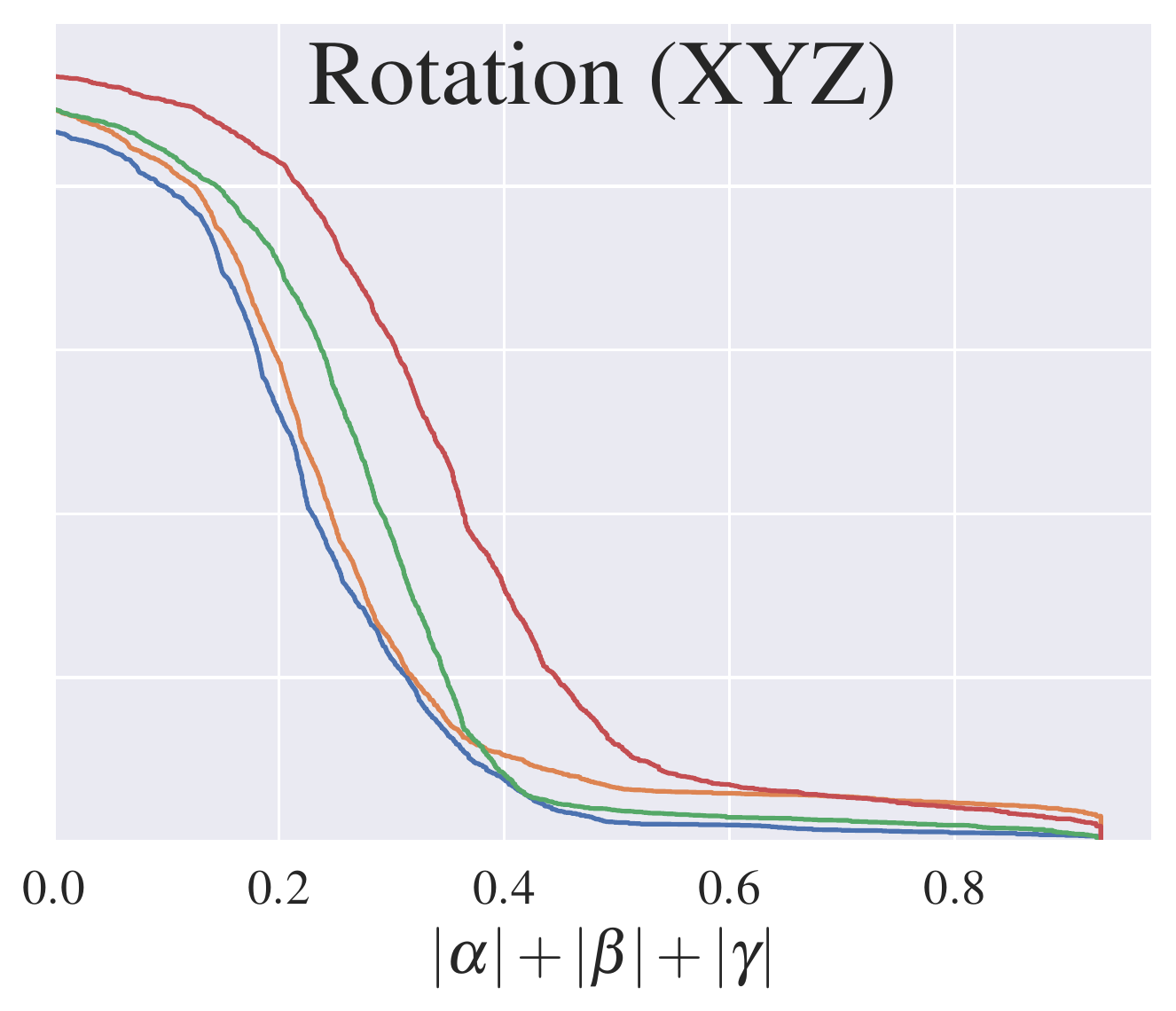}
    \includegraphics[trim={0cm 0cm 0cm 0cm},width=.19\linewidth,height=3cm,clip]{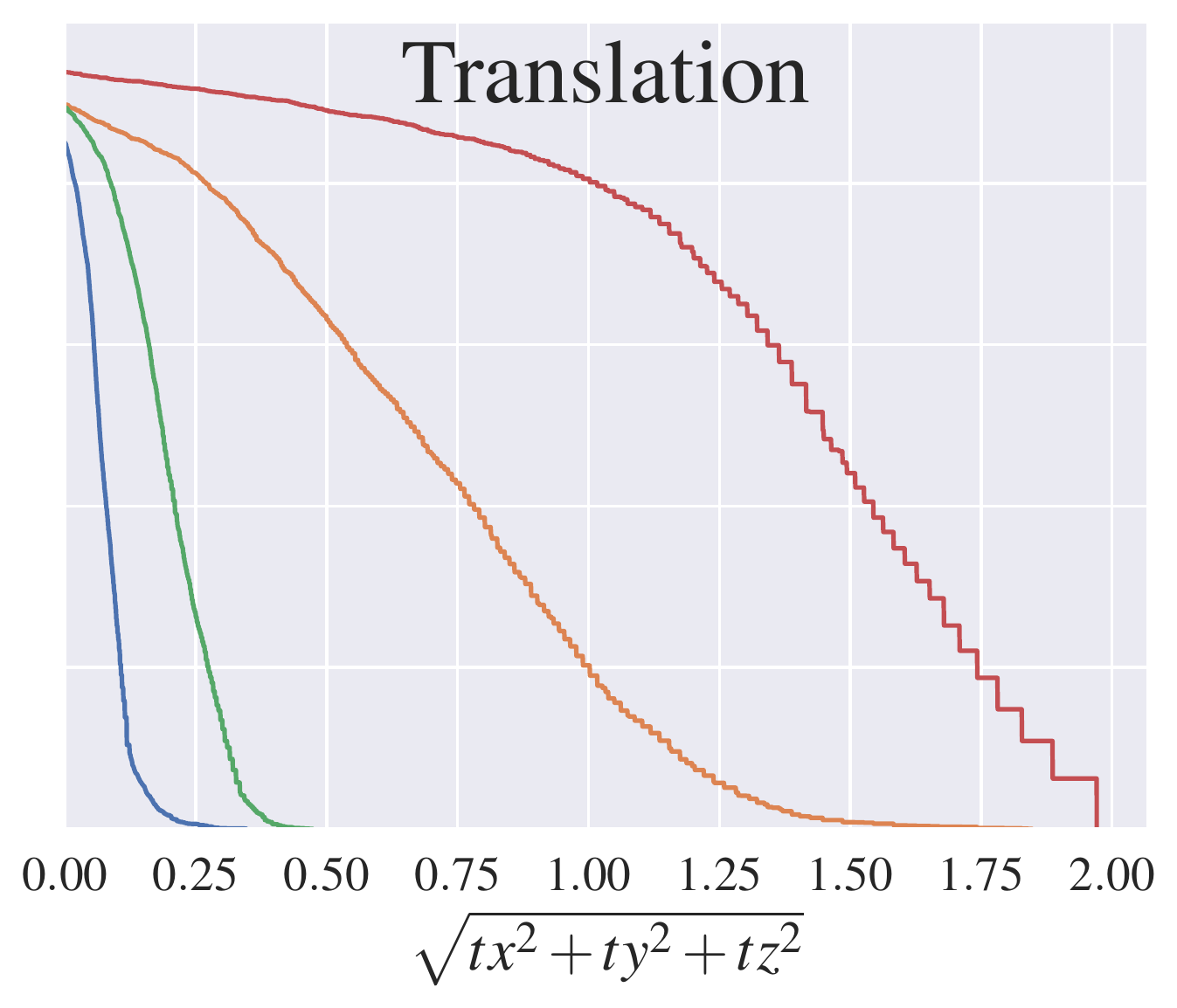}
    \includegraphics[trim={0cm 0cm 0cm 0cm},width=.19\linewidth,height=3cm,clip]{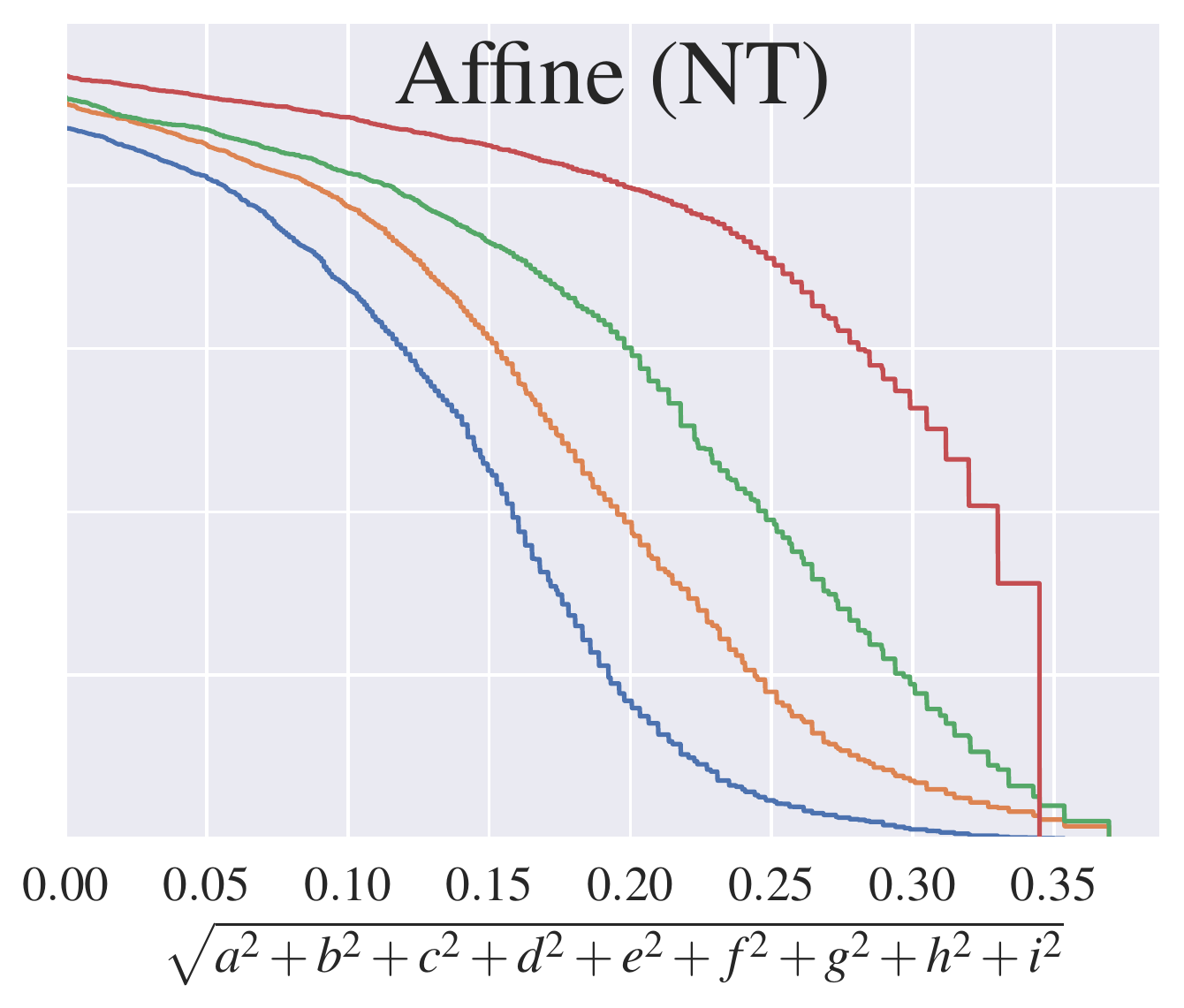}
    \includegraphics[trim={0cm 0cm 0cm 0cm},width=.22\linewidth,height=3.07cm,clip]{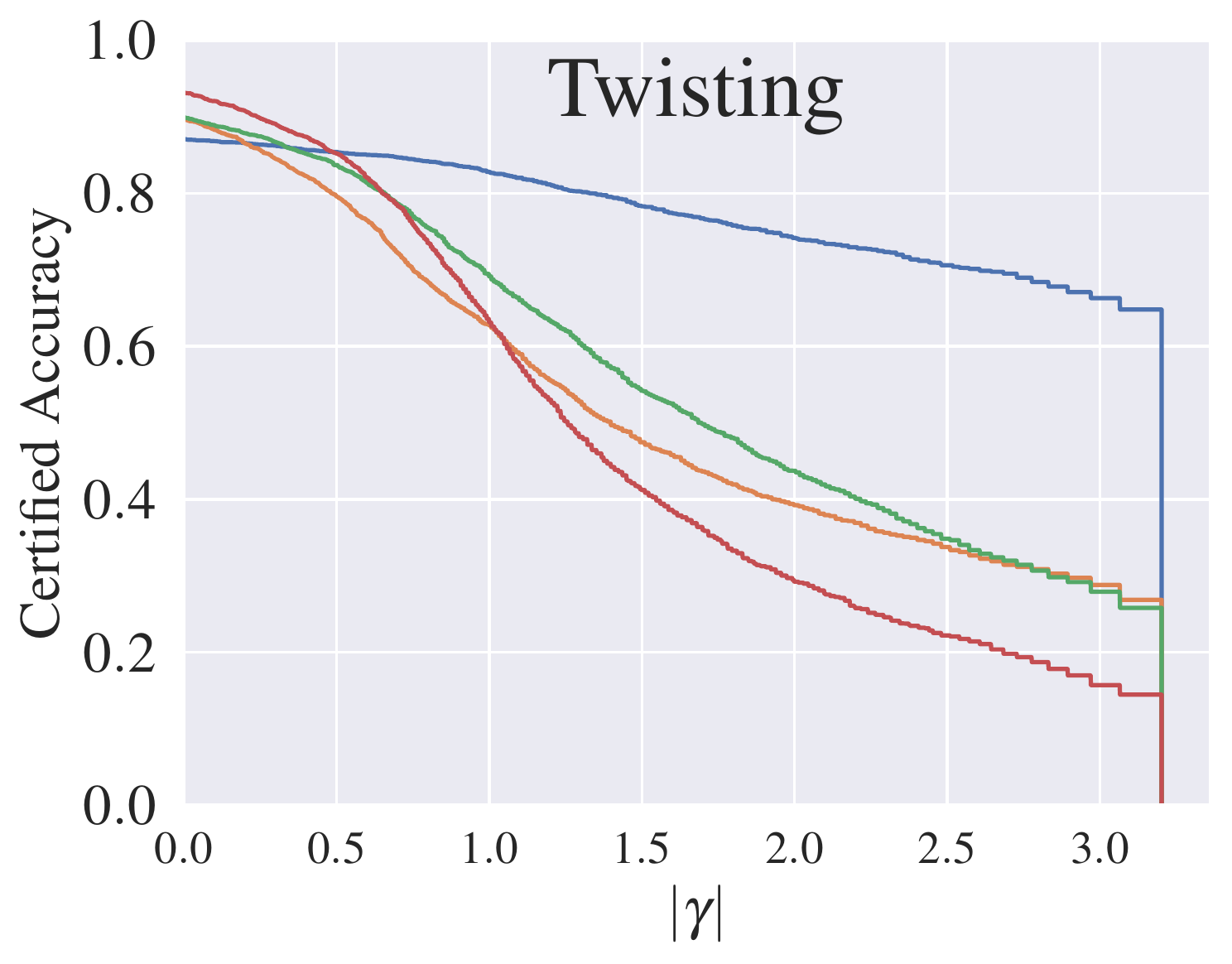}
    \includegraphics[trim={0cm 0cm 0cm 0cm},width=.19\linewidth,height=3cm,clip]{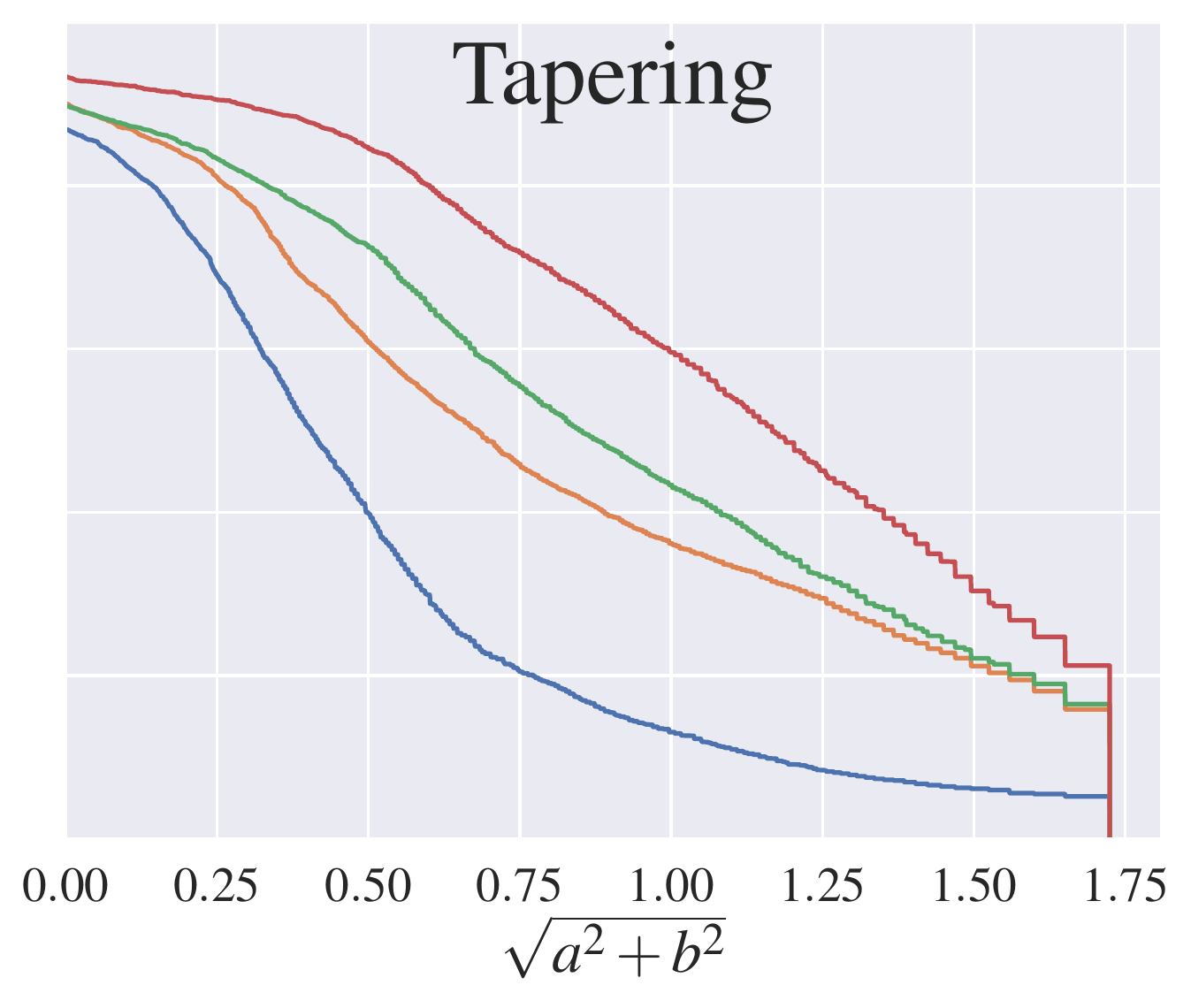}
    \includegraphics[trim={0cm 0cm 0cm 0cm},width=.19\linewidth,height=3cm,clip]{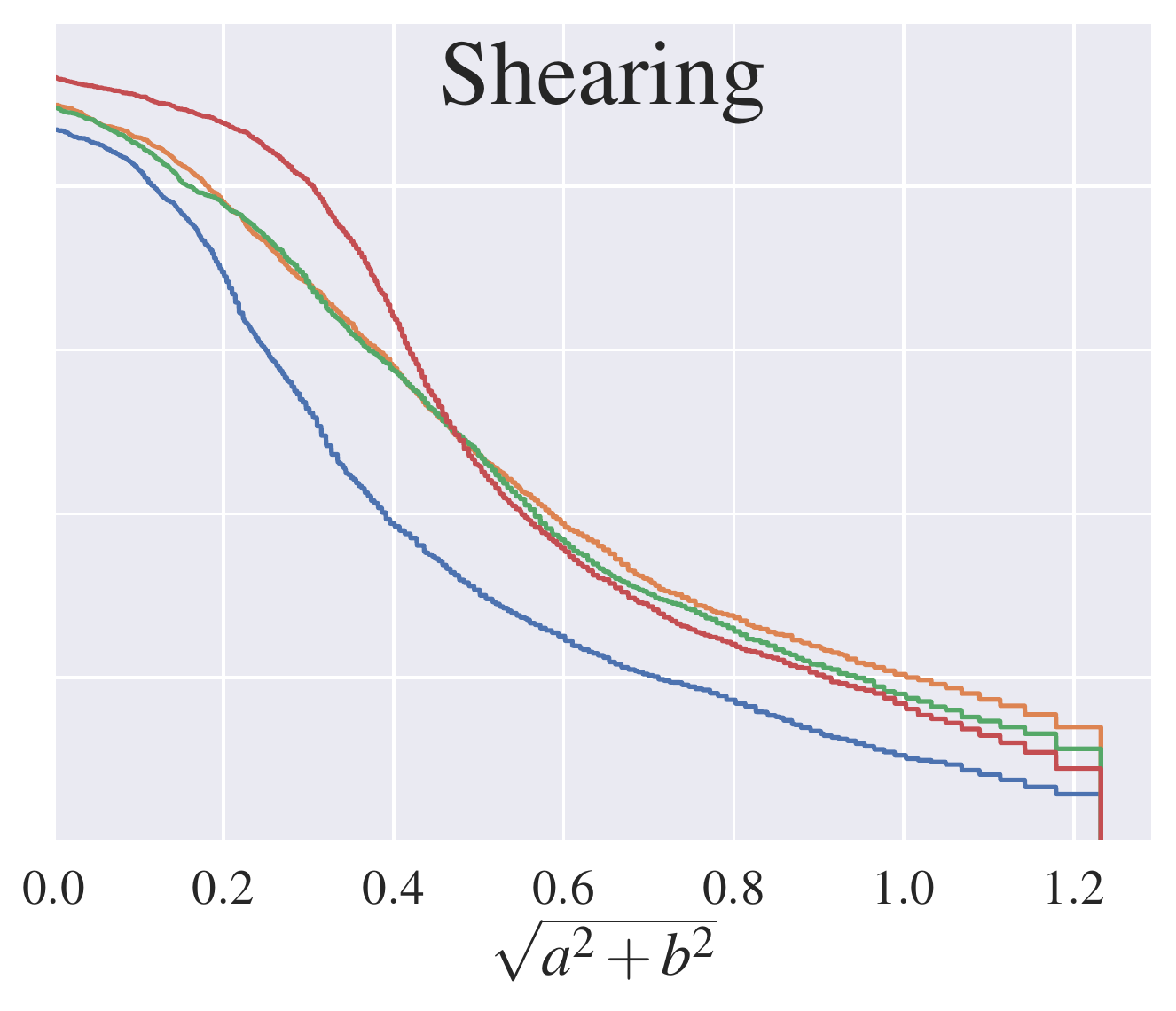}
    \includegraphics[trim={0cm 0cm 0cm 0cm},width=.19\linewidth,height=3cm,clip]{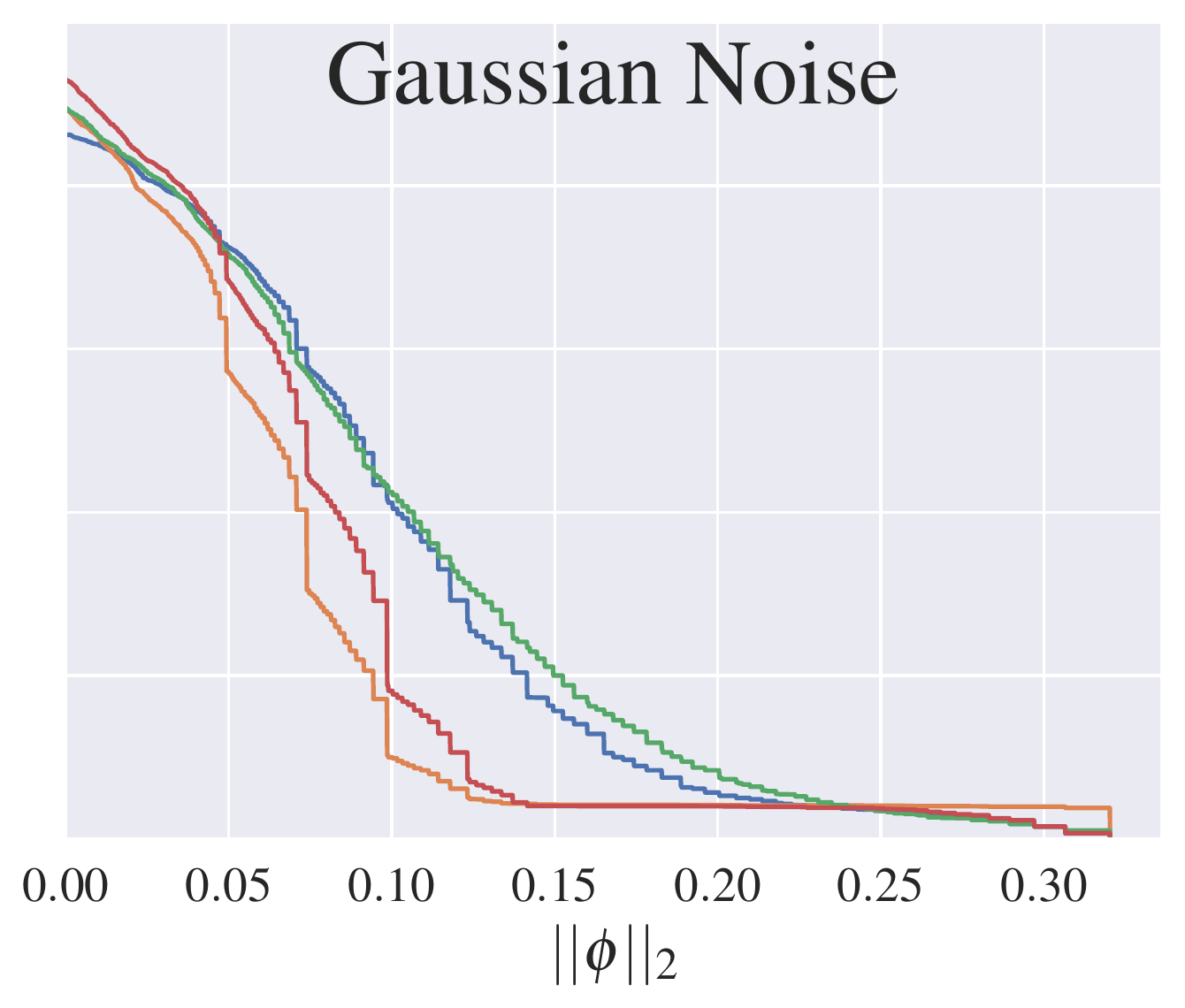}
    \includegraphics[trim={0cm 0cm 0cm 0cm},width=.19\linewidth,height=3cm,clip]{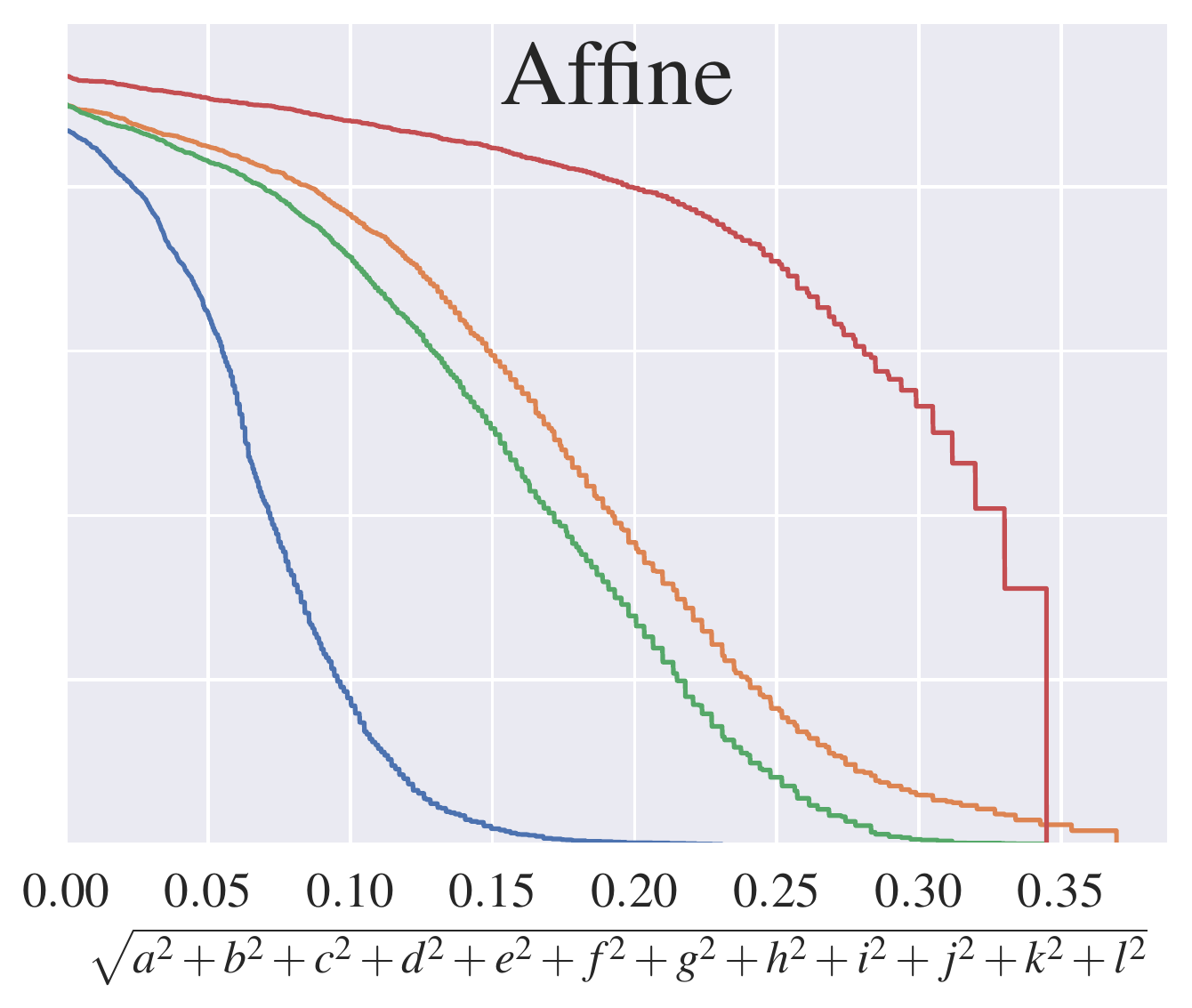}
    \caption{\label{fig:MainCurveModelNet40}
    \textbf{3DeformRS Certification on ModelNet40} of four point cloud DNNs against 10 transformations.
    }
\end{figure*}

\begin{figure*}
    \centering
    \includegraphics[trim={1cm 1cm 1cm 0.5cm},width=.5\linewidth,height=0.3cm,clip]{images/horizontalLegend.png}\\
    \includegraphics[trim={0cm 0cm 0cm 0cm},width=.22\linewidth,height=3.07cm,clip]{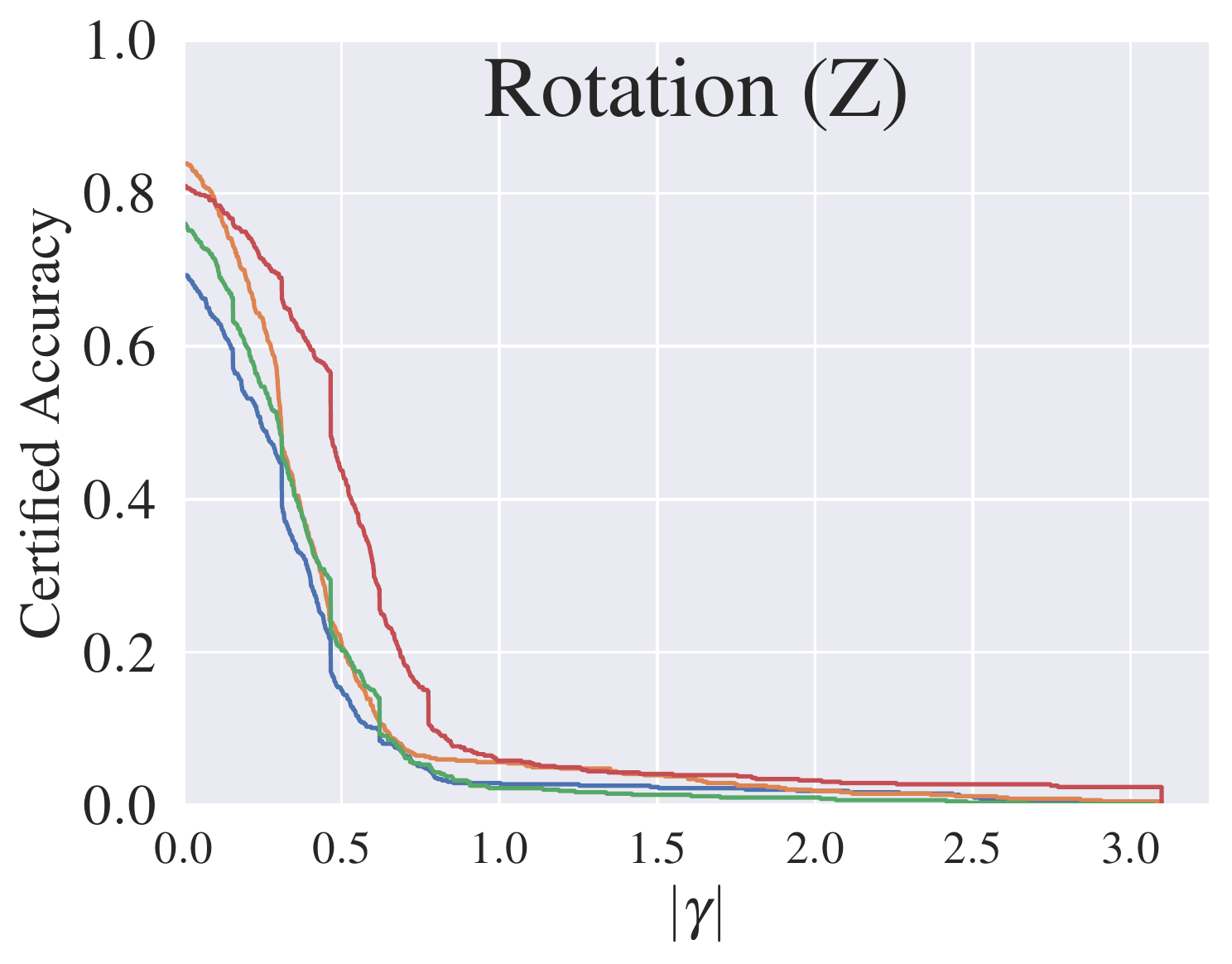}
    \includegraphics[trim={0cm 0cm 0cm 0cm},width=.19\linewidth,height=3cm,clip]{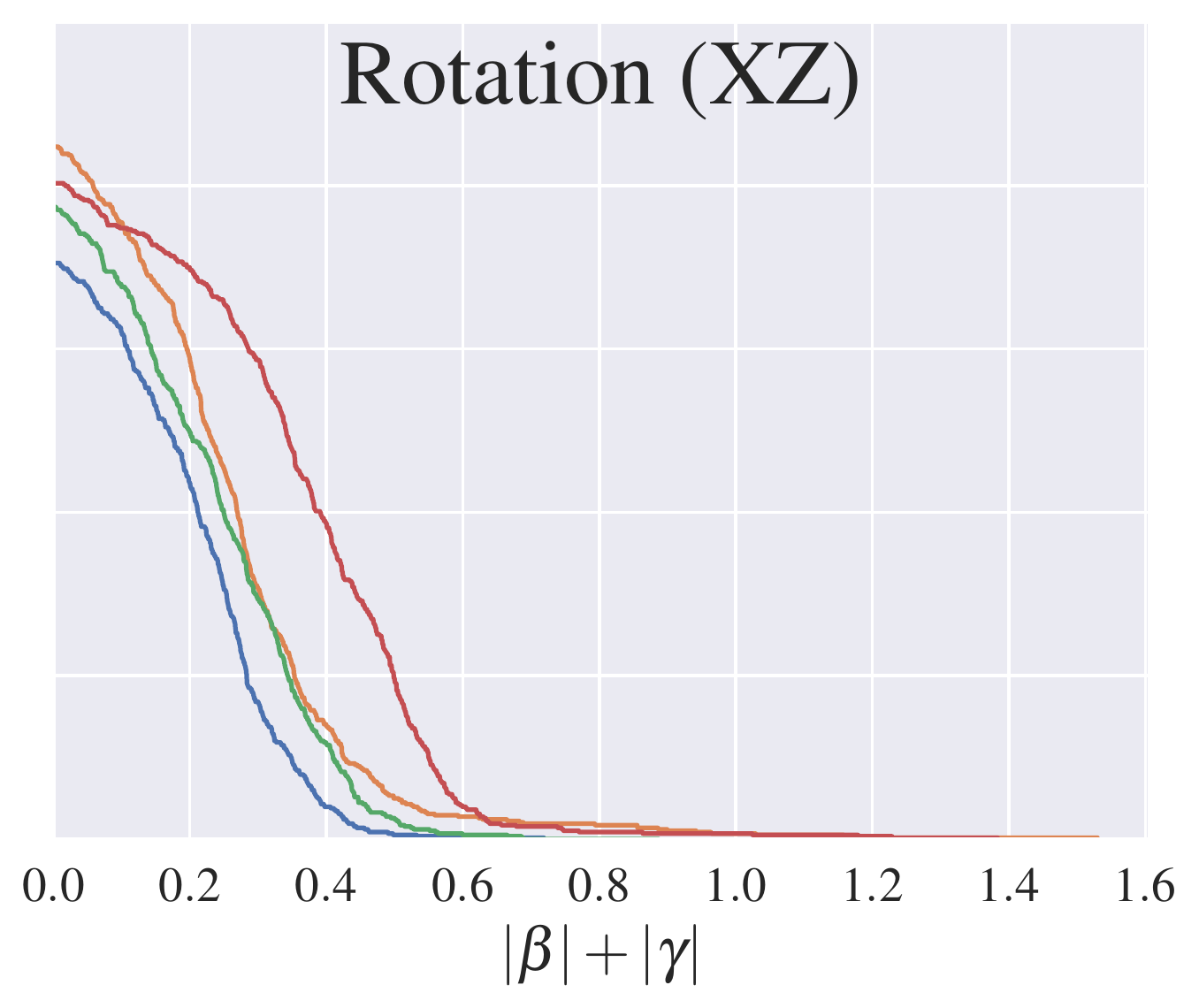} 
    \includegraphics[trim={0cm 0cm 0cm 0cm},width=.19\linewidth,height=3cm,clip]{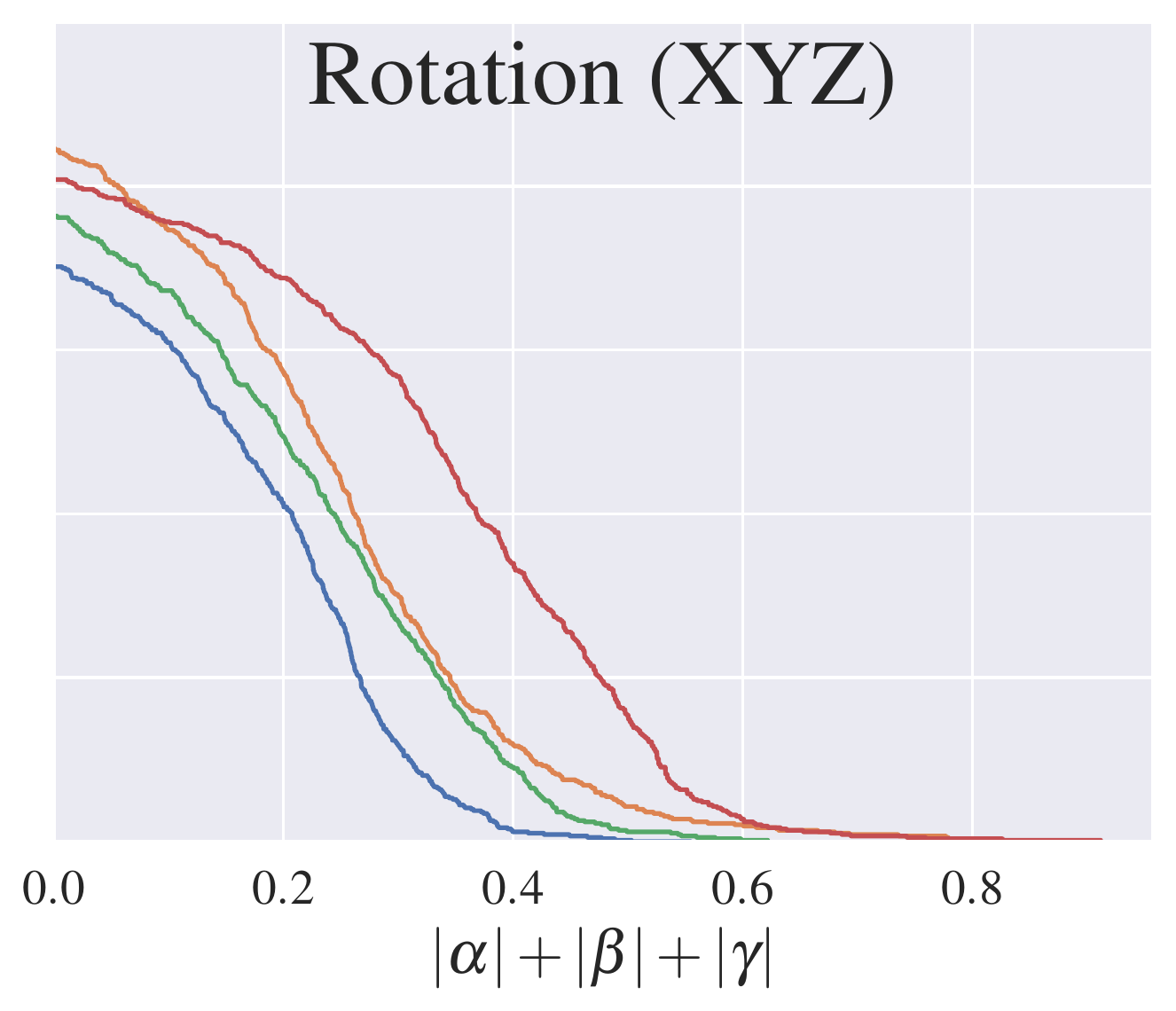}
    \includegraphics[trim={0cm 0cm 0cm 0cm},width=.19\linewidth,height=3cm,clip]{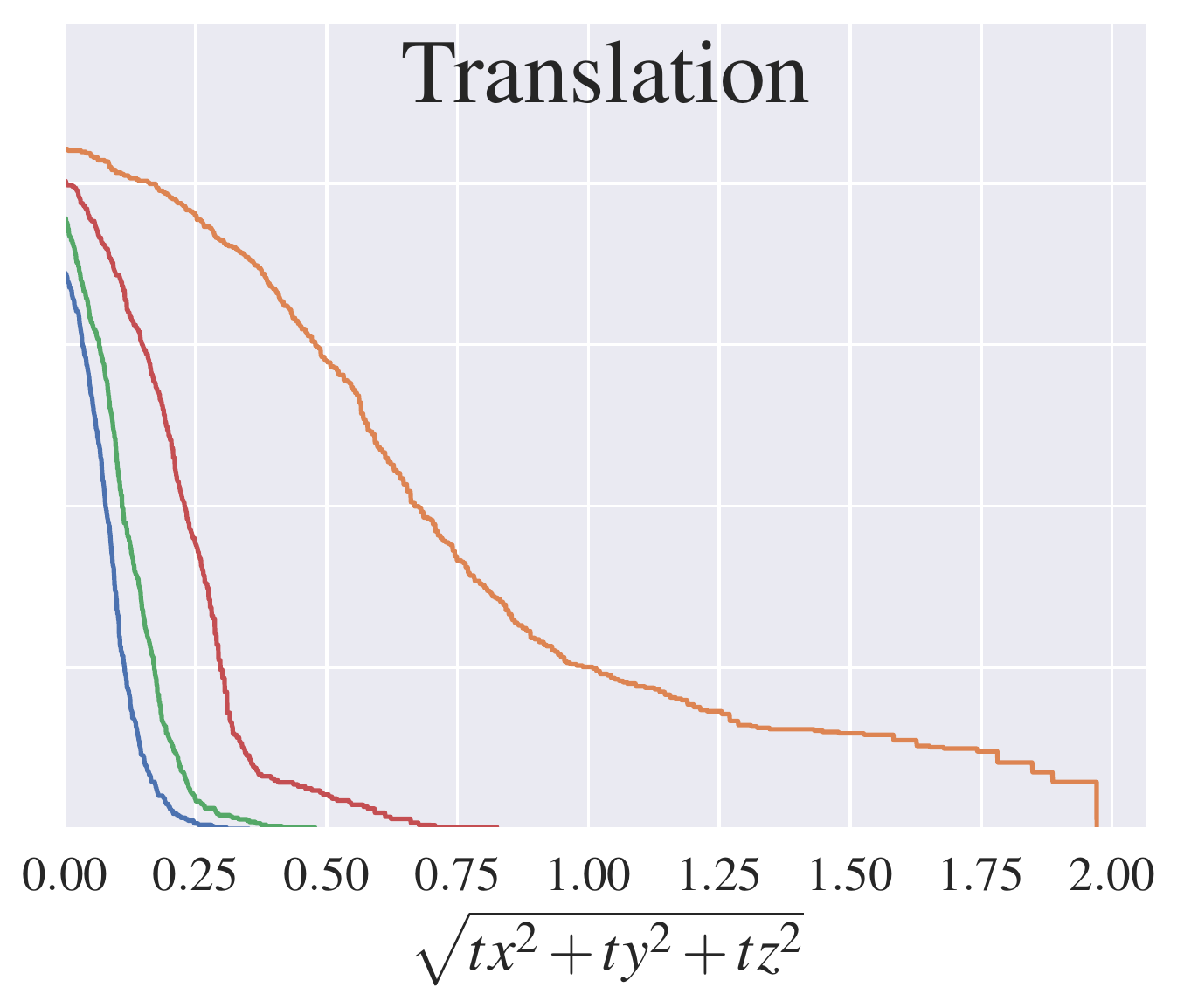}
    \includegraphics[trim={0cm 0cm 0cm 0cm},width=.19\linewidth,height=3cm,clip]{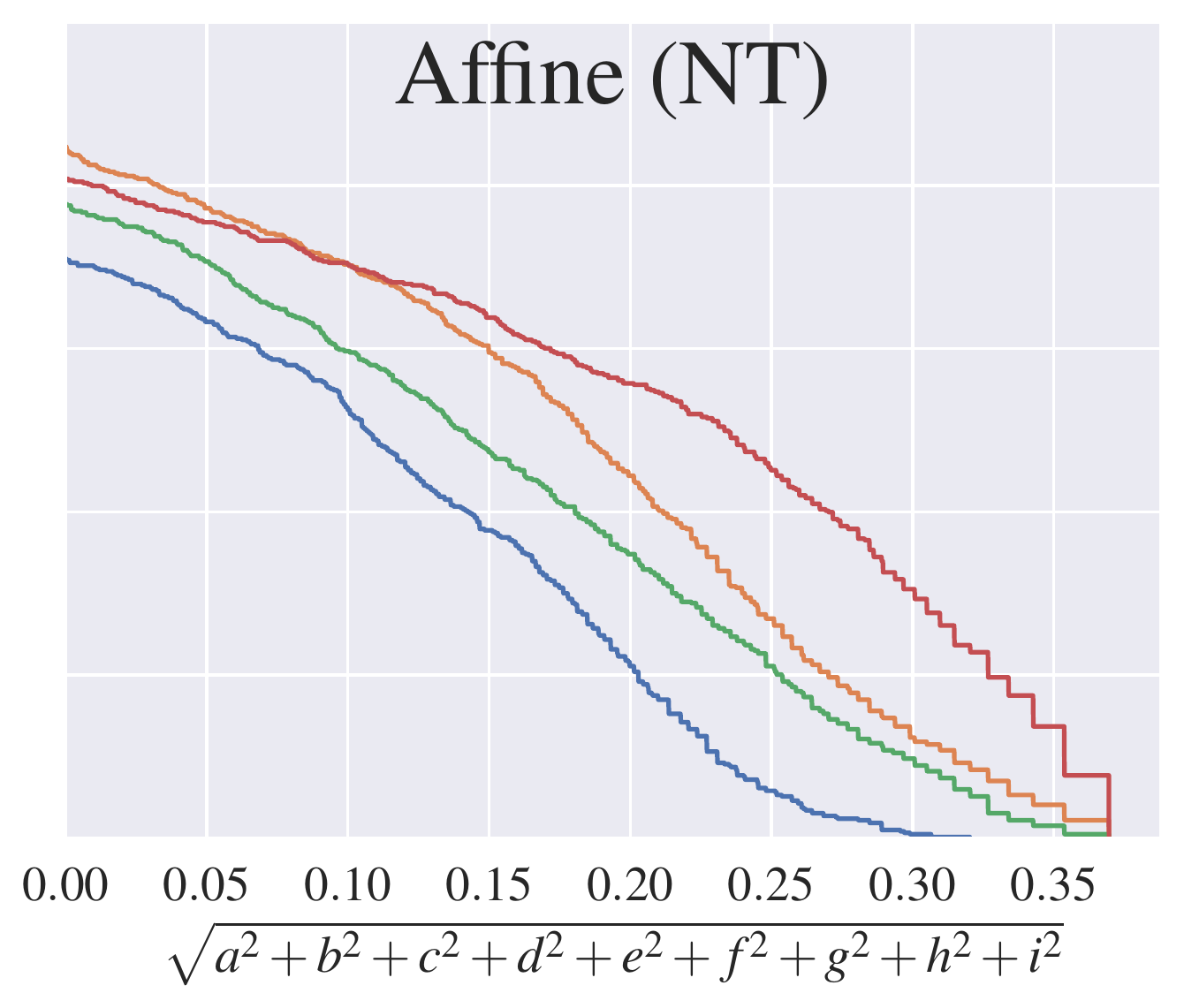}
    \includegraphics[trim={0cm 0cm 0cm 0cm},width=.22\linewidth,height=3.07cm,clip]{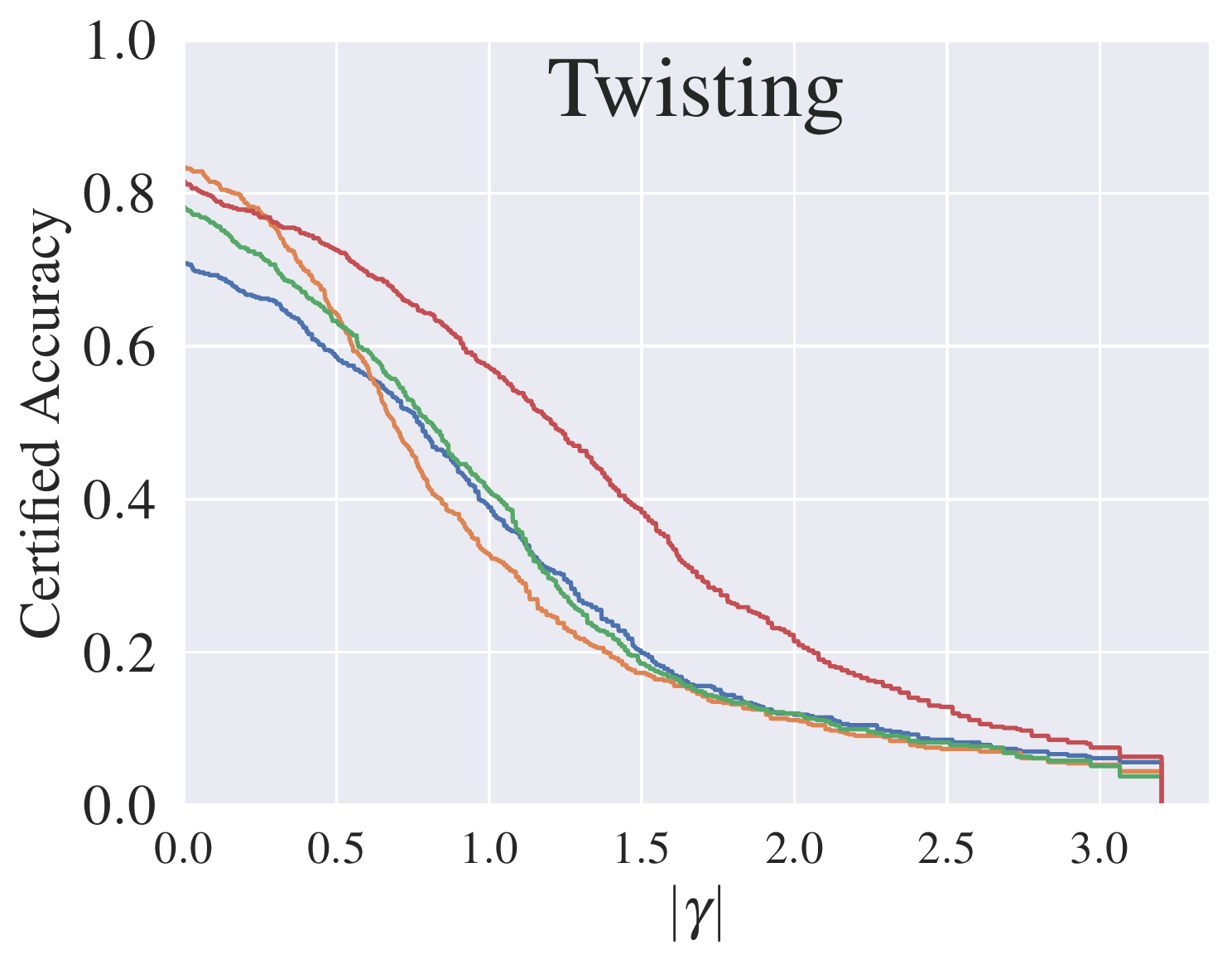}
    \includegraphics[trim={0cm 0cm 0cm 0cm},width=.19\linewidth,height=3cm,clip]{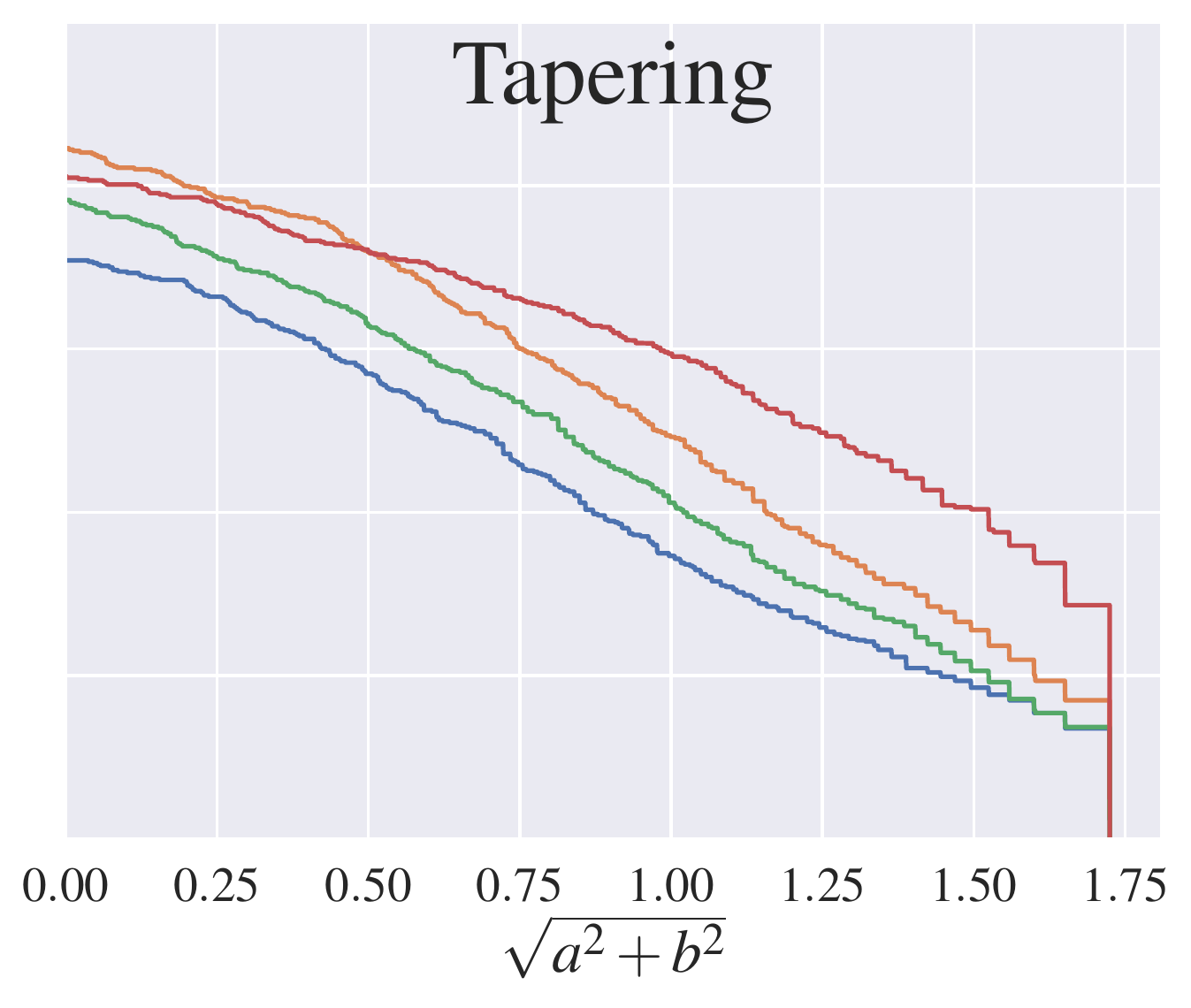}
    \includegraphics[trim={0cm 0cm 0cm 0cm},width=.19\linewidth,height=3cm,clip]{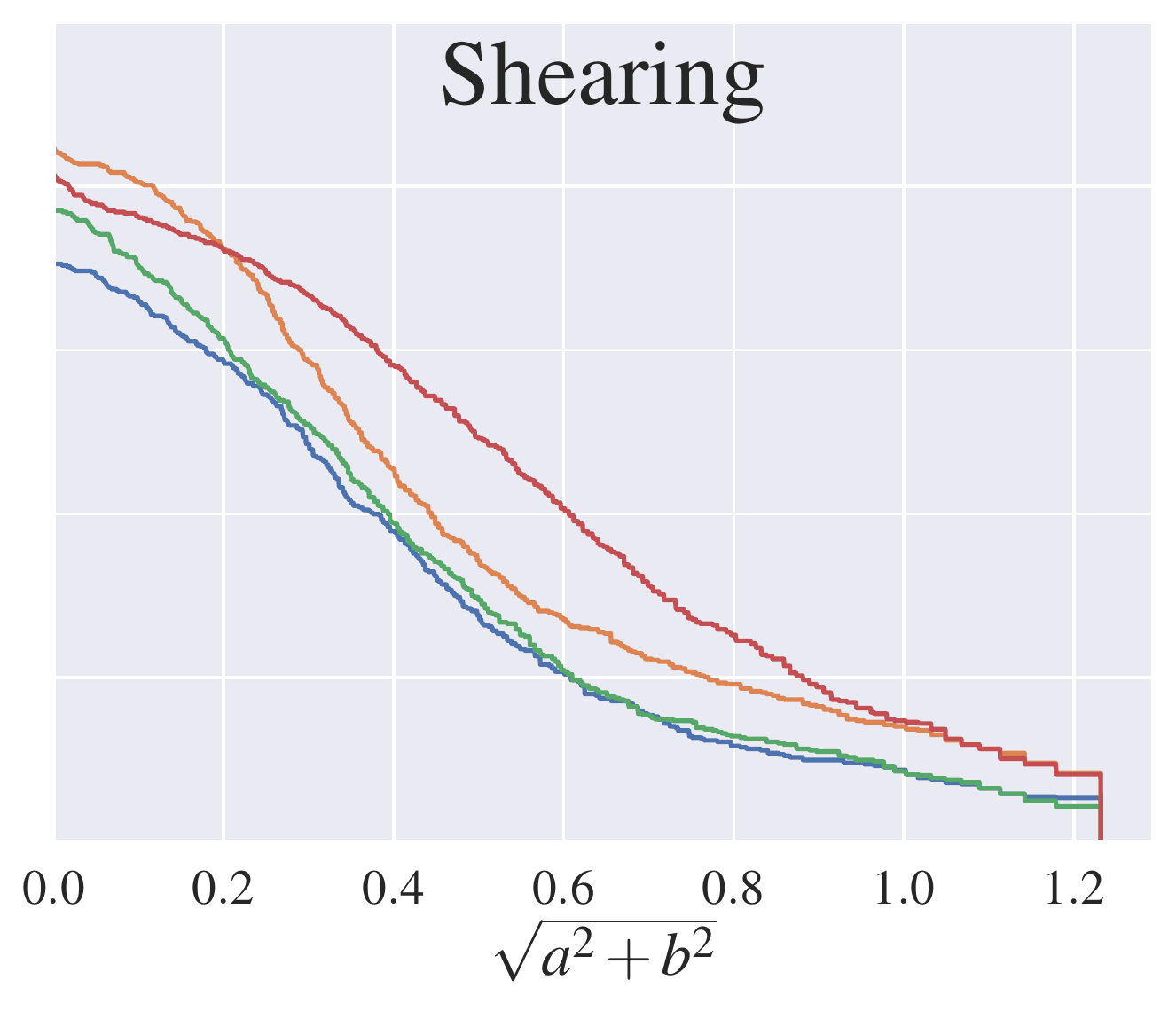}
    \includegraphics[trim={0cm 0cm 0cm 0cm},width=.19\linewidth,height=3cm,clip]{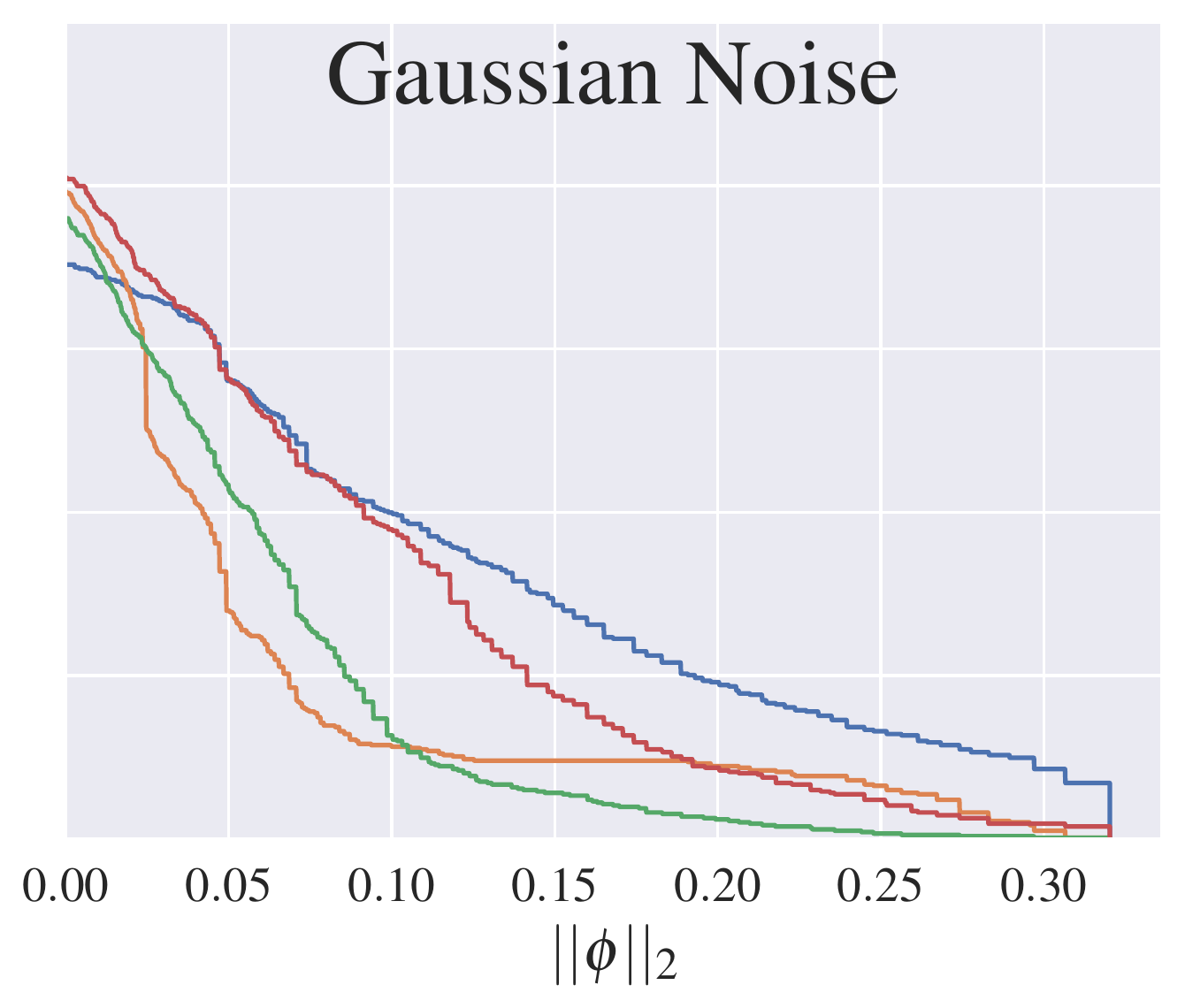}
    \includegraphics[trim={0cm 0cm 0cm 0cm},width=.19\linewidth,height=3cm,clip]{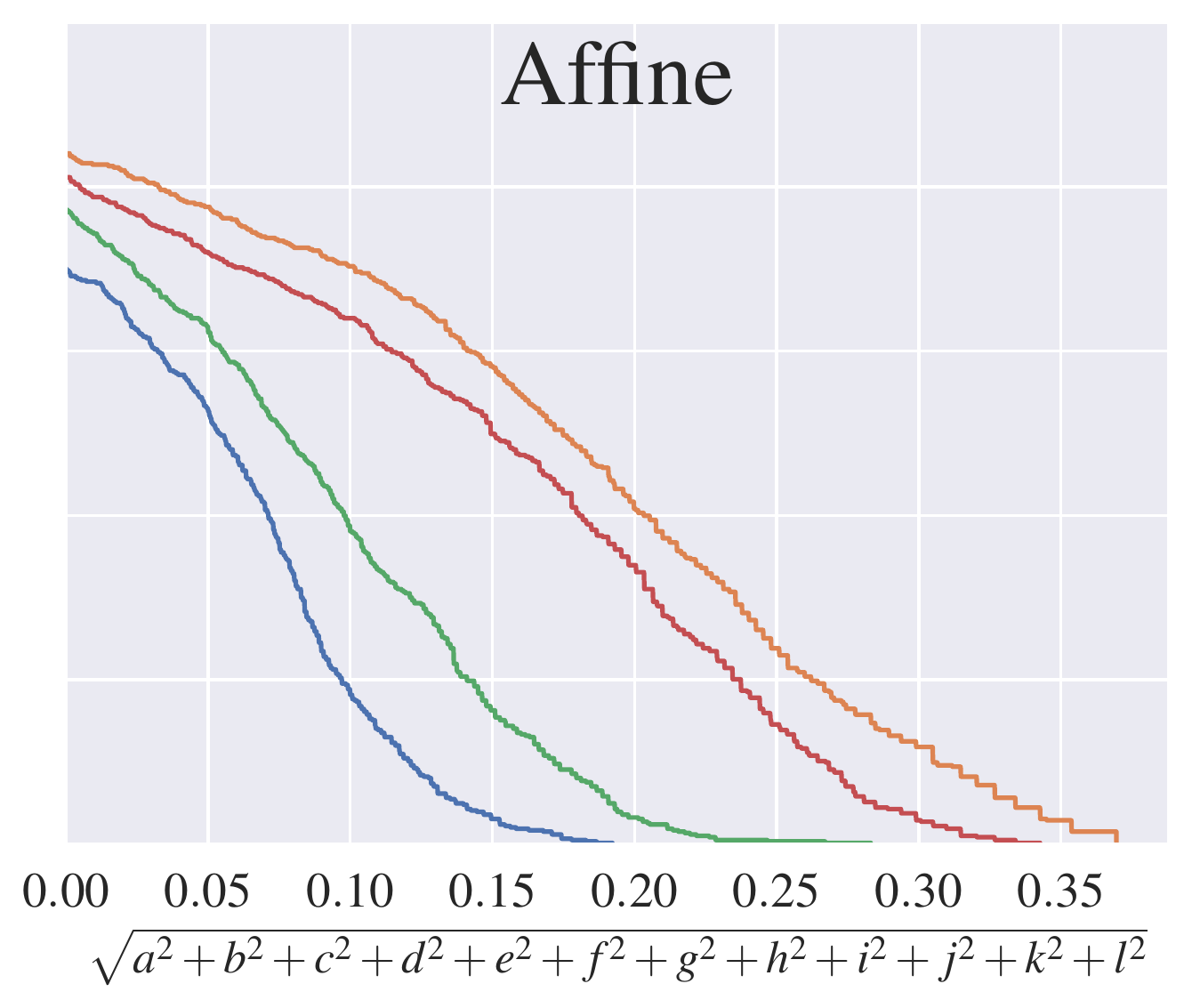}
    \caption{\label{fig:MainCurveScanObjectNN}
    \textbf{3DeformRS Certification on ScanObjectNN} of four point cloud DNNs against 10 transformations.
    }
\end{figure*}

\section{Experiments}
\subsection{Setup}

\paragraph{Datasets}
We experiment on ModelNet40~\cite{ModelNet} and ScanObjectNN~\cite{ScanObjectNN}.
ModelNet40 comprises $12,311$ 3D CAD models from $40$ classes. 
ScanObjectNN has $2,902$ real-world 3D scans of $15$ classes. 
While ModelNet40 has synthetic objects, ScanObjectNN used 3D cameras and so contains natural and challenging self-occlusions. 
We use the ScanObjectNN variant that omits background data.
For both datasets, the shapes are dimension-normalized, origin-centered and sampled with $1024$ points.

\paragraph{Models}
We study four point cloud DNNs: PointNet~\cite{PointNet}, PointNet++~\cite{PointNet++}, DGCNN~\cite{DGCNN} and CurveNet~\cite{CurveNet}.
For PointNet, we used 3DCertify's implementation, which uses $z-$rotation augmentation on ModelNet40. 
For CurveNet, we used the official implementation's weights, trained on ModelNet40 with axis-independent $[.66,1.5]$ scaling and $\pm 0.2$ translation.
On ScanObjectNN, PointNet and CurveNet are trained without augmentation.
For PointNet++ and DGCNN, we used the PyTorch Geometric library~\cite{Fey/Lenssen/2019} with default hyper-parameters and without augmentations.

\begin{table*}
    \centering
    \resizebox{\linewidth}{!}{
    \begin{tabular}{l|cccccccccc}
        \toprule
        \textbf{ModelNet40}                      & $z-$Rot.             & $xz-$Rot.            & $xyz-$Rot.           & Shearing          & Twisting          & Tapering          & Translation       & Affine            & Affine (NT)       & Gaussian Noise \\  \midrule
        PointNet   \cite{PointNet}      & \textbf{2.64}     & \underline{0.31}  & \underline{0.21}  & \underline{0.42}  & \textbf{2.48}     & \underline{0.50}  & \underline{0.07}  & \underline{0.06}  & \underline{0.13}  & 0.09 \\
        PointNet++ \cite{PointNet++}    & 1.45              & 0.34              & 0.25              & 0.55              & 1.66              & 0.81              & 0.64              & 0.16              & 0.17              & \underline{0.07} \\
        DGCNN      \cite{DGCNN}         & 1.29              & 0.33              & 0.26              & 0.54              & 1.78              & 0.90              & 0.18              & 0.14              & 0.20              & \textbf{0.10} \\
        CurveNet   \cite{CurveNet}      & \underline{0.76}  & \textbf{0.39}     & \textbf{0.34}     & \textbf{0.56}     & \underline{1.51}  & \textbf{1.09}     & \textbf{1.32}     & \textbf{0.26}     & \textbf{0.26}     & 0.08 \\ 
        \bottomrule 
        \toprule
        \textbf{ScanObjectNN}                    & $z-$Rot.             & $xz-$Rot.            & $xyz-$Rot.           & Shearing          & Twisting          & Tapering          & Translation       & Affine            & Affine (NT)       & Gaussian Noise \\  \midrule 
        PointNet   \cite{PointNet}      & \underline{0.30}  & \underline{0.16}  & \underline{0.15}  & \underline{0.35}  & 0.90              & \underline{0.72}  & \underline{0.06}  & \underline{0.05}  & \underline{0.11}  & \textbf{0.10} \\
        PointNet++ \cite{PointNet++}    & 0.38              & 0.23              & 0.22              & 0.45              & \underline{0.89}  & 0.93              & \textbf{0.67}     & \textbf{0.16}     & 0.17              & \underline{0.05} \\
        DGCNN      \cite{DGCNN}         & 0.32              & 0.20              & 0.19              & 0.37              & 0.92              & 0.81              & 0.10              & 0.08              & 0.14              & \underline{0.05} \\
        CurveNet   \cite{CurveNet}      & \textbf{0.51}     & \textbf{0.31}     & \textbf{0.29}     & \textbf{0.51}     & \textbf{1.24}     & \textbf{1.02}     & 0.19              & 0.14              & \textbf{0.20}     & 0.08 \\
        \bottomrule
    \end{tabular}
    }
    \caption{\label{tab:MainACR}
    \textbf{Certified Robustness Assessment on ModelNet40 and ScanObjectNN.} 
    We report the Average Certification Radius (ACR) of all DNNs against 10 deformations, on each dataset. 
    For each deformation, we \textbf{embolden} the best performance and \underline{underline} the worst.
    }
\end{table*}

\paragraph{Certification}
In our experiments, following~\cite{salman2019provably, zhai2019macer, deformrs}, we construct the hard version of the smooth classifier in Eq.~\eqref{eq:3d-domain-smooth-classifier} for each deformation. 
Moreover, we follow common practice and adapt the public implementation from~\cite{cohen2019certified} for 
estimating the certified radius via Monte Carlo sampling. 
In particular, we use $1{,}000$ samples to estimate the certified radius with a probability of failure of $10^{-3}$. 
For all experiments, we provide envelope certified accuracy curves cross validated at several values of smoothing parameters, detailed in the \textbf{Appendix}.
Since all deformations used Gaussian smoothing, the certificates we find are in the $\ell_2$ sense (except for rotation, which used uniform smoothing and so its certificate is in the $\ell_1$ sense).

\subsection{Benchmarking 3D Networks}\label{sec:benchmarking}
We present certification curves in Figures~\ref{fig:MainCurveModelNet40} and~\ref{fig:MainCurveScanObjectNN} for ModelNet40 and ScanObjectNN, respectively.
We also report each curve's associated Average Certification Radius (ACR) in Table~\ref{tab:MainACR} as a summary metric.
We further highlight nine main observations from these results, and leave a detailed analysis, together with ablations on RS' hyper-parameters, to the \textbf{Appendix}.

\paragraph{Vulnerability against Rigid Transformations}
Our analysis considers, among others, rotation and translation.
Remarkably, we find that DNNs are significantly vulnerable even against these rigid perturbations.
For most DNNs, the certified accuracy plots in Figures~\ref{fig:MainCurveModelNet40} and~\ref{fig:MainCurveScanObjectNN} show that performance drops dramatically as the perturbation's magnitude increases.
This phenomenon supports further research on increasing the robustness of point cloud DNNs against simple transformations that could happen in the real world.

\paragraph{Deformation Complexity}
Each spatial deformation is parameterized by a certain number of values.
The number of parameters can be associated with the deformation's complexity: ``simple'' deformations, \ie rotation around the $x$ axis, require few parameters, while ``complex'' deformations, \ie affine, require several parameters.
Under these notions, we notice that as the deformation's complexity increases, the DNNs' certified accuracies drop more rapidly.
This observation agrees with intuition: complex transformations should be harder to resist than simple ones.

\paragraph{Gaussian Noise}
This deformation is arguably the most general and information-destroying, as it may not preserve distances, angles nor parallelism.
Experimentally, we indeed observe that DNNs are rather brittle against this noise: even small magnitudes can break the DNNs' performance.

\paragraph{CurveNet Performs Remarkably}
In terms of ACR, CurveNet displays larger robustness than competitors across the board.
In particular, Table~\ref{tab:MainACR} (top), shows that, on ModelNet40, CurveNet achieves the best ACR for seven of the 10 deformations we present, while scoring last only in two deformations ($z-$Rot and Twisting).
Analogously, for ScanObjectNN (Table~\ref{tab:MainACR}, bottom), CurveNet is the best performer, displaying the best robustness for seven of the 10 deformations we consider, while never scoring last.

\paragraph{PointNet Performs Poorly}
On ModelNet40, reported in Table~\ref{tab:MainACR} (top), PointNet achieves the lowest ACR values for nine of the 10 deformations, while only achieving the best score in two deformations ($z-$Rot and Twisting).
Similarly, for ScanObjectNN, in Table~\ref{tab:MainACR} (bottom), PointNet consistently scores last across all transformations, holding the last position in eight out of 10 transformations, and only holding the first position in only one (Gaussian noise).

\paragraph{Training-time Augmentations Boost Certified Robustness}
On ModelNet40, PointNet was trained with $z$-rotation augmentations.
Indeed, we find that, when PointNet is evaluated on ModelNet40, it displays superior performance against rotation-based transformations along the $z$ axis, \ie $z-$Rot and Twisting.
In particular, its ACR is  $>0.8$ more than the runner-up in $z-$Rot and  $\sim0.4$ in Twisting, while its robust accuracy is mostly maintained across the entire rotation regime (from $-\pi$ to $+\pi$ radians, \ie all possible rotations).
That is, PointNet correctly classifies most objects, independent of whether they are rotated or twisted around the $z$ axis.
More interestingly, we also remark how PointNet's dominance is \emph{not} observed in ScanObjectNN, where it was \emph{not} trained with $z$-rotation augmentations.
We thus attribute PointNet's robustness to the training-time augmentations it enjoyed, and so we further study this phenomenon in the next subsection.

\paragraph{Certified \emph{vs.} Regular Accuracy}
Overall, our analysis finds that the best performer is CurveNet, while the worst performer is PointNet.
Notably, this fact agrees with each DNN's plain test performance: CurveNet has an advantage over PointNet of about 7\% and 10\% in ModelNet40 and ScanObjectNN, respectively.
Thus, our results find a correlation between regular and certified performance.

\paragraph{ModelNet40 and ScanObjectNN}
CurveNet is, arguably, the best performer on both datasets in terms of ACR.
However, we note that \emph{all} of CurveNet's certified accuracies drop significantly from ModelNet40 to ScanObjectNN.
Thus, our results agree with how ModelNet40's synthetic nature (compared to ScanObjectNN's realistic nature) implies that ModelNet40 is a ``simpler'' dataset than ScanObjectNN.

\begin{table}
    \centering
    \resizebox{0.85\linewidth}{!}{
    \begin{tabular}{l|cc}
        \toprule
                                        & ModelNet40    & ScanObjectNN \\ \midrule
        PointNet   \cite{PointNet}      & 85.94\%       & 71.35\% \\
        PointNet++ \cite{PointNet++}    & 90.03\%       & 83.53\% \\
        DGCNN      \cite{DGCNN}         & 90.03\%       & 78.04\% \\
        CurveNet   \cite{CurveNet}      & 93.84\%       & 81.47\% \\
        \bottomrule
    \end{tabular}
    }
    \caption{\label{tab:plain-performances}
        \textbf{Test Set Accuracy} on ModelNet40 and ScanObjectNN.
    }
\end{table}

\paragraph{Sizable Variations in Robustness}
For a given deformation and dataset, we note sizable robustness differences across DNNs.
Specifically, we observe that one DNN may obtain an ACR even 10$\times$ larger than other DNN.
This phenomenon happens even when the plain accuracy of the DNNs being certified is mostly comparable, as reported in Table~\ref{tab:plain-performances}.
Thus, we argue that certified robustness should be a design consideration when developing DNNs.
That is, our results suggest that \emph{(i)} plain accuracy may not provide the full picture into a DNN's performance, and \emph{(ii)} certified accuracy may be effective and efficient for assessing DNNs.

\subsection{Simple but Effective: Augmented Training}
\begin{figure}[t]
    \centering
    \includegraphics[width=\columnwidth]{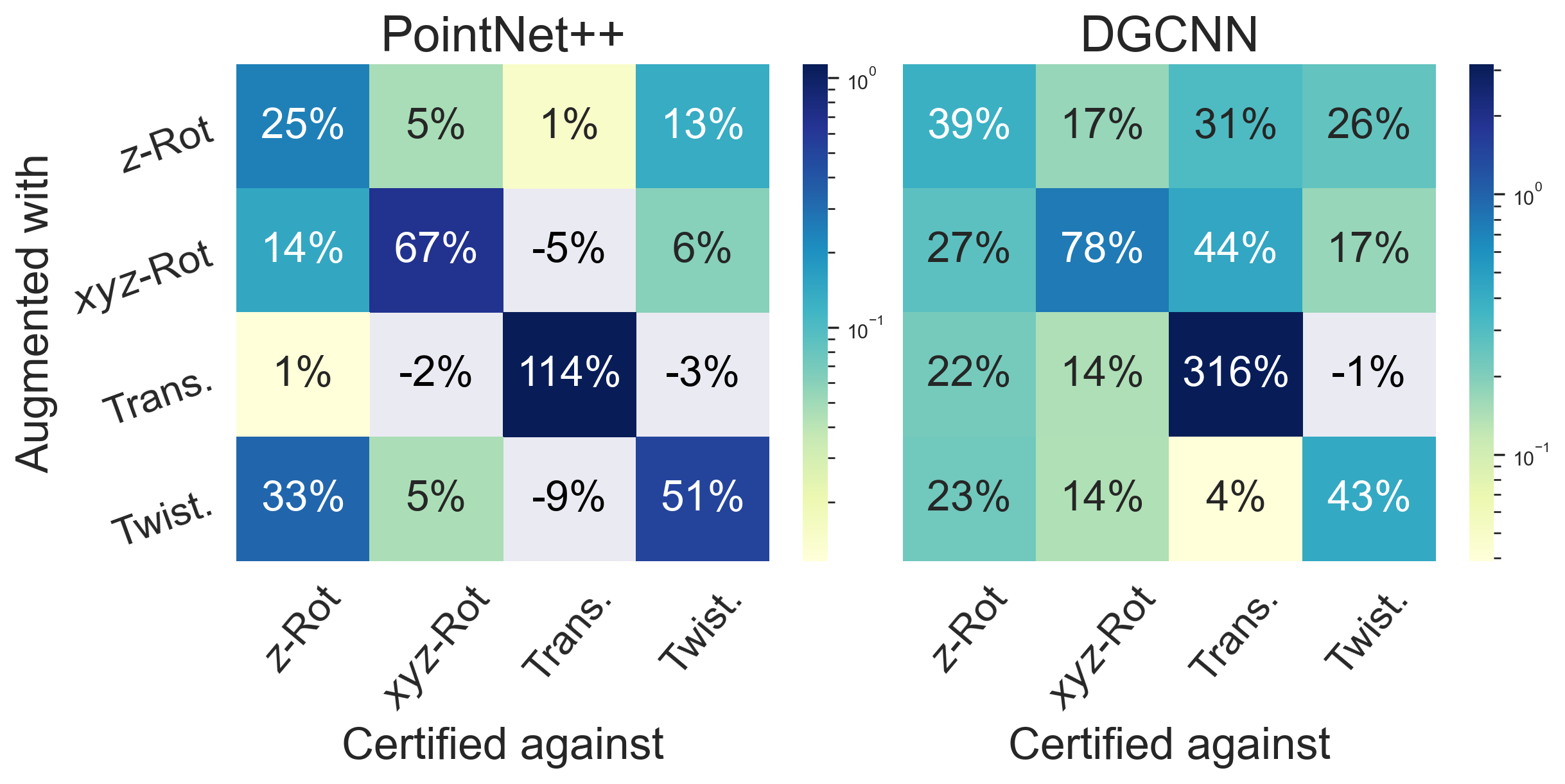}
    \caption{\label{fig:augmentation_results}
        \textbf{Relative effects of training augmentations on Average Certified Accuracy.} 
        We conduct augmentations during training and record relative improvements on certified accuracy. 
        Most training augmentations improve certification across the board.
    }
\end{figure}

Previously, in Figure~\ref{fig:MainCurveModelNet40}, we observed PointNet's dominant performance on $z-$axis rotation and twisting. 
We attributed this superiority to augmentations PointNet enjoyed during training.
Here we investigate the effect of other training augmentations on certified robustness.
In particular, we train two DNNs with four different deformations and then assess their certified robustness against such deformations.
We report the results of this experiment in Figure~\ref{fig:augmentation_results}. 

We draw the following three conclusions from these results. 
\textbf{(i)}~Conducting augmentation with one deformation significantly increases the robustness against that same deformation. 
This is an expected phenomenon, as the model trained on deformed versions of inputs.
\textbf{(ii)}~Training on some deformations yields robustness against other deformations (\eg, augmenting with twisting results in robustness against rotation). 
This observation aligns with out earlier result where PointNet displayed superiority in $z-$axis rotation and twisting. 
This result further suggests that simple training augmentations strategies are effective for ``robustifying'' models against deformations. 
\textbf{(iii)}~The larger certificates come at virtually no cost on clean accuracy: each model's clean accuracy varied less than $\pm1.5\%$ w.r.t. the model trained without augmentations (refer to Table~\ref{tab:performance-change}).

\begin{table}
    \centering
    \resizebox{\linewidth}{!}{
    \begin{tabular}{l|c|cccc}
        \toprule
                    & Plain     & $z$-Rot.  & $xyz$-Rot.    & Trans.    & Twist. \\ \midrule
        PointNet++  & 90.1\%   & +0.2\%    & -0.3\%        & -0.4\%    & -0.1\% \\
        DGCNN       & 89.8\%   & +1.3\%    & +0.8\%        & +1.3\%    & -0.2\% \\
        \bottomrule
    \end{tabular}
    }
    \caption{\label{tab:performance-change}
        \textbf{Augmentation impact on accuracy on ModelNet40.}
    }
\end{table}

\subsection{Comparison with 3DCertify}\label{sec:comparison}
We further compare 3DeformRS and 3DCertify~\cite{lorenz2021robustness}, which is, to the best of our knowledge, the current state-of-the-art certification approach against spatial deformations.
3DCertify's analysis focuses exclusively on PointNet and ModelNet40.
We used their official implementation and pre-trained weights, and note that 3DCertify certifies a 100-instance subset of the test set. 
We note that the certified accuracies reported in~\cite{lorenz2021robustness} are w.r.t. the correctly-classified samples from such subset.
Moreover, we underscore that 3DCertify’s analysis focuses exclusively on PointNet trained on ModelNet40.

\paragraph{Technical Details}
We first discuss the intricacies of providing a fair comparison between 3DCertify (an exact verification approach) and 3DeformRS (a probabilistic certification approach).
We underscore fundamental differences regarding the input each approach receives when certifying \textbf{(i)}~a given DNN, \eg PointNet, \textbf{(ii)}~w.r.t. a desired transformation, \eg Rotation, on \textbf{(iii)}~a selected point cloud input, \eg a plane.
In particular, 3DCertify receives as additional input the exact magnitude of the transformation.
3DCertify then uses the transformation and its magnitude, \eg $3^\circ$, to return a boolean value stating if the given DNN's output is certifiable against a transformation of such magnitude.
On the other hand, 3DeformRS mechanism is fundamentally different: it computes a tight certification radius for the same (DNN-transformation-point cloud) tuple, but leveraging a random distribution with which the input was smoothed.
The two approaches can be compared if the certified radius provided by 3DeformRS is evaluated against the transformation magnitude received by 3DCertify.

However, we note that the certification radius computed by 3DeformRS depends on the hyper-parameter $\sigma$ (or $\lambda$ for uniform smoothing) from Corollary~\ref{cor:parametric-certification}.
That is, 3DeformRS will compute a different radius for each $\sigma$ considered.
While there exists a $\sigma^\star$ with which 3DeformRS provides the \emph{largest} certified radius for the transformation being considered, $\sigma^\star$ is not known \emph{a priori}.
Thus, we are required to run a grid search of $\sigma$ values to compute each instance's largest certified radius.
While this procedure is computationally intensive, the output of \emph{each} experiment---considering one $\sigma$ over the whole dataset---provides an entire certified accuracy \emph{curve}, in contrast to \emph{a single point} in this curve, as provided by exact verification approaches. 
We underscore how this curve can provide insights into a DNN's robustness.
In practice, when comparing with 3DCertify on specific transformation magnitudes, we run a grid search of at most 18 $\sigma$ values to obtain each instance's largest certified radius.
Hence, we report certified accuracy envelope curves, \ie certified accuracy curves that consider each instance's largest certified radius.
Additionally, to circumvent the problem arisen from considering different norms (3DCertify's $\ell_\infty$ norm \emph{vs.} 3DeformRS's $\ell_1$ or $\ell_2$), we limit our comparison to single-parameter transformations.

\paragraph{Results} 
We compare against 3DCertify, in their reported experimental setup, along three dimensions: \textbf{(i)}~certificate magnitude, \textbf{(ii)}~point cloud cardinality and \textbf{(iii)}~speed.
Overall, we find that 3DeformRS provides comparable-to-better certificates, scales well with point cloud size, and has manageable computational cost.

\paragraph{Certificate magnitude}
We consider rotations w.r.t. each axis, and evaluate 3DCertify's official 64-points DNN against rotations of $\gamma \in \{1,2,3,4,5,6,7,8,10,15\}$ degrees.
Whenever possible, we used 3DCertify's reported results, otherwise we used their public implementation to run certification.
We find that 3DeformRS achieves comparable-to-better certificates, while providing the additional benefit of full certified accuracy curves, instead of individual points in these plots.
Refer to the \textbf{Appendix} for such certified accuracy curves.

\begin{table}
    \centering
    \resizebox{\linewidth}{!}{
    \begin{tabular}{l|ccccccc}
        \toprule
        Points  & 16 & 32 & 64 & 128 & 256 & 512 & 1024 \\ \midrule
        Boopathy~\etal~\cite{boopathy2019cnn} & 3.7 & 3.6 & 3.3 & 2.2 & 4.4 & 5.6 &                                       6.7\\
        DeepPoly  & 95.1 & 94.0 & 91.3 & 72.2 & 51.1 &                                          39.3 & 28.1\\
        3DCertify (Taylor 3D)  & 97.5 & 94.0 & 93.5 &                                                      81.1 & 66.7 & 49.4 & 37.1\\\midrule
        3DeformRS (ours) & \textbf{98.8} & \textbf{97.6} & \textbf{97.8} & \textbf{100} & \textbf{100} & \textbf{97.8} & \textbf{100} \\
        \bottomrule
    \end{tabular}
    }
    \caption{\label{tab:diffSamplVerif}
        \textbf{$3^\circ$ $z-$ Rotation Certificates when varying Point Cloud Cardinality.}
        The three previous methods consider linear relaxations and use the DeepPoly verifier, while 3DeformRS is based on Randomized Smoothing.
        Baselines taken from~\cite{lorenz2021robustness}.}
\end{table}

\paragraph{Point Cloud Cardinality}
3DCertify provides exact verification on point clouds of limited size~(64 points reported in~\cite{lorenz2021robustness}).
However, DNNs may deal with point clouds of at least $1024$ points in practice.
We experiment on this setup and compare with previous approaches by varying the point clouds' size and certifying against a $3^\circ$ $z-$rotation.
For this experiment, we follow 3DCertify's setup~\cite{lorenz2021robustness} and use the same DNN weights for each certification method.
Table~\ref{tab:diffSamplVerif} shows that our approach provides a better certificate for DNNs when trained and tested on large point clouds.
In particular, we mark three main observations.
First, 3DeformRS performs up to 60\% better on the realistic setup of $1024$ points.
Second, 3DeformRS provides a certified accuracy of over 97\% across the board; while large robustness is expected from a model augmented with $z-$rotations, our approach shows that providing such certificates \emph{is} possible. 
Third, 3DeformRS breaks the trend of decreasing certificates from which other approaches suffered when handling larger point clouds. 
Thus, we find that 3DeformRS enjoys scalability and invariance to larger input size. 

\paragraph{Speed}
Certification via 3DCertify~\cite{lorenz2021robustness} is computationally expensive, and its current official implementation cannot benefit from GPU hardware accelerators.
Furthermore, 3DCertify's computational cost scales with the perturbation's magnitude, hindering certification against large perturbations.
For example, certifying PointNet with $64$ points against an $z-$ rotation on a CPU requires $\sim13$k seconds for $1^\circ$, and up to $\sim89$k seconds for $10^\circ$.

Unlike 3DCertify, 3DeformRS enjoys a virtually-constant computation cost, as it requires a fixed number of forward passes and $\sigma$ values.
In addition, our native implementation can leverage GPUs for accelerating certification.
Certifying against $z-$ rotation with 3DeformRS on \emph{the same} CPU requires only $\sim40$ seconds per $\sigma$~(independent of $\sigma$'s magnitude, as reported in Table~\ref{tab:running times}).
Even the extreme case of certifying with $100$ values for $\sigma$, arguably an unnecessary amount, still requires only $\sim4$k seconds.
That is, certifying small perturbations with 3DeformRS attains a $\sim3\times$ speed boost, while large perturbations can even enjoy a $\sim20\times$ boost.
Moreover, accelerating 3DeformRS via an Nvidia V100 GPU provides a $5\times$ boost over its CPU counterpart, further improving runtime over 3DCertify.

\begin{table}[t]
    \centering
    \resizebox{0.5\linewidth}{!}{
    \begin{tabular}{c|rrr}
        \toprule
        \multirow{2}{*}{Device} & \multicolumn{3}{c}{$\sigma$} \\
                                & 0.01  &  0.2  &  0.4  \\ \midrule
        CPU                     & 38.8  & 39.9  & 40.4  \\ 
        GPU                     & 7.3   &  7.4  &  7.7  \\
        \bottomrule
    \end{tabular}
    }
    \caption{\label{tab:running times}
        \textbf{Runtimes for 3DeformRS.} 
        We compare certification runtime for a single $\sigma$ value both in CPU and GPU~(values in seconds).
        3DeformRS's CPU version enjoys reasonable certification times, and leveraging a GPU lowers the runtime by $\sim5\times$.
    }
\end{table}

\subsection{Ablations}\label{sec:ablations}
The stochastic nature of 3DeformRS requires a Monte Carlo method followed by a statistical test to bound the probability of returning an incorrect prediction.
This statistical test is parameterized by a failure ratio $\alpha$ which, throughout our experiments, was set to the default value of $10^{-3}$, following~\cite{cohen2019certified}.
Here, we analyze the sensitivity of 3DeformRS to other failure probabilities $\alpha$.
We show in the \textbf{Appendix} the certified accuracy curves for failure probabilities $\alpha \in \{10^{-2},10^{-4},10^{-5}\}$. 
We underscore that, in our assessment, 3DeformRS shows negligible variation in certified accuracy w.r.t. changes in $\alpha$.
In particular, we notice that all ACRs are $\sim0.22$ for all the $\alpha$ values we considered.

\section{Conclusions and Limitations}
In this work, we propose 3DeformRS, a method for certifying point cloud DNNs against spatial deformations.
Our method provides comparable-to-better certificates than earlier works while scaling better to large point clouds and enjoying practical computation times.
These virtues or 3DeformRS allow us to conduct a comprehensive empirical study of the certified robustness of point cloud DNNs against semantically-viable deformations. 
Furthermore, 3DeformRS' practical runtimes may enable its usage in real-world applications.
While our stochastic approach is practical with its faster and top-performing certification, its stochasticity may also raise concerns when comparing against exact verification approaches.
Moreover, we note that our work solely focused on assessing the 3D robustness of point cloud DNNs against input deformations, disregarding other types of perturbations. 
Possible avenues for future work include incorporating better training algorithms such as MACER~\cite{zhai2019macer} and SmoothAdv~\cite{salman2019provably} for further robustness improvements.

\textbf{Acknowledgements.}
This publication is based upon work supported by the King Abdullah University of Science and Technology (KAUST) Office of Sponsored Research (OSR) under Award No. OSR-CRG2019-4033. 
We would also like to thank Jesús Zarzar for the help and discussions.
{
    \small
    \bibliographystyle{ieee_fullname}
    \bibliography{macros,main}
}
\newpage
\clearpage
\appendix



\section{Full Certified Accuracy Curves}
\begin{figure*}[ht]
    \centering
    \includegraphics[width=0.34\linewidth]{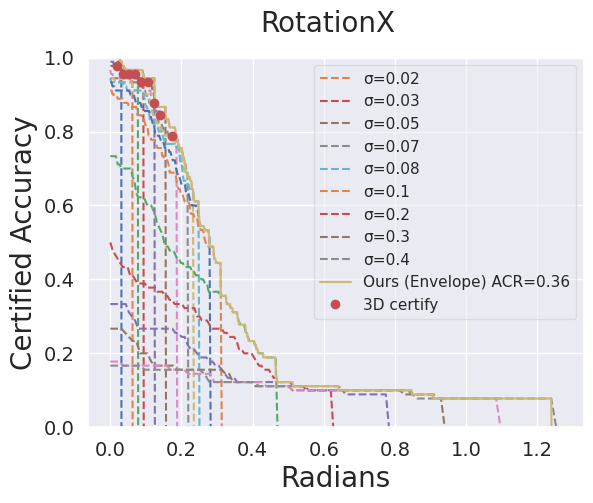}
    \includegraphics[width=0.32\linewidth]{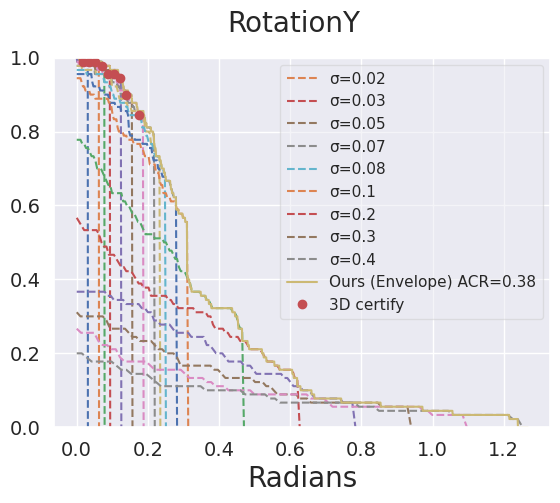}
    \includegraphics[width=0.32\linewidth]{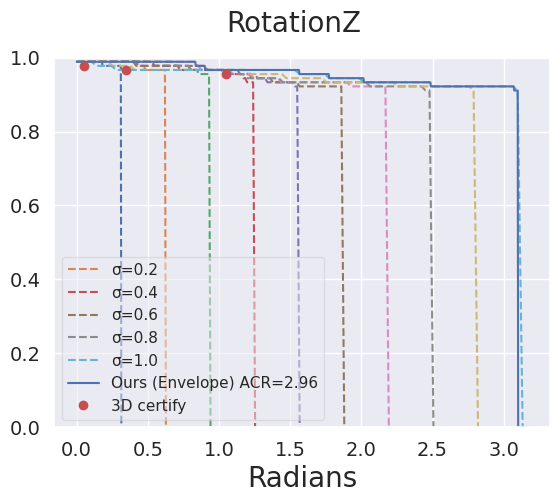}
    \caption{\textbf{Certified accuracy against $x-$, $y-$ and $z-$ rotations for PointNet's 64-point version.}
    The $z-$augmented PointNet maintains a certified accuracy of over $90\%$ for up to $360^\circ$ rotations.
    }
    \label{fig:64pointnetRotationZ}
\end{figure*}

Figure~\ref{fig:64pointnetRotationZ} reports the full certified accuracy curves that compare 3DeformRS against a previous certification approach (3DCertify~~\cite{lorenz2021robustness}).
Note that 3DeformRS achieves comparable-to-better certificates, while also enjoying full certified accuracy curves instead of individual points in these plots.

\section{Certified Accuracy Curves for Individual $\sigma$ values}
Every envelope curve we presented is the result of certifying with various $\sigma$ values and computing the maximum certified accuracy achieved at each certified radius.
Here, we elaborate on the results shown in the main paper, and report the curves corresponding to all the $\sigma$ values we considered.
Figures~\ref{fig:SuppRotationXYZ}--\ref{fig:SuppGaussianNoise} report these curves for ModelNet40 and ScanObjectNN.
Each curve, corresponding to a particular $\sigma$, is reported as a dashed line underneath the envelope.

\section{Detailed Analysis per Deformation}
\begin{figure*}[ht]
    \centering
    \includegraphics[width=0.95\linewidth,trim=0 1.5cm 0 0, clip]{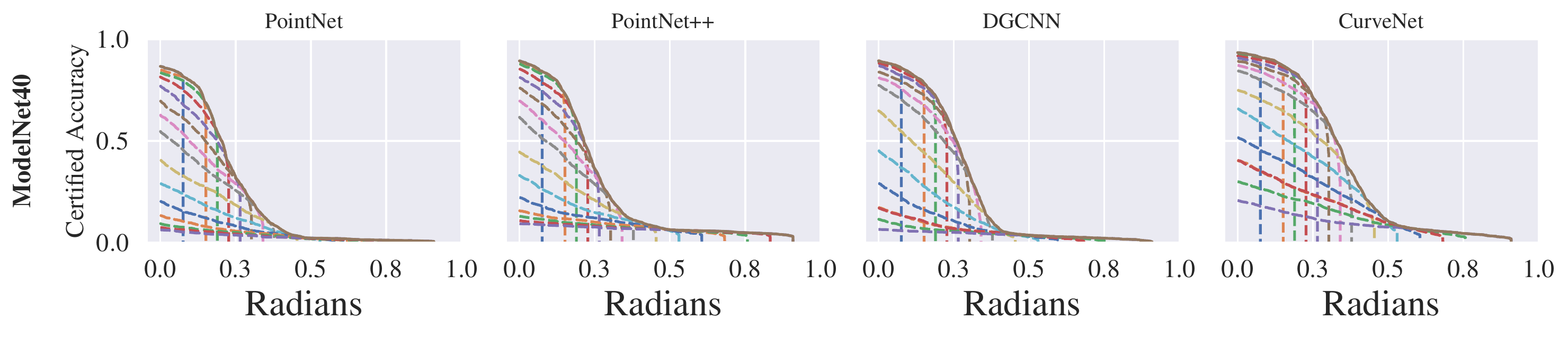}
    \includegraphics[width=0.95\linewidth,trim=0 0.3cm 0 0.55cm,clip]{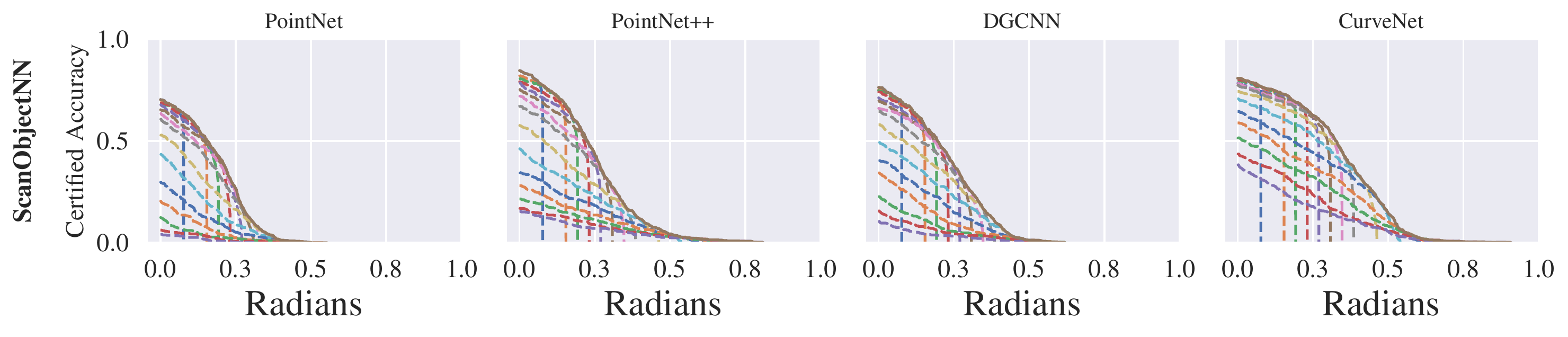}
    \includegraphics[width=0.8\linewidth]{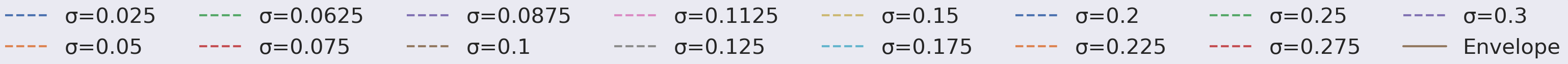}
    \caption{\textbf{RotationXYZ}. Certified accuracy for PointNet, PointNet++, DGCNN and CurveNet on ModelNet40 and ScanObjectNN.}
    \label{fig:SuppRotationXYZ}
\end{figure*}

\begin{figure*}[ht]
    \centering
    \includegraphics[width=0.95\linewidth,trim=0 1.5cm 0 0, clip]{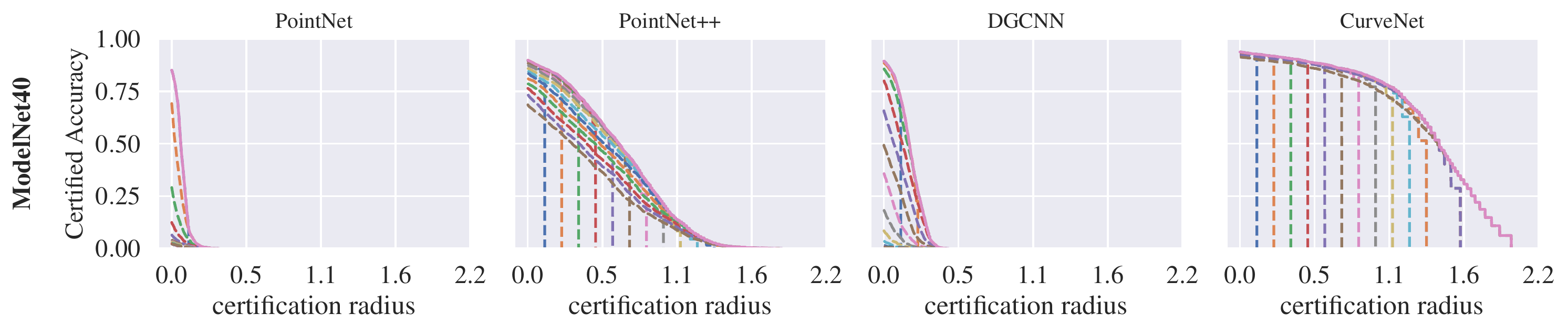}
    \includegraphics[width=0.95\linewidth,trim=0 0.3cm 0 0.55cm,clip]{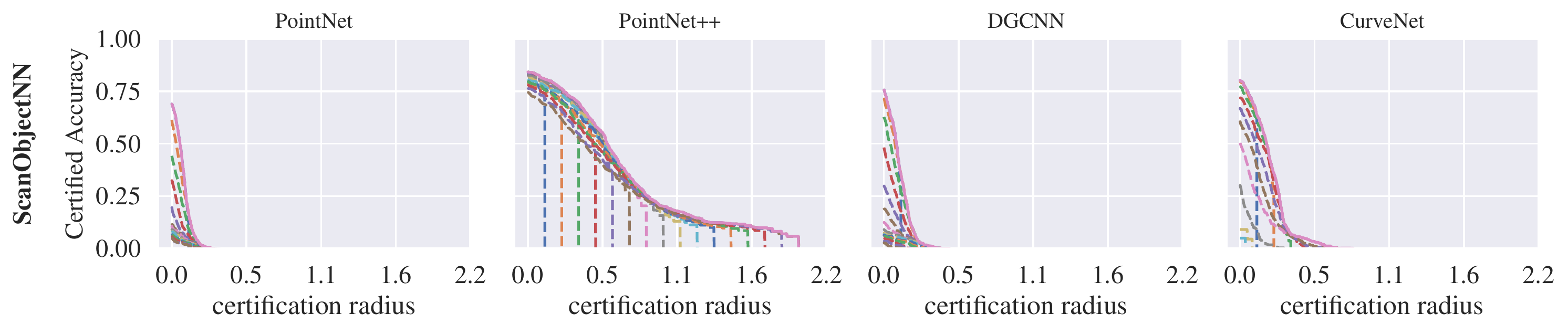}
    \includegraphics[width=0.8\linewidth]{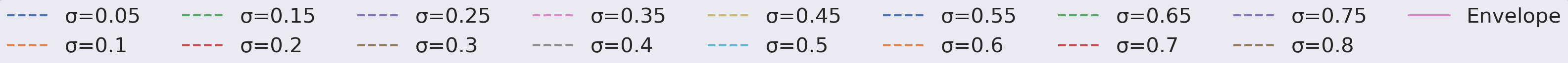}
    \caption{\textbf{Translation}. Certified accuracy for PointNet, PointNet++, DGCNN and CurveNet on ModelNet40 and ScanObjectNN.}
    \label{fig:SuppTranslation}
\end{figure*}

\begin{figure*}[ht]
    \centering
    \includegraphics[width=0.95\linewidth,trim=0 1.5cm 0 0, clip]{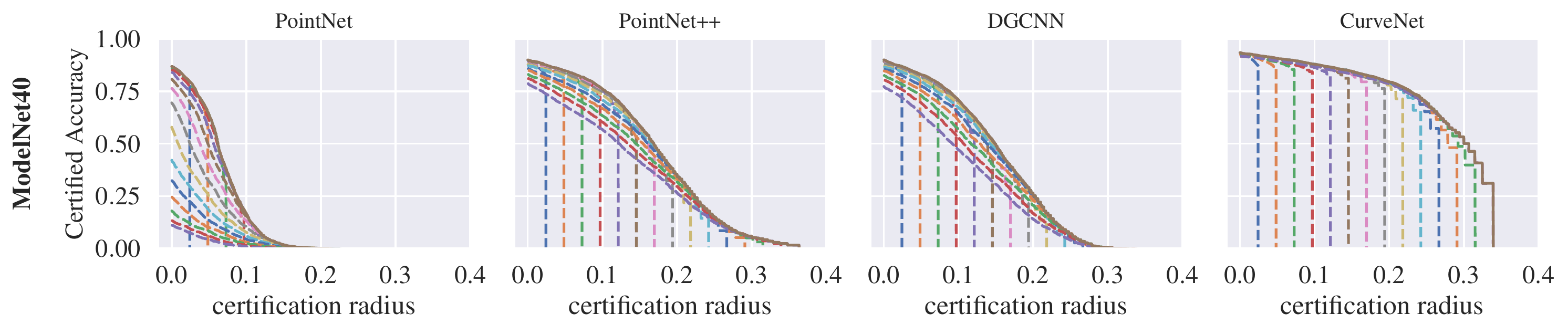}
    \includegraphics[width=0.95\linewidth,trim=0 0.3cm 0 0.55cm,clip]{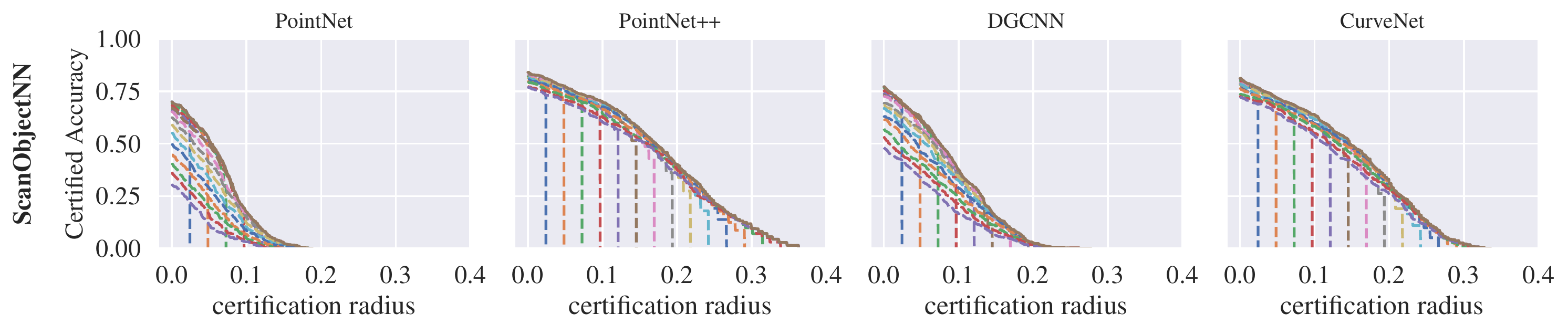}
    \includegraphics[width=0.8\linewidth]{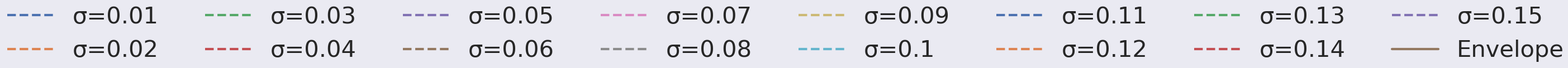}
    \caption{\textbf{Affine}. Certified accuracy for PointNet, PointNet++, DGCNN and CurveNet on ModelNet40 and ScanObjectNN.}
    \label{fig:SuppAffine}
\end{figure*}

\begin{figure*}[ht]
    \centering
    \includegraphics[width=0.95\linewidth,trim=0 1.5cm 0 0, clip]{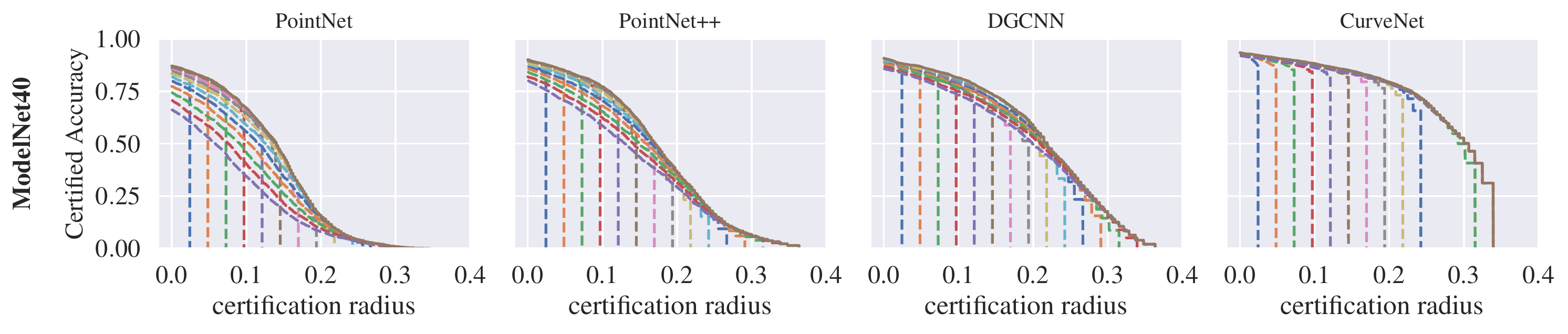}
    \includegraphics[width=0.95\linewidth,trim=0 0.3cm 0 0.55cm,clip]{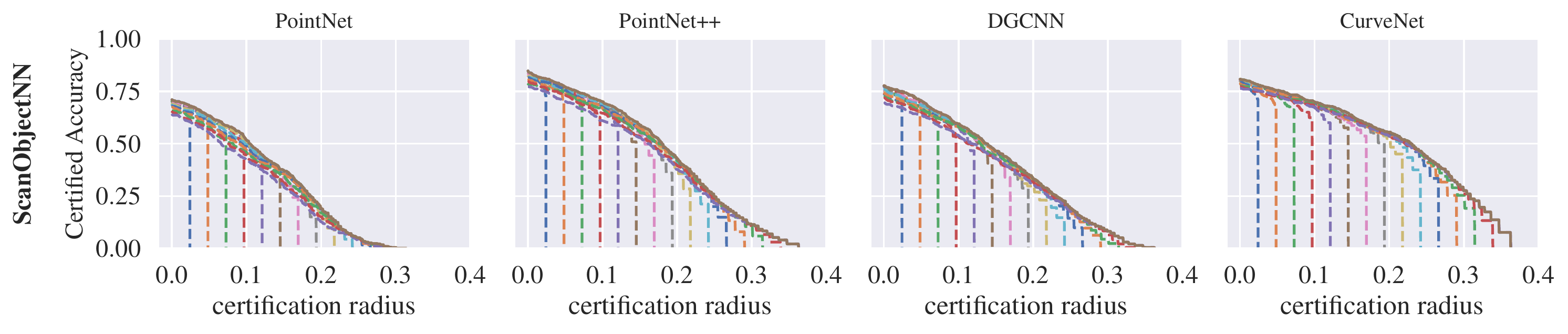}
    \includegraphics[width=0.8\linewidth]{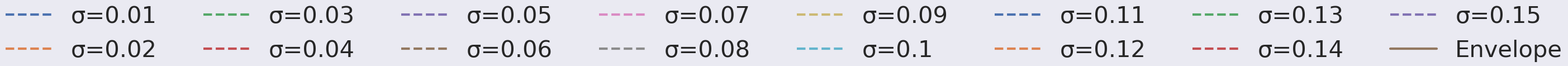}
    \caption{\textbf{Affine (NT)}. Certified accuracy for PointNet, PointNet++, DGCNN and CurveNet on ModelNet40 and ScanObjectNN.}
    \label{fig:SuppAffineNoTranslation}
\end{figure*}

\begin{figure*}[ht]
    \centering
    \includegraphics[width=0.95\linewidth,trim=0 1.5cm 0 0, clip]{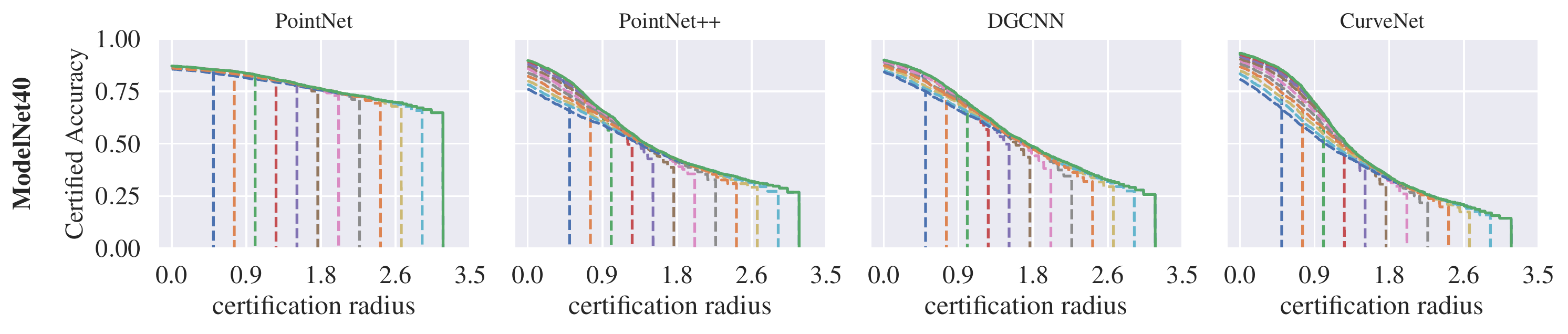}
    \includegraphics[width=0.95\linewidth,trim=0 0.3cm 0 0.55cm,clip]{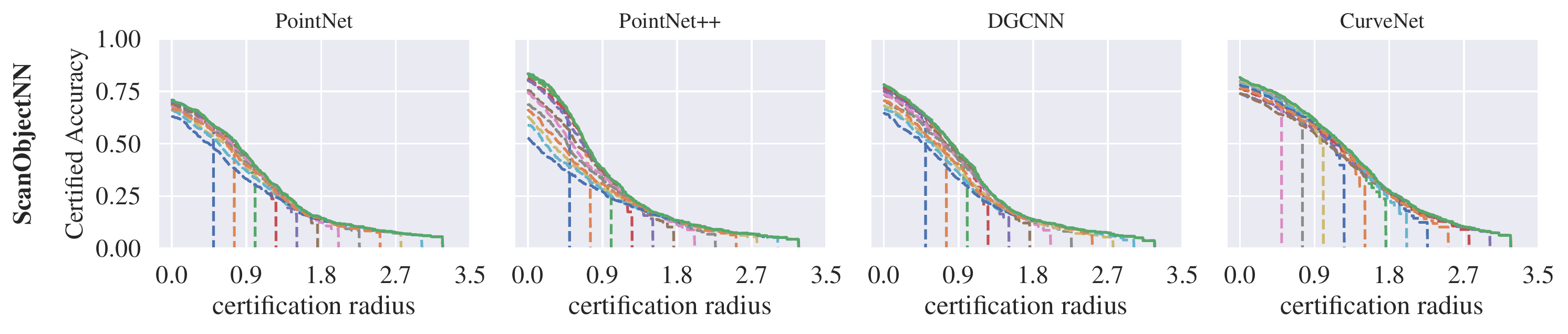}
    \includegraphics[width=0.8\linewidth]{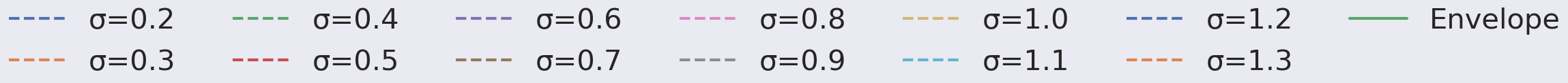}
    \caption{\textbf{Twisting}. Certified accuracy for PointNet, PointNet++, DGCNN and CurveNet on ModelNet40 and ScanObjectNN.}
    \label{fig:SuppTwisting}
\end{figure*}

\begin{figure*}[ht]
    \centering
    \includegraphics[width=0.95\linewidth,trim=0 1.5cm 0 0, clip]{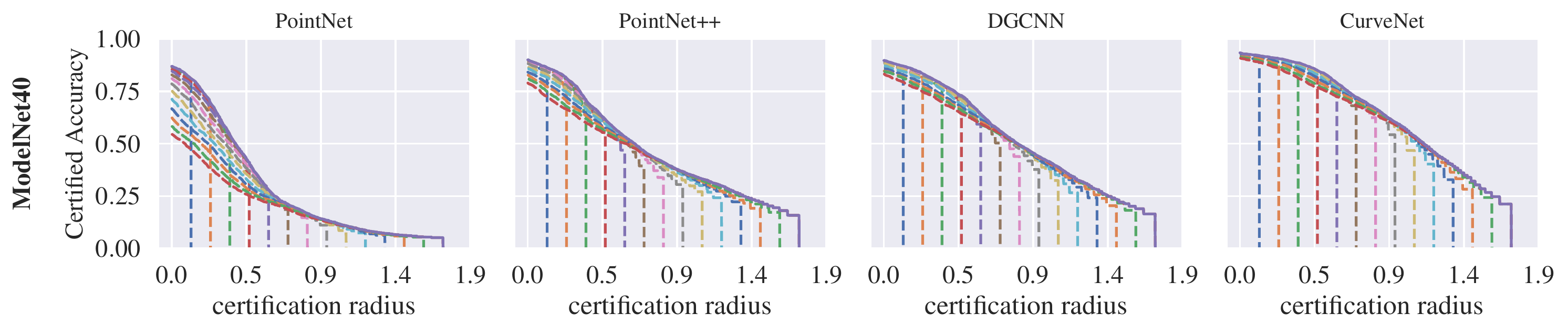}
    \includegraphics[width=0.95\linewidth,trim=0 0.3cm 0 0.55cm,clip]{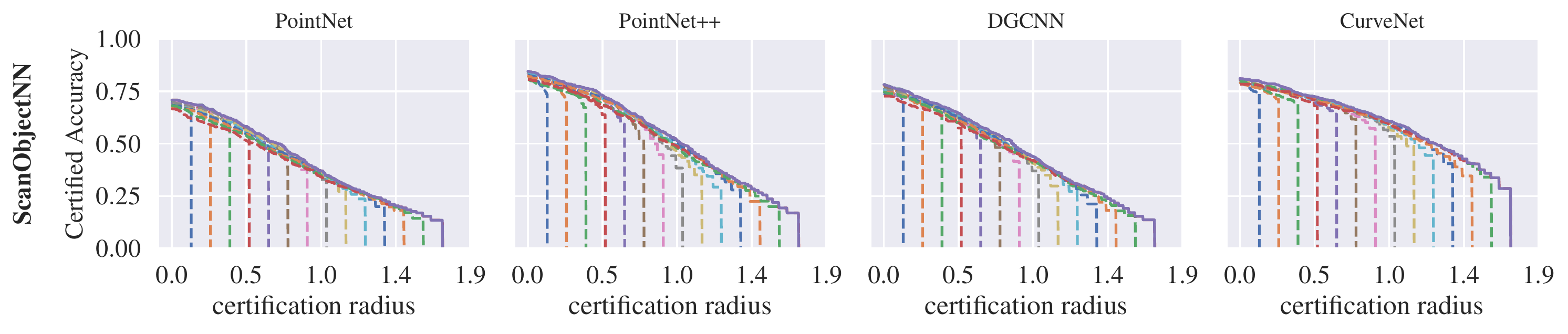}
    \includegraphics[width=0.8\linewidth]{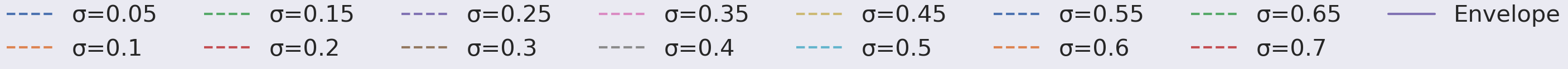}
    \caption{\textbf{Tapering}. Certified accuracy for PointNet, PointNet++, DGCNN and CurveNet on ModelNet40 and ScanObjectNN.}
    \label{fig:SuppTapering}
\end{figure*}

\begin{figure*}[ht]
    \centering
    \includegraphics[width=0.95\linewidth,trim=0 1.5cm 0 0, clip]{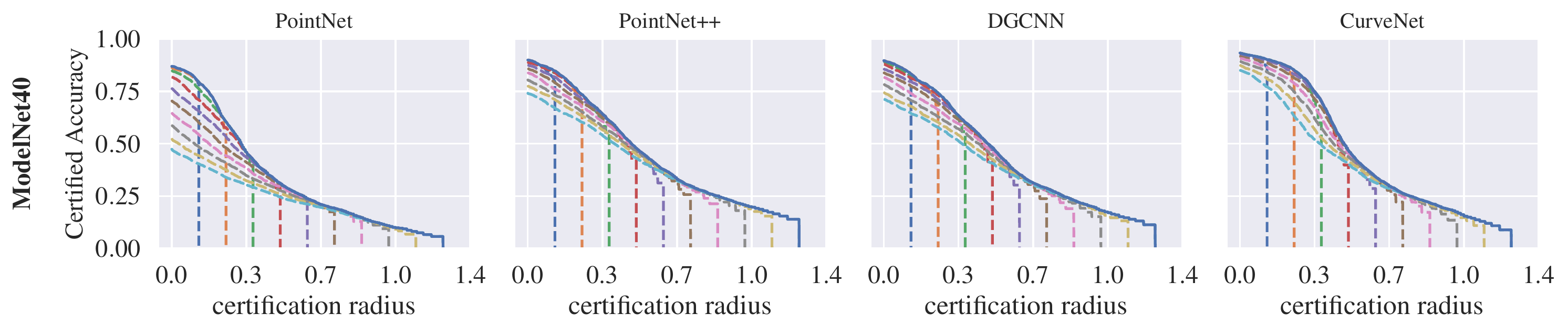}
    \includegraphics[width=0.95\linewidth,trim=0 0.3cm 0 0.55cm,clip]{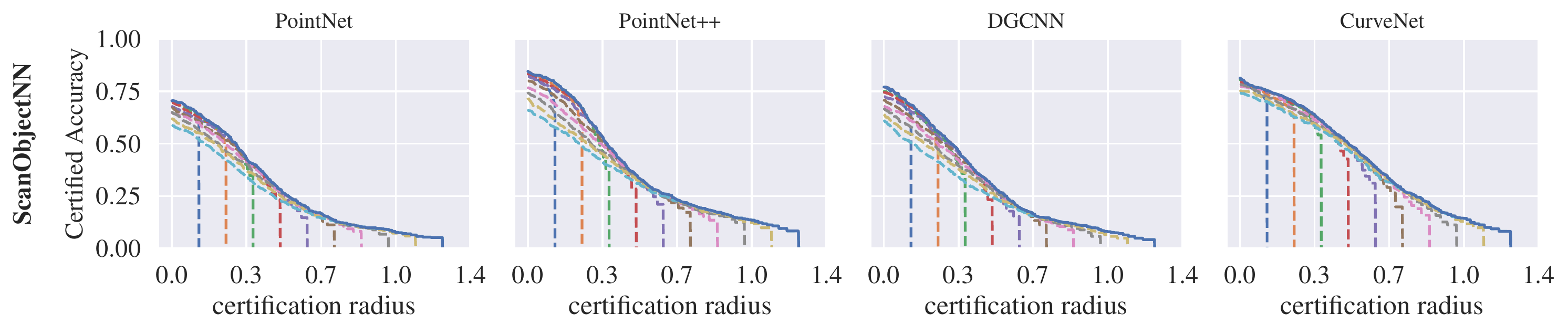}
    \includegraphics[width=0.8\linewidth]{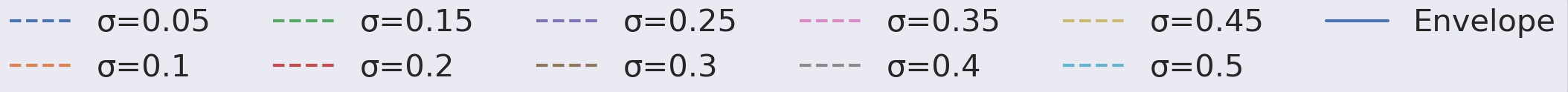}
    \caption{\textbf{Shearing}. Certified accuracy for PointNet, PointNet++, DGCNN and CurveNet on ModelNet40 and ScanObjectNN.}
    \label{fig:SuppShearing}
\end{figure*}

\begin{figure*}[ht]
    \centering
    \includegraphics[width=0.95\linewidth,trim=0 1.5cm 0 0, clip]{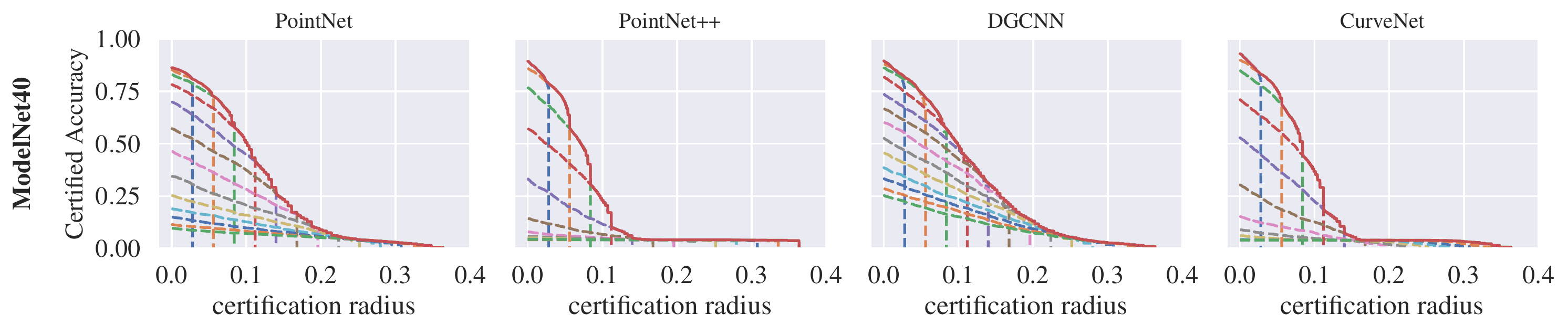}
    \includegraphics[width=0.95\linewidth,trim=0 0.3cm 0 0.55cm,clip]{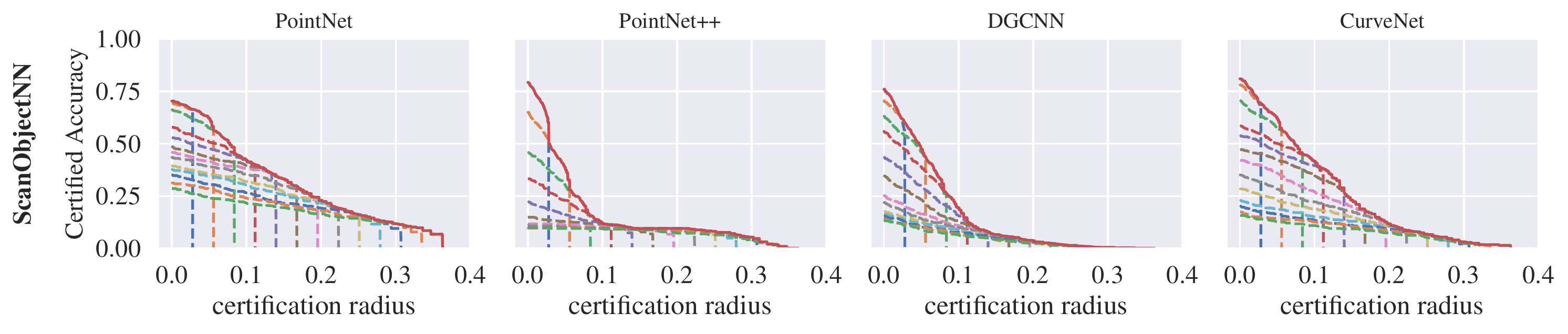}
    \includegraphics[width=0.8\linewidth]{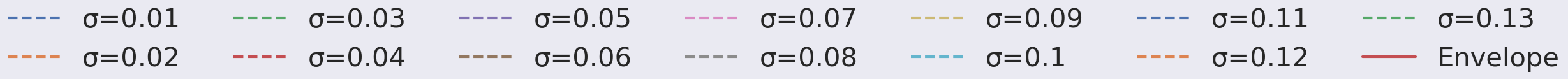}
    \caption{\textbf{Gaussian Noise}. Certified accuracy for PointNet, PointNet++, DGCNN and CurveNet on ModelNet40 and ScanObjectNN.}
    \label{fig:SuppGaussianNoise}
\end{figure*}

Next, we provide a comprehensive and detailed analysis of all the results reported in Section~\ref{sec:benchmarking}.
The envelope certified accuracy curves are reported in Figures~\ref{fig:MainCurveModelNet40} and~\ref{fig:MainCurveScanObjectNN}, and Table~\ref{tab:MainACR} in the main paper. Figures~\ref{fig:SuppRotationXYZ}--\ref{fig:SuppGaussianNoise} in this supplementary material also detail the certified accuracy curve per deformation, with individual $\sigma$ values.
Here we report analyses of the main observations we draw for each deformation.

\paragraph{Rotations}
\underline{\textit{Rot Z}}. 
In ModelNet40, PointNet is vastly superior to all other point cloud DNNs against rotations: its ACR is over 80\% more than the runner-up (PointNet++).
Furthermore, its robust accuracy is, for the most part, maintained across the entire regime we consider (from $-\pi$ to $+\pi$ radians, \emph{i.e.} the entire range of possible rotations).
That is, PointNet correctly classifies most objects independently of their rotation around the $z$ axis.
In stark contrast, PointNet does \textit{not} show this superiority in ScanObjectNN.
As mentioned in the main paper, we attribute this phenomenon almost exclusively to the training augmentations that this version of PointNet enjoyed.
Furthermore, we also find that other point cloud DNNs also show some robustness against rotations in the $z$ axis (although to a lesser extent).
That is, there is a non-negligible amount of objects that point cloud DNNs can accurately recognize regardless of $z$ rotations.
\underline{\textit{Rot XZ}}. 
Combining $z$ rotations with $x$ rotations dramatically changes the phenomena we observed with $z$ rotations alone.
In particular, PointNet's superiority in ModelNet40 entirely disappears: while its certified accuracy was larger than that of other point cloud DNNs for most $z$ rotations, this is no longer the case in Rot XZ, where PointNet now ranks last with an ACR of $0.31$.
In turn, CurveNet displays the largest robustness with an ACR of $0.39$.
PoinNet's drop in ranking, from first to last, remarks the generalization problem of training augmentations.
Specifically, we interpret this observation as follows: despite the robustness boost that training augmentations can provide against a target transformation, they may catastrophically fail to generalize to different (yet semantically similar) transformations.
Notably, we also observe that CurveNet enjoys a larger performance margin (compared to the other DNNs) in ScanObjectNN than in ModelNet40, suggesting CurveNet's architecture generalizes well to the more realistic objects from ScanObjectNN.
\underline{\textit{Rot XYZ}}.
Moreover, introducing another rotation preserves the patterns from Rot XZ: CurveNet shows better performance than other models, while PointNet shows the worse performance of all.
Notably, for the most general XYZ rotation, CurveNet's performance is about 30\% larger than that of the runner-up (DGCNN).
Furthermore, we also note that, for most of the regime we consider, DGCNN is superior to PointNet++ in ModelNet40.
However, this order is reversed when experimenting in ScanObjectNN.

\paragraph{Translation}
For each dataset, the ranking of models is preserved across the entire range of values we consider: for ModelNet40 the ranking is CurveNet, PointNet++, DGCNN, and PointNet; on the other hand, for ScanObjectNN the ranking is PointNet++, CurveNet, DGCNN and PointNet.
That is, we observe that \textit{(i)} PointNet is the worse performer in both cases, and \textit{(ii)} CurveNet is superior to PointNet++ in ModelNet40 (although this order is reversed in ScanObjectNN).

\begin{table*}[t]
    \centering
    \resizebox{\linewidth}{!}{
    \begin{tabular}{c|l|c||c|l|c}
    \toprule
    \midrule
        Name & Deformation Flow & $\phi$ & Name & Deformation Flow & $\phi$  \\
        \midrule
        $x-$ Rotation
        & $\begin{array}{l}
          \Tilde{x} = 0  \\
          \Tilde{y} = (c_\alpha - 1) y - s_\alpha z \\
          \Tilde{z} = s_\alpha y + (c_\alpha-1) z
        \end{array}$
        & $\begin{bmatrix} \alpha \end{bmatrix}$  &
        $xz-$ Rotation
        & $\begin{array}{l}
          \Tilde{x} = (c_\gamma  - 1) x - s_\gamma c_\alpha y + s_\gamma s_\alpha z  \\
          \Tilde{y} = (s_\gamma )x + (c_\gamma c_\alpha -1) y - c_\gamma s_\alpha z  \\
          \Tilde{z} = s_\alpha y + ( c_\alpha -1) z
        \end{array}$
        & $\begin{bmatrix} \alpha \\ \gamma  \end{bmatrix}$ \\
        \midrule
        $y-$ Rotation
        & $\begin{array}{l}
          \Tilde{x} = (c_\beta-1) x + s_\beta z \\
          \Tilde{y} = 0 \\
          \Tilde{z} = -s_\beta x + (c_\beta-1) z
        \end{array}$
        & $\begin{bmatrix} \beta \end{bmatrix}$   &
        $xyz-$ Rotation
        & $\begin{array}{l}
          \Tilde{x} = (c_\gamma c_\beta - 1) x + (c_\gamma s_\beta s_\alpha -s_\gamma c_\alpha)y + (c_\gamma s_\beta c_\alpha +s_\gamma s_\alpha) z  \\
          \Tilde{y} = (s_\gamma c_\beta)x +(s_\gamma s_\beta s_\alpha +c_\gamma c_\alpha -1) y + (s_\gamma s_\beta c_\alpha -c_\gamma s_\alpha)z  \\
          \Tilde{z} = -s_\beta x  + c_\beta s_\alpha y + (c_\beta c_\alpha -1) z
        \end{array}$
        & $\begin{bmatrix} \alpha \\ \beta \\ \gamma \end{bmatrix}$ \\
        \bottomrule
    \end{tabular}
    }
    \caption{
    \textbf{Point flows for rotations along $x-$, $y-$, $xz-$, $xyz-$axes}.
    }
    \label{tab:supp_deformations}
\end{table*}

\begin{table*}[ht]
    \centering
    \resizebox{\textwidth}{!}{
    \begin{tabular}{c|c|c|c|c}
    \toprule
    \midrule
        $x-$Rotation [$\alpha$] & $z-$Rotation [$\gamma$] & $xyz-$Rotation [$\alpha$,$\beta$,$\gamma$] & Translation [$t_x$,$t_y$,$t_z$]  & Affine [$a,\dots,l$] \\ \midrule
        $\begin{bmatrix} 
    	1 & 0 & 1 & 0\\
    	0 & c_\alpha & -s_\alpha & 0\\
    	0 & s_\alpha & c_\alpha & 0\\
    	0 & 0 & 0 & 1\\
    	\end{bmatrix}$ 
    	& 
        $\begin{bmatrix} 
    	c_\gamma & -s_\gamma & 0& 0\\
    	s_\gamma & c_\gamma & 0& 0\\
    	0 & 0 & 1 & 0\\
    	0 & 0 & 0 & 1\\
    	\end{bmatrix}$ 
    	&
    	$\begin{bmatrix}
    	c_\gamma c_\beta &c_\gamma s_\beta s_\alpha -s_\gamma c_\alpha &c_\gamma s_\beta c_\alpha +s_\gamma s_\alpha & 0\\
    	s_\gamma c_\beta &s_\gamma s_\beta s_\alpha +c_\gamma c_\alpha &s_\gamma s_\beta c_\alpha -c_\gamma s_\alpha & 0\\
    	-s_\beta &c_\beta s_\alpha &c_\beta c_\alpha & 0\\
    	0 & 0 & 0 & 1\\
    	\end{bmatrix}$
    	& 
        $\begin{bmatrix} 
    	1 & 0 & 0 & t_x\\
    	0 & 1 & 0 & t_y\\
    	0 & 0 & 1 & t_z\\
    	0 & 0 & 0 & 1\\
    	\end{bmatrix}$ 
    	& 
        $\begin{bmatrix} 
	    a+1 & b   & c   & d \\
    	e   & f+1 & g   & h \\
    	i   & j   & k+1 & l \\
    	0   & 0   & 0   & 1 \\
    	\end{bmatrix}$ 
    	\\
    	\bottomrule \midrule
    	$y-$Rotation ($\beta$)& $z-$Twisting ($\gamma$) & $z-$Tapering  ($a$,$b$)  & $z-$Shearing ($a$,$b$) & Affine (NT) [$a,\dots,k$] \\ \midrule
        $\begin{bmatrix} 
    	c_\beta & 0 & s_\beta & 0\\
    	0 & 1 & 0 & 0\\
    	-s_\beta  & 0& c_\beta & 0\\
    	0 & 0 & 0 & 1\\
    	\end{bmatrix}$ 
    	&
        $\begin{bmatrix} 
    	c_{\gamma z} & -s_{\gamma z} & 0& 0\\
    	s_{\gamma z} & c_{\gamma z} & 0& 0\\
    	0 & 0 & 1 & 0\\
    	0 & 0 & 0 & 1\\
    	\end{bmatrix}$ 
    	& 
        $\begin{bmatrix} 
    	\frac{1}{2}a^2z+bz+1 & 0 & 0 & 0\\
    	0 & \frac{1}{2}a^2z+bz+1 & 0 & 0\\
    	0 & 0 & 1 & 0\\
    	0 & 0 & 0 & 1\\
    	\end{bmatrix}$ 
    	& 
        $\begin{bmatrix} 
    	1 & 0 & a & 0\\
    	0 & 1 & b & 0\\
    	0 & 0 & 1 & 0\\
    	0 & 0 & 0 & 1\\
    	\end{bmatrix}$ 
    	& 
        $\begin{bmatrix} 
	    a+1 & b   & c   & 0 \\
    	d   & e+1 & f   & 0 \\
    	g   & h   & i+1 & 0 \\
    	0   & 0   & 0   & 1 \\
    	\end{bmatrix}$ 
    	\\
    	\bottomrule
    \end{tabular}
    }
    \caption{
    \textbf{Transformation matrices $\mathbf{T} \in \mathbb{R}^{4\times4}$ for common perturbations under homogeneous coordinates.}
    A point cloud $p_i$ in homogeneous coordinates is transformed into $\tilde p_i = \mathbf{T}(\phi)\:p_i$.
    $c_\alpha$ and $s_\alpha$ correspond to $\sin(\alpha)$ and $\cos(\alpha)$, respectively.
    ``$xyz-$Rotation'' corresponds to the extrinsic Euler angles in the specific order: ``$x-$Rotation'', ``$y-$Rotation'' then ``$z-$Rotation''.
    }
    \label{tab:Deformations}
\end{table*}

\paragraph{Affine and No-Translation Affine}
In ModelNet40, we notice that the affine and affine (NT) transformations do not share exactly the same DNN ranking: while CurveNet and PointNet are the best and worst performers, respectively, the affine (NT) curve suggests DGCNN performs better than PointNet++, while the affine curve suggests the opposite.
Additionally, we note that the affine transformation (which \textit{includes} translation), suggests exactly the same DNN ranking as the translation transformation.
This observation suggests that the translation transformation is a consequential component in the affine transformation.
On the other hand, for ScanObjectNN, we also see that the DNN ranking suggested by the affine and affine (NT) transformations is not exactly the same: both suggest DGCNN and PointNet are the two worst performers (in that order), but affine (NT) suggests CurveNet performs better than PointNet++, while affine suggests the opposite.
Finally, analogous to our observation of these transformations on ModelNet40, we notice that the translation and affine transformations suggest the \textit{same} DNN ranking.

\paragraph{Twisting}
The twisting transformation we considered is around the $z$ axis.
In ModelNet40, PointNet shows remarkable performance superiority (compared to the other DNNs) against twisting. 
As mentioned in the main paper, we attribute this phenomenon to the $z$ rotation augmentations with which PointNet was trained.
Regarding the rest of the DNNs, we observe that CurveNet (analogous to the results on $z$ rotation) also displays low performance for most rotations.
This last observation may suggest that ModelNet40 has some idiosyncratic properties that induce noise into the assessment, since, in ScanObjectNN, CurveNet is remarkably superior to other methods for almost all rotation magnitudes.

\paragraph{Tapering}
In both ModelNet40 and ScanObjectNN, CurveNet and PointNet are, respectively, the best and worst performers.
Remarkably, the slope of the curves, for both datasets (but specially for ScanObjectNN), are rather smooth, showing a slow but steady decrease in performance as the transformation grows stronger.
This observation suggests that point cloud DNNs possess somewhat of an inherent robustness against tapering.

\paragraph{Shearing}
For the entire regime we consider, PointNet is the worst performer in both datasets.
In ModelNet40, the other DNNs, \ie PointNet++, DGCNN and CurveNet, have similar performances for the entire regime.
In ScanObjectNN, on the other hand, the performances display larger variation across DNNs.

\paragraph{Gaussian Noise}
In ModelNet40, performances drop rapidly as the perturbation's magnitude increases.
These drops are, however, consistent across most of the regime we consider.
In particular, we notice that DGCNN performs slightly better than PointNet; in turn, PointNet performs better than CurveNet which performs better than PointNet++.
The performances of all DNNs change dramatically when testing on ScanObjectNN.
Specifically, the performance drops in ScanObjectNN are not as rapid as in ModelNet40.
Remarkably, PointNet, which consistently scored the worst in most of the transformations, scores the \textit{best} against Gaussian noise, displaying superior performance against the other DNNs for almost the entirety of the deformation regime.
This observation may invite for a thorough study into why, apparently, there is little (or even inverse) correlation between robustness against spatial transformations and per-point perturbations with Gaussian noise.

\section{Formulation for other Rotations}
Table~\ref{tab:supp_deformations} reports the detailed formulation for the more complex types of rotations we considered in our work, \ie $xz$-Rotation \& $xyz$-Rotation.

\section{Homogeneous Coordinates}
Table~\ref{tab:Deformations} reports details into how the deformations we considered in our work are modeled in the standard homogeneous-coordinates form.

\begin{figure*}[ht]
    \centering
    \includegraphics[trim=8cm 0 8cm 0,width=.7\textwidth]{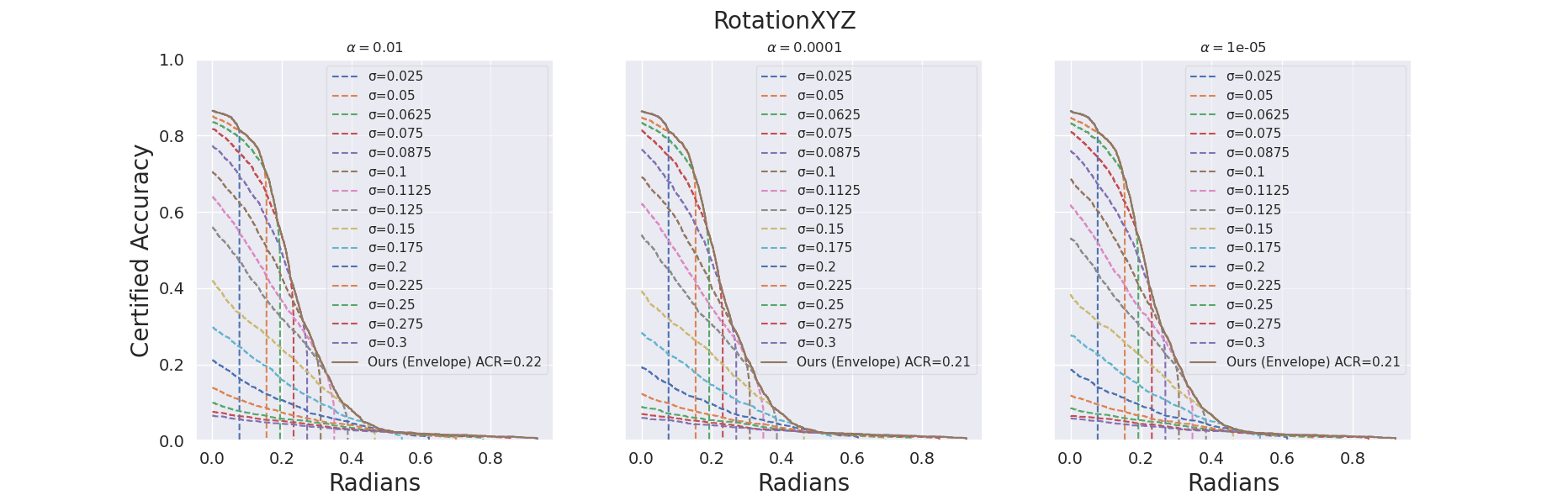}
    \caption{
    \textbf{Certified accuracy curves with failure ratio $\alpha \in [10^{-2},10^{-4},10^{-5}]$ under $xyz$-Rotation deformations using 3DeformRS.}
    Our certification method appears to be insensitive to the failure ratio $\alpha$.
    }
    \label{fig:alphaComparison}
    \vspace{-.3cm}
\end{figure*}

\section{Curves for Failure Ratios $\alpha$}
Figure~\ref{fig:alphaComparison} reports the certified accuracy curves for the failure ratios $\alpha$ we considered in Section~\ref{sec:ablations}.

\section{3DCertify comparison}

\subsection{Locally-run experiments}

We compared against 3DCertify for Rotations around single axes.
Whilst the important $z$-Rotation results were reported in 3DCertify, we ran local experiments for $x$-Rotation and $y$-Rotation using their public implementation (available at \texttt{https://github.com/eth-sri/3dcertify}). 

\begin{table}[t]
\centering
\begin{tabular}{c|c|c|c}
\toprule
Degrees   & RotX & RotY & RotZ \\ \hline
1  & 0.98 & 0.99 & -- \\ 
2  & 0.96 & 0.99 & -- \\ 
3  & 0.96 & 0.99 & 0.99 \\ 
4  & 0.96 & 0.98 & -- \\ 
5  & 0.93 & 0.96 & -- \\ 
6  & 0.93 & 0.96 & -- \\ 
7  & 0.88 & 0.94 & -- \\ 
8  & 0.84 & 0.90 & -- \\ 
10 & 0.79 & 0.84 & -- \\ 
15 & 0.54 & 0.67 & -- \\ 
20 & --   & --   & 0.97 \\ 
60 & --   & --   & 0.96 \\ \bottomrule
\end{tabular}
\caption{\textbf{Certified Accuracy from 3DCertify.} Taylor relaxation and improved MaxPool layer with DeepPoly for single-axis rotations.}
\label{tab:oneAxis3DCertifyCertifiedAcc}
\end{table}

Table~\ref{tab:oneAxis3DCertifyCertifiedAcc} reports the certified accuracy achieved by 3DCertify.
We used 3DCertify's original experimental setup (Taylor relaxation with $2^\circ$ splits).

Apart from the  $x$-Rotation and $y$-Rotation experiments, we also computed a single value for $z$-Rotation ($3^\circ$) to verify consistency with the larger values reported in 3DCertify ($20^\circ$ and $60^\circ$).
These results are reported in Figure~\ref{fig:64pointnetRotationZ}.

\begin{table}[t]
\centering
\begin{tabular}{c|c|c|c}
\toprule
Degrees   & RotX & RotY & RotZ \\ \hline
1  & 13346 & 14095 & -- \\ 
2  & 26614 & 26785 & -- \\ 
3  & 38960 & 39247 & 32832 \\ 
4  & 49931 & 49803 & -- \\ 
5  & 58679 & 58699 & -- \\ 
6  & 66781 & 66460 & -- \\ 
7  & 73440 & 73238 & -- \\ 
8  & 79223 & 78798 & -- \\ 
10 & 89327 & 88744 & -- \\ 
15 & 89974 & 93572 & -- \\ 
20 & --   & --   & -- \\ 
60 & --   & --   & -- \\ \bottomrule
\end{tabular}
\caption{\textbf{Running Time (seconds) for 3DCertify.} Taylor relaxation and improved MaxPool layer with DeepPoly for single-axis rotations.}
\label{tab:oneAxis3DCertifyRuntime}
\end{table}

In Table~\ref{tab:oneAxis3DCertifyRuntime} we report the runtimes for all experiments.
It is worth to notice the proportionality between the runtimes and the deformation parameter: certifying large deformation parameters would take significantly longer.
These values oscillate in the tenths of thousands of seconds and, beyond $15^\circ$, runtimes above a hundred thousand seconds are expected.

\subsection{Point Sampling}
\begin{figure*}[ht]
    \centering
    \includegraphics[width=\linewidth]{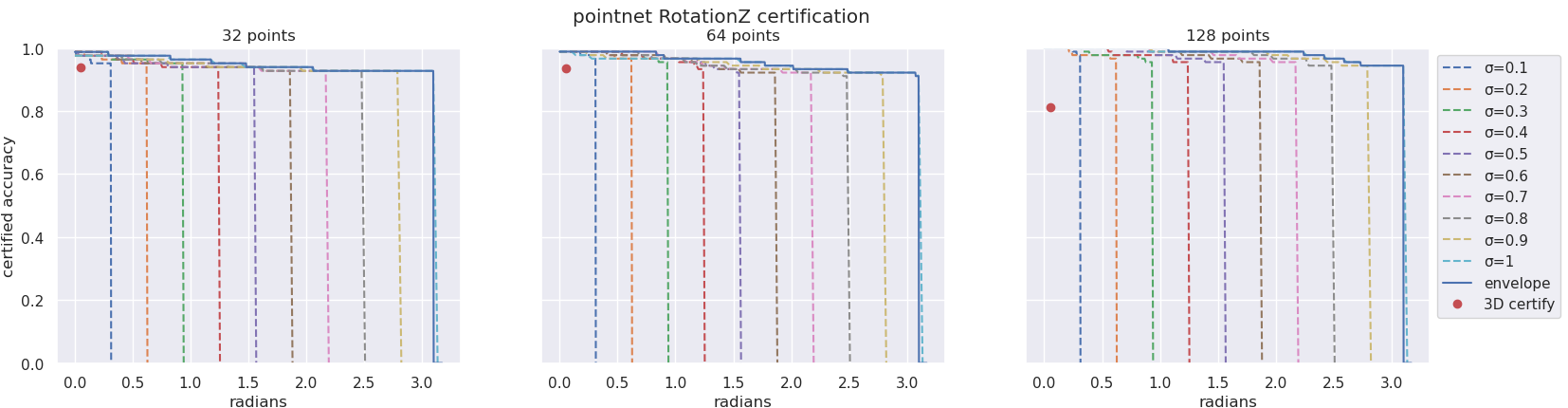}\\
    \includegraphics[width=\linewidth]{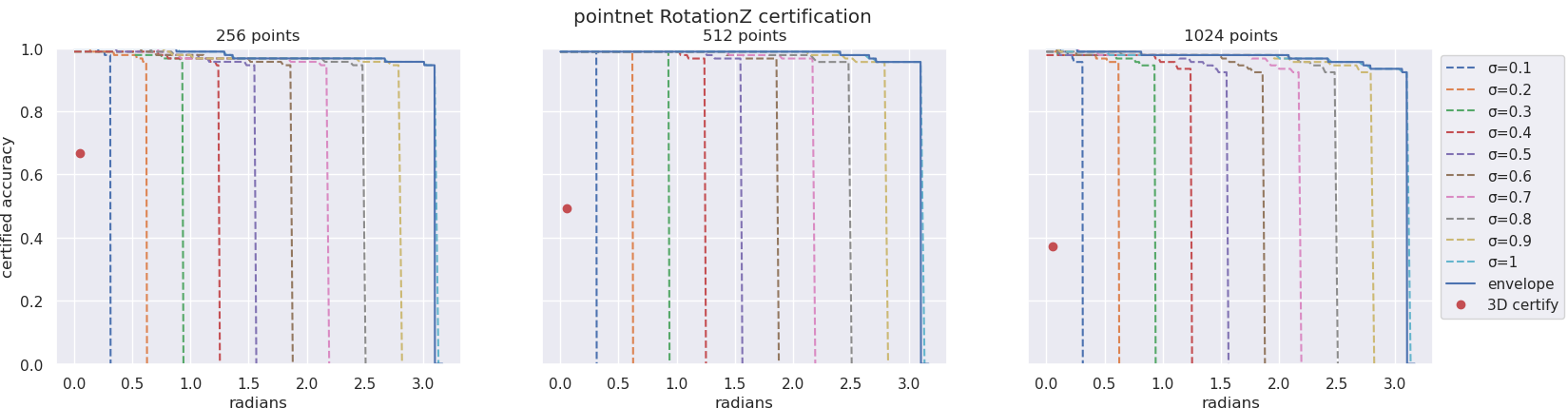}
    \caption{
    \textbf{Certification under different Point Cloud Cardinality.}
    Larger point sampling leads to severe loss in certification ability for 3DCertify. 
    In contrast, the variation of 3DeformRS performance  is minimal.}
    \label{fig:PointAmountComparison}
\end{figure*}

As seen on Table~\ref{tab:diffSamplVerif}, 3DeformRS outperforms previous certification approaches based on linear relaxations for $3^\circ$ $z$-Rotations.
Furthermore, Figure~\ref{fig:PointAmountComparison} shows how 3DeformRS consistently certifies PointNet for all $z$-Rotations regardless of point cloud cardinality.
Note that we used the weights for PointNet provided by the authors, pre-trained with the corresponding point cloud size.

\newpage

\begin{figure*}[ht]
    \centering
    \includegraphics[width=0.95\linewidth,trim=0 1.5cm 0 0, clip]{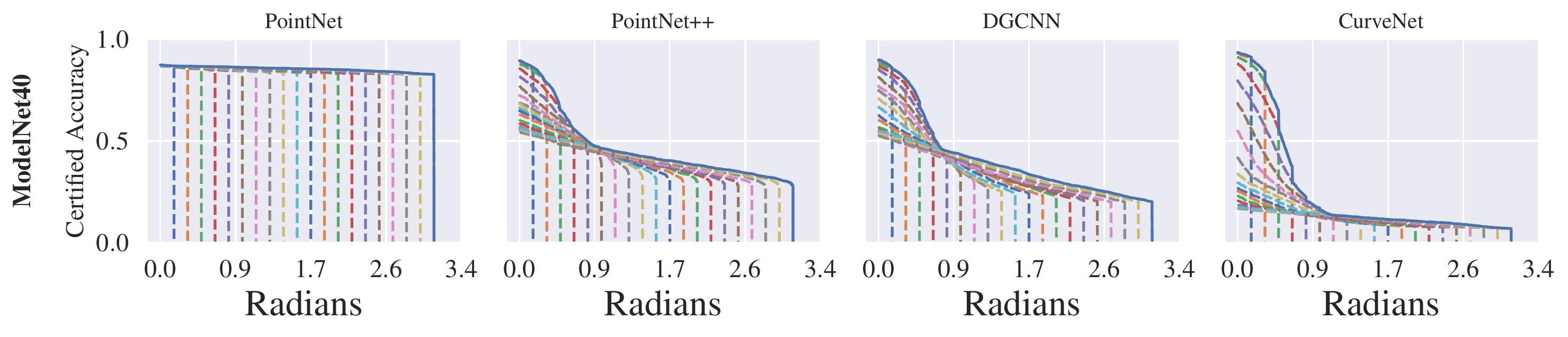}
    \includegraphics[width=0.95\linewidth,trim=0 0.3cm 0 0.55cm,clip]{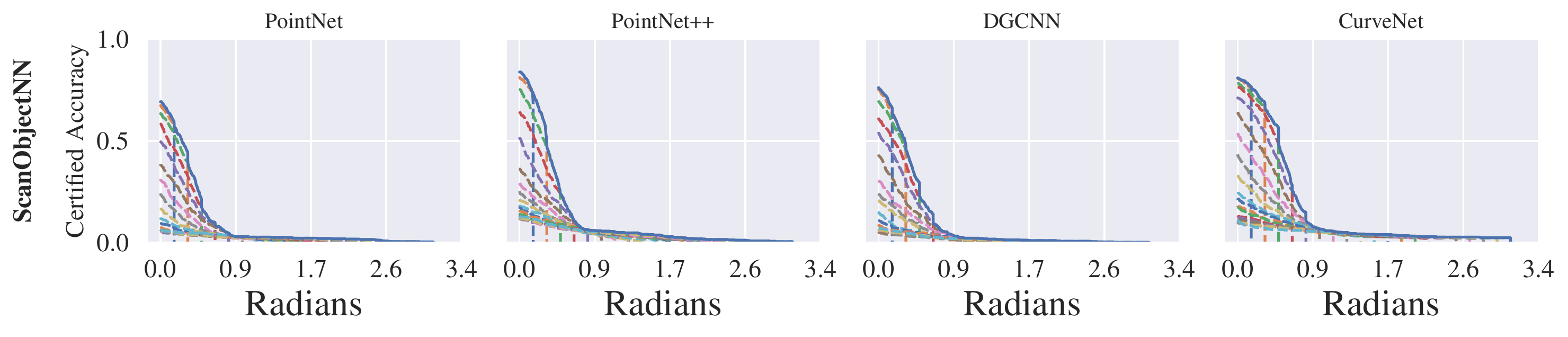}
    \includegraphics[width=0.8\linewidth]{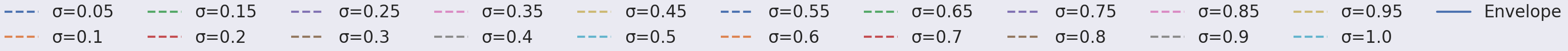}
    \caption{\textbf{RotationZ}. Certified accuracy for PointNet, PointNet++, DGCNN and CurveNet on ModelNet40 and ScanObjectNN.}
    \label{fig:SuppRotationZ}
\end{figure*}

\begin{figure*}[ht]
    \centering
    \includegraphics[width=0.95\linewidth,trim=0 1.5cm 0 0, clip]{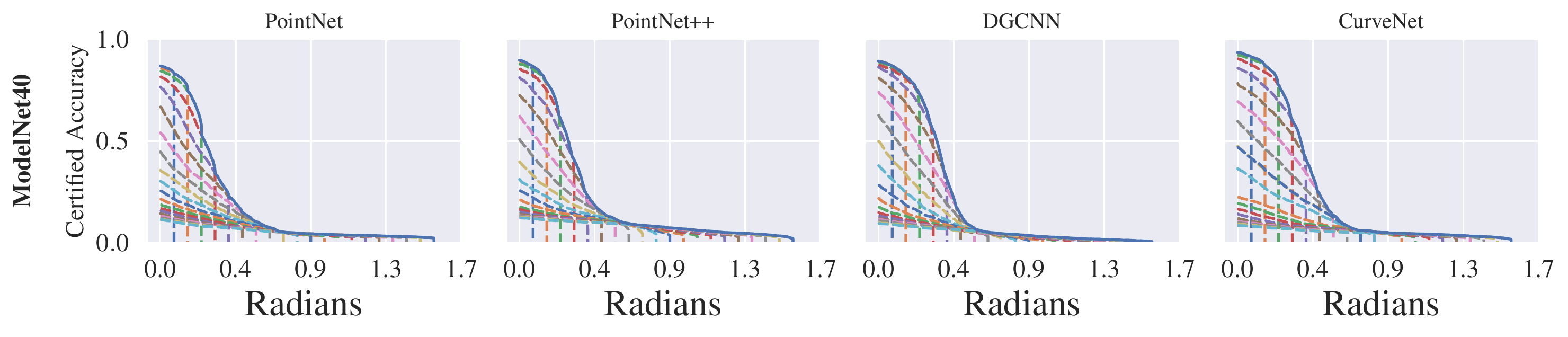}
    \includegraphics[width=0.95\linewidth,trim=0 0.3cm 0 0.55cm,clip]{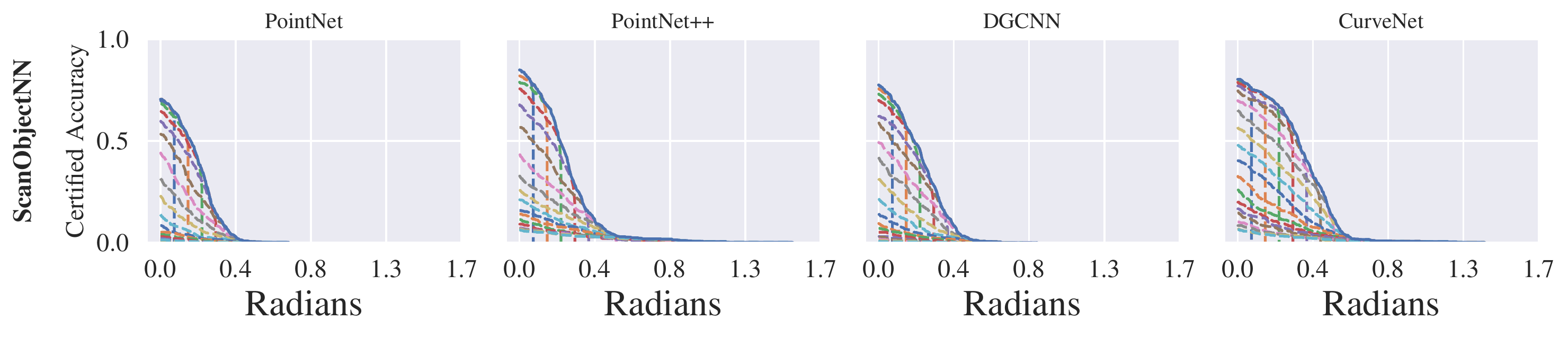}
    \includegraphics[width=0.8\linewidth]{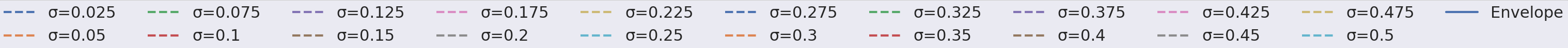}
    \caption{\textbf{RotationXZ}. Certified accuracy for PointNet, PointNet++, DGCNN and CurveNet on ModelNet40 and ScanObjectNN.}
    \label{fig:SuppRotationXZ}
\end{figure*}

\end{document}